\documentclass[fleqn, 10pt]{wlscirep}
\usepackage[T1]{fontenc}
\usepackage{multirow}
\usepackage{tabularx}

\title{Monocular markerless biomechanics for clinically interpretable gait assessment in spinal cord injury}

\usepackage{lineno}
\usepackage{todonotes}
\usepackage{changepage}
\usepackage{subcaption}
\usepackage{graphicx}
\usepackage[table]{xcolor} 
\usepackage{overpic}
\usepackage{amsmath}
\usepackage{float}
\usepackage{placeins} 
\usepackage{mdframed}
\usepackage{bm}

\renewcommand{\textit}[1]{#1}

\usepackage{hyperref}
\newcommand{\figref}[1]{\hyperref[#1]{Figure~\ref*{#1}}}
\newcommand{\tabref}[1]{\hyperref[#1]{Table~\ref*{#1}}}
\newcommand{\sfigref}[1]{\hyperref[#1]{Supplementary Figure~\ref*{#1}}}
\newcommand{\stabref}[1]{\hyperref[#1]{Supplementary Table~\ref*{#1}}}
\newcommand{\eqhref}[1]{\hyperref[#1]{Equation~\ref*{#1}}}

\usepackage{silence}
\author[1,2*]{Shreyasvi Natraj}
\author[1,2]{Mathieu Ruepp}
\author[1,2]{Yanke Li}
\author[1,3,4]{Robert Riener}
\author[2,5,6]{Inge Eriks-Hoogland}
\author[1,2,7*]{Diego Paez-Granados}
\affil[1]{Spinal Cord Artificial Intelligence (SCAI) Lab, ETH Zürich, Zürich, Switzerland}
\affil[2]{Swiss Paraplegic Research, Swiss Paraplegic Centre, Nottwil, Switzerland}
\affil[3]{University Hospital Balgrist, Zurich, Switzerland}
\affil[4]{Graduate School of Engineering, Tohoku University, Sendai, Japan}
\affil[5]{Faculty of Health Sciences and Medicine, University of Lucerne, Lucerne, Switzerland}
\affil[6]{Swiss Paraplegic Centre, Nottwil, Switzerland}
\affil[7]{University of Tsukuba, 305-8577, Ibaraki, Tsukuba, Japan}

\affil[*]{Correspondence to: {shreyasvi.natraj@hest.ethz.ch, diego.paez}@hest.ethz.ch}

\begin{abstract}

Three-dimensional gait analysis guides rehabilitation after spinal cord injury but depends on marker-based motion capture and force plates, which few clinics have. Monocular markerless pipelines have been established in fewer healthy adult cohorts but not in neurological cohorts. We present the \texttt{SCAI SCI Gait} dataset, comprising 239 adult individuals with spinal cord injury with synchronized video, motion capture, and force-plate measurements, we fitted a parametric body mesh to a single sagittal-view video, driving an anthropometrically scaled \texttt{OpenSim} model via virtual markers. Markerless lower-body kinematics showed state-of-the-art agreement with motion-capture measurements ($r = 0.68\text{--}0.90$, $p < 0.001$, and $\mathrm{RMSE} = 4.18\text{--}6.49^{\circ}$), and accurate kinematics-based predicted ground-reaction forces closely matched those measured by force plates ($r = 0.85\text{--}0.87$, $p < 0.001$, and $\mathrm{RMSE} = 2.13\text{--}2.19\,\mathrm{N\,kg^{-1}}$). Conditional-dependence graph analysis with Markov blankets revealed that waveform components were conditionally associated with functional independence, and speed-stratified clustering revealed distinct mechanical strategies among individuals walking at similar speeds. These findings establish the use of monocular video as a scalable approach for clinically meaningful biomechanical assessment and data-driven phenotyping in patients with spinal cord injury. The dataset and code is available at \url{https://github.com/SCAI-Lab/SCAI-SCI-Gait-Dataset}.

\end{abstract}

\begin{document}

\flushbottom
\maketitle
%
%
\thispagestyle{empty}

\section*{Introduction}

Three-dimensional gait analysis provides a quantitative description of walking that helps distinguish primary neuromotor deficits from compensatory strategies such as hip hiking, circumduction and trunk lean, and it links movement patterns to joint loading, falls, and secondary injury\cite{McGuirk2022, Albani2014}. In routine clinical practice, however, observational assessment often cannot determine whether a deviation is a cause or a consequence of weakness, spasticity, pain, or balance limitation, and similar outward patterns can reflect different underlying mechanisms\cite{Baker2013MeasuringWalking, Robertson2013}. Clinical 3D gait analysis addresses this ambiguity by combining kinematics with external loading (GRFs and CoP) across the gait cycle, enabling clinically interpretable joint-angle and joint-moment estimates that can guide rehabilitation, orthoses, and surgery, and that can support longitudinal outcomes assessment when gait speed changes are difficult to anticipate\cite{Winter2009, McGuirk2022, Hulleck2022}. In spinal cord injury (SCI), where impairment ranges from subtle interlimb asymmetry to profound paresis with assistive device use, quantitative gait analysis can reveal locomotor function beyond broad clinical scales, sensitively capture change over time, and help characterize strategy-level differences among individuals walking at similar speeds\cite{Werner2021, Werner2023}. Conventionally, this is performed with multicamera, marker-based motion capture (MoCap) and force plates\cite{Baker2013MeasuringWalking, Winter2009}. Despite its clinical utility and integration with musculoskeletal toolchains such as \texttt{OpenSim}\cite{OpenSim}, MoCap is expensive, labor intensive, and operator dependent, which limits frequent assessment and broader clinical deployment\cite{Hulleck2022, McGuirk2022, Robertson2013}. Moreover, soft-tissue artifacts and marker-placement variability can introduce nontrivial joint-angle errors, with downstream impacts on derived joint moments\cite{Leardini2005STA, PETERS20101, LEBOEUF2023111774, GORTON2009398, Wade2022}.

\begin{figure}[t]
\centering
\includegraphics[width=1\linewidth]{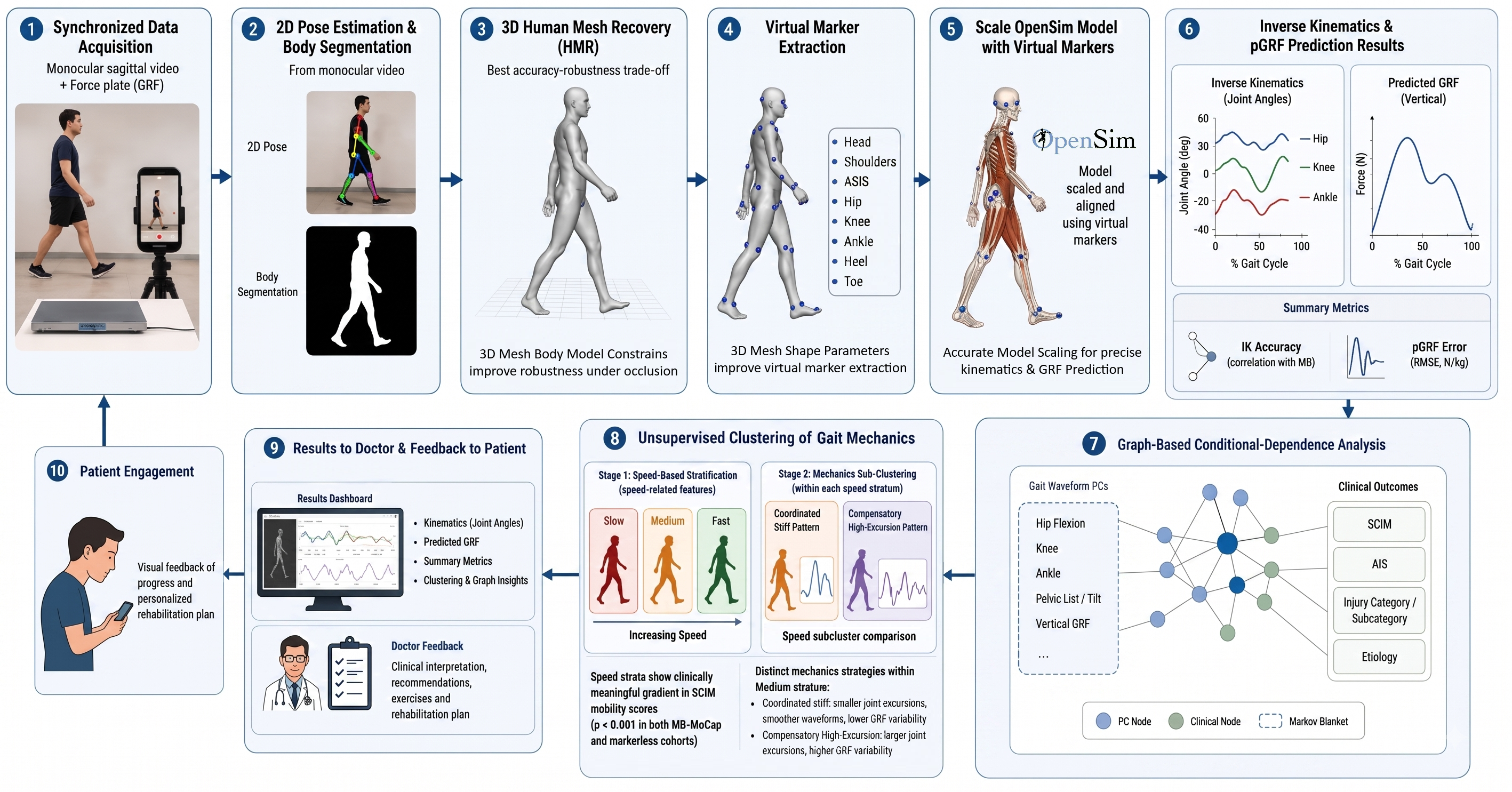}
\caption{\small{Monocular markerless gait analysis workflow and downstream clinical inference used in this study. (1) Synchronized monocular sagittal-plane video and force-plate GRF acquisition. (2) Video preprocessing with 2D pose estimation and body segmentation. (3) Monocular 3D human-mesh recovery (HMR) with body-model constraints for robust occlusion. (4) Extraction of virtual markers from the 3D mesh (head, shoulders, ASIS, hip, knee, ankle, heel, and toe). (5) Participant-specific OpenSim model scaling/alignment to enable kinematics and GRF prediction. (6) Inverse kinematics (IK) yields joint-angle waveforms; kinematics-based prediction of vertical GRFs (pGRFs) provides kinetic estimates; performance is summarized by IK accuracy (correlation with MB) and pGRF error (RMSE, N/kg). (7) Graph-based conditional-dependence analysis links principal components of kinematic/GRF features to clinical outcomes (SCIM, AIS, injury category/subcategory, and etiology), with Markov blankets highlighting conditionally relevant associations. (8) Two-stage unsupervised clustering: speed stratification (slow/medium/fast) followed by within-stratum mechanics subclustering to identify coordination/compensatory strategies. (9) Results dashboard for clinicians (kinematic waveforms, predicted GRFs, summary metrics, and clustering/graph insights) to support interpretation and rehabilitation planning. (10) Patient-facing visual feedback and a personalized rehabilitation plan close the loop between quantitative gait assessment, clinician guidance, and rehabilitation planning.} }
\label{fig:intro_sn}
\end{figure}

Over the past decade, several approaches have been developed for higher-fidelity 3D kinematics extraction by pairing 2D
detection with multi-camera calibration and triangulation, as implemented in commercial systems (\texttt{Theia3D}) and open-source workflows such as \texttt{OpenCap} and \texttt{Pose2Sim} \cite{opencap, Pagnon_2022_JOSS, KANKO2021110665}. Across studies, these multi-view keypoint pipelines show encouraging agreement with MoCap for many spatiotemporal outcomes and for several sagittal-plane joint-angle waveforms, whereas the agreement is less consistent for distal joints, non-sagittal-plane rotations, and patient cohorts \cite{Riazi2022, pose2sim_accuracy, Scataglini24, SEUTHE2025113008, Dsouza2024, Wade2022, Pardell24, Lam2023}. This has supported semi-markerless gait analysis for clinical outcomes assessment, where video provides 3D kinematics while GRFs are still measured using force plates so that inverse dynamics can
be retained and the clinical implications of kinematic error for joint moments and loading can be quantified \cite{Dsouza2024, Song2023, Yang27052025, Rode2025SciRep, McGuirk2022}.

To reduce reliance on multi-camera infrastructure and enable more accessible deployment in clinical settings, a large body of work targets 3D pose estimation from fewer views, including monocular video. \emph{2D-to-3D lifting} maps 2D keypoints to 3D joint coordinates under learned kinematic and anthropometric priors \cite{Martinez2017Simple3D}. For gait, temporal models further exploit motion continuity to stabilize trajectories, reduce frame-to-frame jitter, and partially resolve depth ambiguity in the presence of self-occlusion \cite{Pavllo2019VideoPose3D}. Despite rapid progress, monocular 3D poses remain limited by scale ambiguity, out-of-plane rotations, and viewpoint-dependent occlusions, leading to keypoint noise and occasional mistracking that abruptly shift joints, yielding anatomically implausible configurations such as extreme limb angles and inconsistent limb lengths compared with those of multi-camera approaches \cite{3dpose_fail}. These are exacerbated in real-world gait videos with assistive devices and atypical postures \cite{GUO2025}.

Building upon these pose estimation advances, computer vision models have also shifted from sparse joints to dense, person-specific body reconstructions via human mesh recovery (HMR) and parametric models such as \texttt{SMPL}, originally developed for graphics and virtual avatar creation \cite{Loper2015, Kanazawa2018HMR, Tian2023, MUHAMMAD2022103628}. With respect to biomechanics, meshes are appealing because they provide temporally consistent surface correspondences that can, in principle, be mapped to anatomically motivated virtual markers and used to drive musculoskeletal modeling toolchains. Recent ``in-the-wild'' monocular biomechanics pipelines such as \texttt{OpenCap} \textit{Monocular} combine single-view video with learned body-model priors and optimization/smoothing to recover kinematics without a dedicated multi-camera lab setup \cite{OpenCapMonocular2026}. Complementary directions integrate stronger foreground/part segmentation and video object tracking (segmentation-driven capture such as \texttt{SAM4DCap}) \cite{sam4dcap} to improve robustness under occlusion, truncation, and complex backgrounds, which are common in clinical gait recordings with assistive devices. In practice, however, mesh-to-skeleton conventions, virtual-marker definitions, and model-scaling procedures are not yet standardized, and mesh-based monocular gait pipelines remain under-validated for atypical or neurological gait, which limits confidence for clinical interpretation \cite{Ueno2024Heliyon, OpenCapMonocular2026, HORSAK2023111801, HORSAK2025112986}. A scope comparison of representative systems and their support for \texttt{OpenSim}, GRFs/kinetics, clinical phenotyping, and SCI evaluation is provided in \stabref{tab:system_comparison}.

Even with accurate kinematics, real-world biomechanics remain constrained by the need for force instrumentation. Standard inverse dynamics require the GRFs and center-of-pressure (CoP) to compute joint moments, which are central to clinically interpretable estimates of joint loading; however, force plates are expensive, spatially restrictive, and difficult to deploy in remote clinics or homes. To mitigate this bottleneck, GRF estimation from motion has been studied in several settings, including multi-camera (multi-view) markerless capture, where 3D kinematics are well constrained \cite{Lichtwark2024GRF}, as well as pipelines that use marker-based or markerless kinematics as the input to learning- or optimization-based GRF predictors \cite{OpenCapMonocular2026, Katsu2024GRFMV, tan2025gaitdynamics}. In these controlled configurations, the agreement with force-plate measurements can be strong, supporting the feasibility of GRF prediction from kinematics \cite{Lichtwark2024GRF, Katsu2024GRFMV, tan2025gaitdynamics}. However, these successes have typically not been extensively validated across diverse populations for GRF estimation, and the corresponding fully markerless inverse-dynamics pipeline, with joint moments computed from predicted GRFs/CoP, has not been broadly validated beyond the datasets and conditions used for model development.

The existing approaches have several clinically important limitations that impede their translation to monocular ``in-the-wild'' gait analysis. Many GRF/CoP predictors have been developed and validated under controlled conditions, and robustness under clinically relevant domain shifts, including viewpoint, clothing, background clutter, assistive devices, atypical postures, and altered foot--ground contact mechanics, remains insufficiently characterized. More broadly, the field has not yet converged on a rigorously validated end-to-end pipeline that jointly yields fully markerless monocular kinematics together with sufficiently accurate 3D GRFs for clinical populations. In addition, evidence remains limited on how HMR model choice affects mesh fidelity in specific pathologies and how reconstructions should be standardized through virtual-marker definitions and model scaling to produce trustworthy downstream kinematic and kinetic outputs. Critically, to our knowledge, no prior work has combined such a fully monocular, fully markerless kinematics--kinetics pipeline with a comprehensive, pathology-focused biomechanical analysis in SCI, including clinically interpretable biomechanical biomarkers that stratify individuals into mechanistically distinct subgroups.

In this study, we demonstrate and validate a monocular, calibration-free markerless pipeline for comprehensive gait assessment in individuals with spinal cord injury (SCI), bridging the gap between computer-vision body recovery and clinical neurorehabilitation (\figref{fig:intro_sn}). Clinically, it is difficult to anticipate who will improve in walking speed, in part because individuals can achieve similar speeds using different coordination and compensation strategies; therefore, an actionable clinical question is whether we can identify reproducible biomechanical patterns \emph{within and between groups}, and then determine which of these patterns are plausible targets for intervention (i.e., which mechanisms to treat and which features to monitor as therapy outcomes)\cite{Hulleck2022, Werner2023}. Our work establishes three primary contributions, including a comprehensive benchmark of five state-of-the-art 3D human mesh recovery models on our pathological walking dataset, \texttt{SCAI SCI Gait} ($N = 239$), and demonstrates that parametric whole-body constraints prevent anatomically implausible limb deformations during pathological asymmetry and occlusion. We quantify the physical limits of kinematics-to-force prediction (\texttt{GaitDynamics}) in service of scalable clinical kinetics, and we apply conditional-dependence graphical models to discover minimal Markov blankets of functional recovery and employ speed-disentangled clustering to uncover latent mechanical phenotypes that are masked by conventional clinical examinations.

\section*{Results}
\subsection*{Parametric body-mesh recovery provides robust 3D lower-limb poses under neurological gait pathology}

First, we determined which monocular reconstruction strategy provides a
sufficiently reliable foundation for downstream biomechanics in SCI.
Accordingly, we benchmarked five state-of-the-art 3D-HMR models:
\texttt{CameraHMR}\cite{patel2024camerahmr}, \texttt{WHAM}\cite{shin2024wham},
\texttt{Multi-HMR}\cite{multi-hmr2024}, \texttt{W-HMR}\cite{yao2024whmr}, and
\texttt{PLIKS}\cite{Shetty_2023_CVPR}.
Except \texttt{WHAM}, which takes a video as input and performs multi-frame
reconstruction over a temporal window, all evaluated methods were run in a
frame-by-frame inference setting, estimating the \texttt{SMPL} pose/shape (and
camera parameters) independently per image (with any temporal consistency
arising implicitly or through lightweight smoothing). For each trial, we ran
every method across the entire video to obtain a temporally ordered \texttt{SMPL}
sequence, and then extracted the same lower-body joint set from each frame and
synchronized it with the corresponding marker-based motion-capture reference.

\textbf{Dataset:} The benchmarking was performed on the \texttt{SCAI SCI Gait dataset},
a clinical gait cohort of 239 adult individuals with SCI recorded during instrumented
level walking on a \textbf{10 m walkway} using a \texttt{Qualisys} motion-capture
(MoCap) system with synchronized lower-limb marker trajectories and force-plate
ground-reaction forces (GRFs) (see \tabref{tab:id_dataset}).
The cohort includes \textbf{176 males and 63 females}, with a mean age of
\textbf{46.62 $\pm$ 15.52 years} (18--83), a mean height of \textbf{1.76 $\pm$ 0.10 m}, and a
mean body weight of \textbf{75.27 $\pm$ 15.69 kg}. These synchronized video-MoCap
recordings provide per-frame reference joint trajectories, enabling direct
joint-position error scoring against motion capture. The dataset includes
recordings from two video acquisition settings (high- and low-resolution) and
comprises \textbf{49,838} video frames in total, of which \textbf{46,717}
contained valid person detections suitable for mesh reconstruction
(\textbf{36,925 high-resolution} and \textbf{9,792 low-resolution} frames).
To enable sharing while protecting participant privacy, we additionally provide
an anonymized release in which each sequence is re-rendered using the
best-performing HMR model from our benchmark, with a black background and a
generic \texttt{SMPL} head; this anonymized dataset is distributed together with
the de-identified medical metadata alongside this paper.

For each frame, every method's \texttt{SMPL}-based output was mapped to this shared
pelvis/hip/knee/ankle/toe definition to produce a comparable 9-joint 3D
trajectory, which was then scored against motion capture using \textit{PA-MPJPE}
(mm). \textit{PA-MPJPE} was computed per frame after an optimal similarity
alignment between the predicted and reference keypoints (see
\eqhref{eq:pa_mpjpe_align}--\eqhref{eq:pa_mpjpe}), reflecting the inherent global
similarity ambiguity of monocular, camera-centered \texttt{SMPL} reconstructions.
In addition to accuracy, we quantified robustness via the number of \emph{valid
frames} (frames with a complete reconstruction and all 9 target joints present;
sensitive to occlusion, blur, truncation, and assistive devices). To mitigate
trial start/end artifacts, we trimmed \textbf{5\%} of the frames at the beginning
and end of each recording before aggregation. The results are stratified by
acquisition setting, \textbf{high resolution} (1920$\times$1088 at 50 FPS) vs.
\textbf{low resolution} (720$\times$576 at 25 FPS), and summarized with an
\textit{Overall} score weighted by valid frame counts
(\tabref{tab:combined_benchmarking}).

\begin{table}[ht]
\centering
\begin{tabular}{lccc}
  \toprule \multicolumn{4}{c}{\textbf{\texttt{SCAI SCI Gait Dataset} \textit{(9 Lower Body Keypoints, PA-MPJPE)}}} \\
  \midrule \textbf{3D pose estimator}                                                                         & \textbf{1920$\times$1088 at 50 FPS} & \textbf{720$\times$576 at 25 FPS} & \textbf{Overall}                  \\
  \midrule \texttt{Multi-Person}                                                                              & 74.6 $\pm$ 11.9 (18,591)            & 112.2 $\pm$ 44.1 (210)            & 81.7 $\pm$ 26.1 (18,801)          \\
  \texttt{PLIKS}                                                                                              & 72.7 $\pm$ 14.2 (29,843)            & 130.2 $\pm$ 54.5 (1,001)          & 94.8 $\pm$ 45.1 (30,844)          \\
  \texttt{W-HMR}                                                                                              & 70.8 $\pm$ 10.9 (27,936)            & 99.3 $\pm$ 23.2 (6,152)           & 83.7 $\pm$ 22.6 (34,088)          \\
  \texttt{WHAM}                                                                                               & 71.1 $\pm$ 13.8 (29,186)            & 99.6 $\pm$ 22.0 (7,396)           & 83.5 $\pm$ 22.7 (36,582)          \\
  \texttt{\textbf{CameraHMR}}                                                                                 & \textbf{70.6 $\pm$ 15.2 (31,083)}   & \textbf{97.9 $\pm$ 25.5 (10,374)} & \textbf{83.0 $\pm$ 24.6 (41,457)} \\
  \bottomrule
\end{tabular}
\caption{\small{\textbf{HMR benchmarking on the \texttt{SCAI SCI Gait dataset} (9 lower body keypoints).} Five monocular \texttt{SMPL}-based methods (\texttt{CameraHMR}\cite{patel2024camerahmr}, \texttt{WHAM}\cite{shin2024wham}, \texttt{Multi-HMR}\cite{multi-hmr2024}, \texttt{W-HMR}\cite{yao2024whmr}, and \texttt{PLIKS}\cite{Shetty_2023_CVPR}) are evaluated on the pelvis/root and bilateral hip, knee, ankle, and toe joints (9 total). The accuracy is \textit{PA-MPJPE} (mm) computed per valid frame after Procrustes similarity alignment (rotation, translation, and scale) to motion capture; entries represent the mean $\pm$ SD, with valid-frame counts in parentheses after trimming 5\% at the start/end of each recording. Valid frames require complete reconstruction with all 9 joints (robustness to occlusion, blur, truncation, and assistive devices). The results are shown for high-resolution (1920$\times$1088 at 50 FPS), low-resolution (720$\times$576 at 25 FPS), and overall aggregated across resolutions weighted by valid-frame counts.}}
\label{tab:combined_benchmarking}
\end{table}

\textbf{Overall accuracy--robustness trade-off, model selection, and relation to pose-only pipelines:} Because \textit{PA-MPJPE} is computed over each method's own set of valid frames, differences in coverage can confound a pure accuracy ranking; therefore, we interpret \autoref{tab:combined_benchmarking} as an accuracy--robustness trade-off.

With respect to all the valid frames, \texttt{CameraHMR} achieved the highest coverage (41,457 valid frames), with the second-lowest overall \textit{PA-MPJPE} (83.0$\pm$24.6~mm), whereas \texttt{Multi-Person} obtained the lowest overall error (81.7$\pm$26.1~mm) but with very limited coverage (18,801 valid frames). The accuracy and coverage of \texttt{CameraHMR} were consistent across acquisition settings: on high-resolution videos, it reached 70.6$\pm$15.2~mm over 31,083 valid frames, and on low-resolution videos, it retained 97.9$\pm$25.5~mm over 10,374 valid frames, indicating comparatively graceful degradation under reduced image quality. Although \texttt{W-HMR} and \texttt{WHAM} were similarly accurate on high-resolution videos, both returned fewer valid frames overall (particularly under low resolution), and \texttt{WHAM} did not match \texttt{CameraHMR} in terms of either total coverage or low-resolution accuracy. \texttt{PLIKS} exhibited the strongest degradation at low resolution (highest error and few valid frames). Because missing frames and greater joint error propagate directly to gaps and errors in virtual-marker trajectories, destabilizing kinematics, we selected \texttt{CameraHMR} as the default HMR method for subsequent musculoskeletal analyses.

Furthermore, the \texttt{SMPL}-constrained mesh recovery procedure reduced aberrant limb configurations under partial visibility by coupling joint predictions through a shared parametric whole-body model. This joint coupling improved the consistency of the relative pose geometry, as reflected by lower \textit{PA-MPJPE}, even in the presence of occlusions and degraded image resolution (see \sfigref{fig:supp_pampjpe_resolution}). This behavior contrasts with that of less constrained 3D pose-reconstruction pipelines, in which joints are estimated in a more independent manner and can thus yield implausible limb lengths and joint angles, particularly when keypoints are missing or contaminated by noise. Consequently, on the \texttt{SCAI SCI Gait dataset}, the mesh-based approach produced more accurate joint keypoint estimates than the pose-only 3D reconstruction methods did\cite{Natraj2025npjDM} because the mesh constraints regularize the solution space toward anatomically plausible limb lengths and joint kinematics. This higher-fidelity keypoint estimation, in turn, enhanced the robustness and validity of subsequent kinematic analyses and other downstream computational tasks.

\subsection*{SMPL-derived virtual markers drive accurate OpenSim inverse kinematics across a clinical SCI cohort}

\begin{figure}[H]
\centering

\begin{subfigure}
  [t]{0.8\linewidth}
  \centering
  \includegraphics[width=\linewidth]{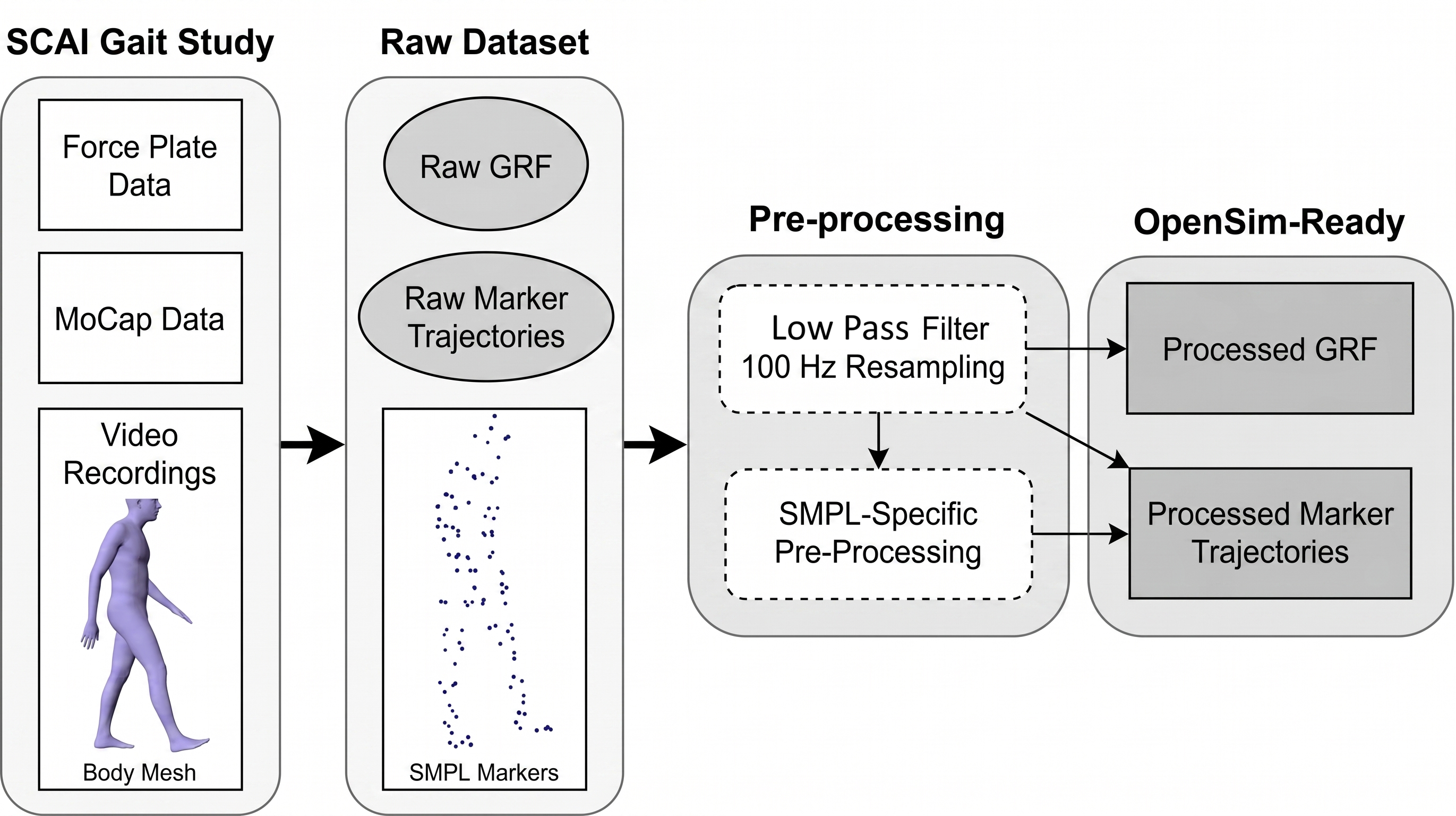}
  \caption{\small{\textbf{(a) Data preparation and kinetics overview.} Raw \texttt{SCAI SCI Gait} dataset inputs are converted to \texttt{OpenSim}-ready data (\texttt{SMPL} virtual markers, filtering, and resampling), enabling inverse dynamics with measured GRFs or predicted GRFs (pGRFs via \texttt{GaitDynamics}).}}
  \label{fig:pipeline_overview}
\end{subfigure}

\par
\medskip

\begin{subfigure}
  [t]{0.54\linewidth}
  \centering
  \includegraphics[width=\linewidth]{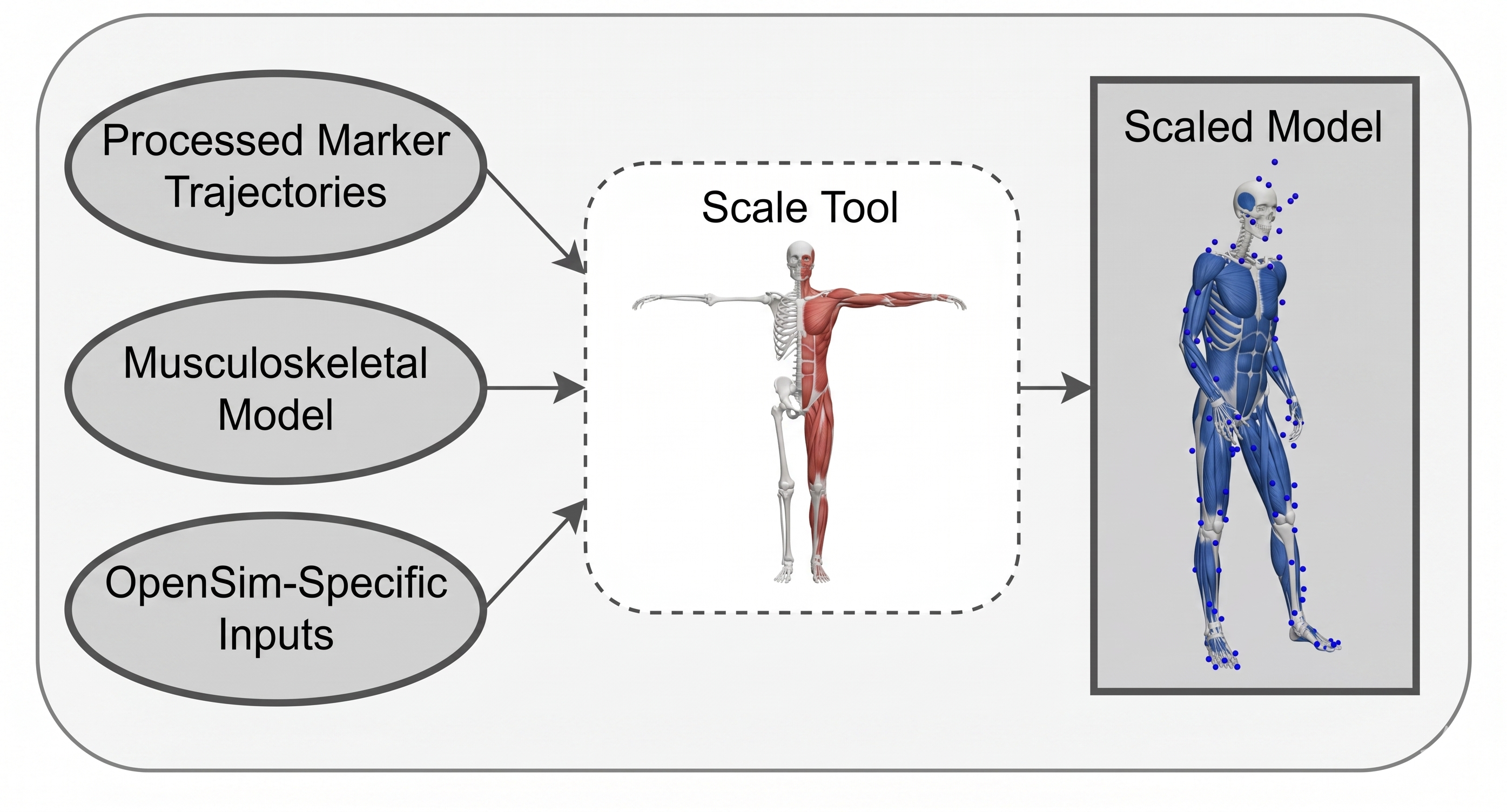}
  \caption{\small{\textbf{Subject-specific model scaling.} The \texttt{OpenSim} model is scaled using either physical MoCap markers or \texttt{SMPL}-derived virtual markers with consistent marker pairs and segment scale factors.}}
  \label{fig:scaling}
\end{subfigure}
\hfill
\begin{subfigure}
  [t]{0.45\linewidth}
  \centering
  \includegraphics[width=\linewidth]{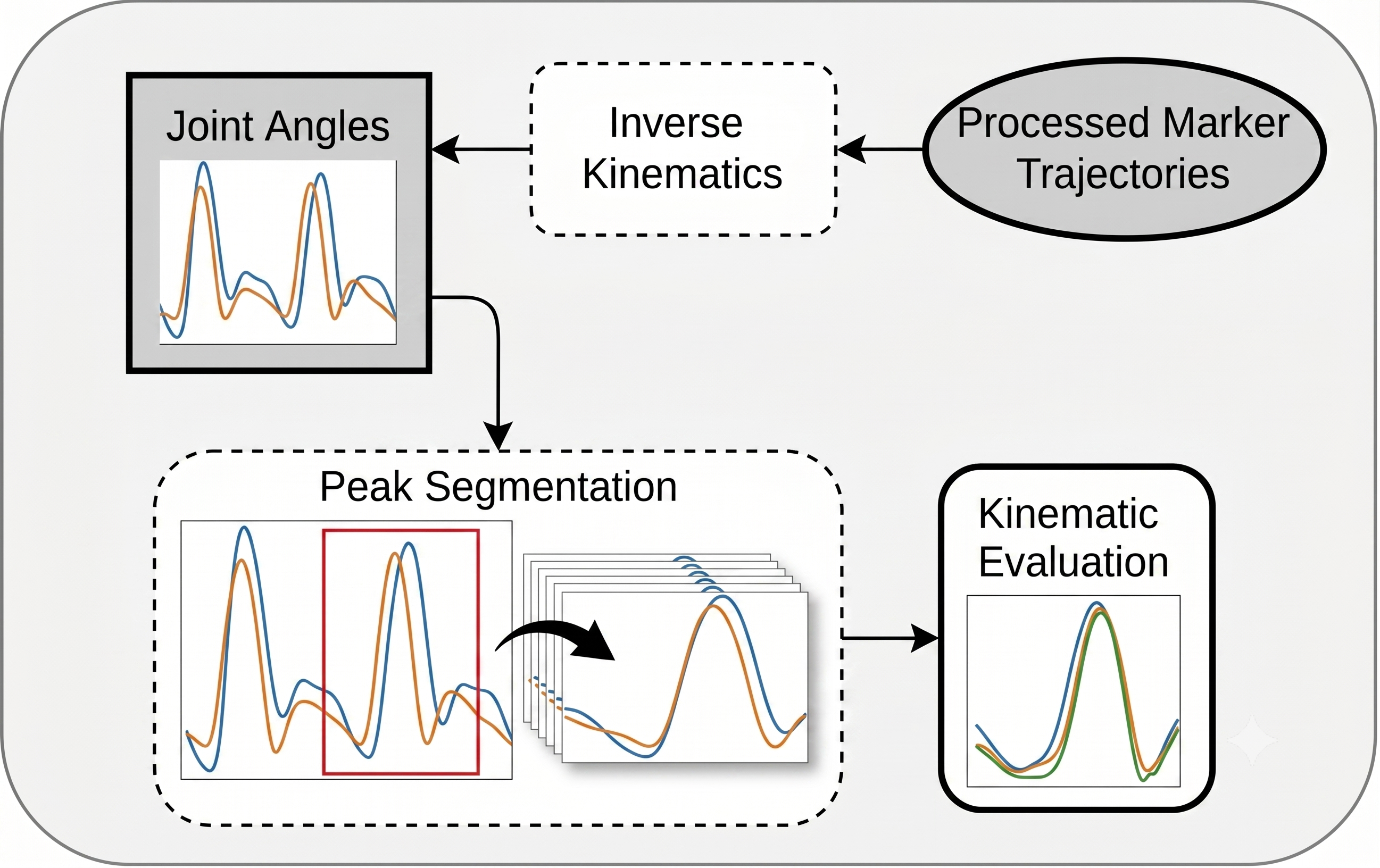}
  \caption{\small{\textbf{Kinematics workflow.} Filtered/resampled trajectories drive \texttt{OpenSim} IK; outputs are trimmed, segmented into gait cycles, time-normalized, temporally aligned, and evaluated with error/shape metrics and SPM.}}
  \label{fig:kinematic_pipeline}
\end{subfigure}

\caption{\small{Figure 2. \textbf{SMPL-to-OpenSim workflow for virtual-marker musculoskeletal analysis.} (a) Subject-specific scaling with physical or \texttt{SMPL}-derived virtual markers. (b) Kinematic processing and evaluation from IK to gait cycle comparison. (c) Data preprocessing and kinetic analysis pipeline, including inverse dynamics with measured or predicted GRFs.}}
\label{fig:osim_pipeline_overview}
\end{figure}

Following the reconstruction of temporally ordered \texttt{SMPL} body meshes from sagittal monocular videos using \texttt{CameraHMR}\cite{patel2024camerahmr}, \texttt{OpenSim}-ready virtual marker trajectories were generated by sampling an anatomically defined set of surface points on the \texttt{SMPL} template (\figref{fig:pipeline_overview}).

\textbf{Scaling:} Defining markers in template space enforces one-to-one anatomical correspondence across time. For each reconstructed frame, the 3D coordinates of the selected \texttt{SMPL} vertices/landmarks were exported to form per-marker trajectories that are directly analogous to optical motion capture (MoCap). Across these 20-marker lower-limb physical marker trajectories acquired using the optical MoCap configuration (\sfigref{fig:mocap_markerset}), 18 markers were retained for both the \texttt{OpenSim} Scale and IK. Virtual markers were generated from \texttt{SMPL} reconstructions by sampling anatomically defined template vertices, yielding 105 virtual markers for the \texttt{Pose2Sim} configuration and 77 virtual markers for the Rajagopal configuration (\sfigref{fig:smpl_grid_full_p2s}, \sfigref{fig:smpl_grid_parts}, \sfigref{fig:smpl_grid_full_raja}).

The trajectories were then smoothed to suppress frame-to-frame jitter, resampled when needed to
reconcile the sampling rates, and verified for consistent units and coordinate conventions prior to use
in musculoskeletal modeling. To preserve a controlled, like-for-like comparison, we applied identical
segment-length definitions and scaling rules across modalities, including the same bilateral
marker-pair distance definitions for pelvis/hip, femur, and tibia scaling and fixed multiplicative
scale factors where direct marker-pair constraints were not used
(\stabref{tab:scaling_marker_settings}; \stabref{tab:markers_tools}). Consistent with the
standard \texttt{OpenSim} Scale workflow (\figref{fig:scaling}), segment
scale factors were estimated by minimizing a weighted marker misfit in the scaling pose while
updating segment inertial properties from the scaled geometry; weights were chosen to emphasize robust
bony landmarks (typically \textbf{weight=10}) over soft-tissue/skin markers (\textbf{weight=1--5}) and,
for \texttt{SMPL}-derived inputs, to stabilize global alignment via moderate upweighting of
pelvis/shoulder markers (\textbf{weight=2}) while downweighting less reliable adipose-region markers
(\textbf{weight=0.1}).

\textbf{Inverse Kinematics:} Joint kinematics were estimated with \texttt{OpenSim} inverse kinematics
(IK) using the scaled model and time-varying marker trajectories. Within each model configuration,
the marker definitions, complete marker-weight map, and solver settings were fixed across marker-based
and markerless runs; thus, the only difference was the input trajectory source. IK solutions were
exported as continuous joint-angle time series, trimmed (first/last 10\%) to remove start/stop
transients, segmented into individual gait cycles, and evaluated cycle by cycle to avoid conflating
intertrial variability with waveform agreement. To focus comparisons on the within-cycle kinematic
signature (rather than absolute offsets), each cycle was standardized (z-scored) prior to computing
the similarity/error metrics.

Residual phase differences (from sampling-rate mismatch, upstream filtering/smoothing, and small
gait-event timing errors) were then corrected via dynamic time warping (DTW) alignment
(\eqhref{dtw_eqn}) so that pointwise comparisons reflected shape differences
rather than trivial time shifts. Agreement was quantified using complementary metric families:
(i) Pearson's correlation ($r$) for temporal structure (\eqhref{pearson_eqn});
(ii) root mean square error/mean absolute error for amplitude disagreement in degrees
(\eqhref{rmse_eqn}--\eqhref{mae_eqn}); and
(iii) distance-based measures (MSM and DTW costs) for residual waveform/warping distance. Finally, we
assessed time-localized waveform differences using \texttt{SPM1D} throughout the gait cycle
(\eqhref{spm_t_eqn}). Because the large number of matched gait cycles
increases statistical sensitivity (even small deviations can reach significance), we also report
\texttt{Cohen's $d$} effect sizes to quantify the magnitude of differences independent of $p$-values
(\eqhref{cohend_eqn}).

\begin{figure}[H]
\centering

\begin{subfigure}
  [t]{0.98\linewidth}
  \centering
  {\scriptsize \setlength{\tabcolsep}{3pt} \renewcommand{\arraystretch}{1.15} \resizebox{\linewidth}{!}{%
  \begin{tabular}{llcccccccccccc}\toprule \multicolumn{14}{c}{\texttt{\textbf{Pose2Sim: Marker-based vs Markerless SCI Flexion Angles (Mean-Centered)}}} \\ \midrule \textbf{Model} & \textbf{Joint} & \textbf{Side} & \textbf{Cycles} & \textbf{Static offset ($^{\circ}$)} & \textbf{$r$} & \textbf{$p$} & \textbf{RMSE ($^{\circ}$)} & \textbf{MAE ($^{\circ}$)} & \textbf{MSM} & \textbf{Norm. MSM} & \textbf{DTW} & \textbf{DTW $z$} & \textbf{Warp eff.} \\ \midrule \multirow{9}{*}{\texttt{Pose2Sim}} & Hip flexion & Right & 286 & $\phantom{-}1.94^{\circ}\,\pm\,5.58^{\circ}$ & 0.8739 & 0.0027 & 4.75 & 3.82 & 114.8116 & 1.1367 & 29.8309 & 2.3416 & 0.0493 \\ & Hip flexion & Left & 312 & $\phantom{-}0.50^{\circ}\,\pm\,5.55^{\circ}$ & 0.9023 & 0.0001 & 4.18 & 3.35 & 112.5776 & 1.1146 & 26.6929 & 2.0927 & 0.0480 \\ & Knee angle & Right & 304 & $\phantom{-}1.28^{\circ}\,\pm\,5.62^{\circ}$ & 0.8667 & 0.0122 & 6.49 & 5.29 & 135.9759 & 1.3463 & 40.2701 & 2.3099 & 0.0489 \\ & Knee angle & Left & 297 & $\phantom{-}0.36^{\circ}\,\pm\,7.40^{\circ}$ & 0.8696 & 0.0001 & 6.44 & 5.26 & 129.9602 & 1.2867 & 38.2411 & 2.1573 & 0.0466 \\ & Ankle angle & Right & 319 & $-15.45^{\circ}\,\pm\,5.70^{\circ}$ & 0.6825 & 0.0324 & 5.20 & 4.17 & 135.2198 & 1.3388 & 37.0781 & 3.9994 & 0.0601 \\ & Ankle angle & Left & 311 & $-14.58^{\circ}\,\pm\,5.34^{\circ}$ & 0.7492 & 0.0126 & 4.22 & 3.34 & 123.6991 & 1.2247 & 29.9821 & 3.5259 & 0.0526 \\ \cmidrule(lr){2-14} & \multirow{3}{*}{\textit{Overall}} & Right & 909 & $-4.08^{\circ}\,\pm\,5.63^{\circ}$ & 0.8077 & 0.0158 & 5.48 & 4.43 & 128.6691 & 1.2739 & 35.7264 & 2.8836 & 0.0528 \\ & & Left & 920 & $-4.57^{\circ}\,\pm\,6.10^{\circ}$ & 0.8404 & 0.0043 & 4.95 & 3.98 & 122.0790 & 1.2087 & 31.6387 & 2.5920 & 0.0491 \\ & & All & 1829 & $-4.33^{\circ}\,\pm\,5.87^{\circ}$ & 0.8241 & 0.0100 & 5.22 & 4.21 & 125.3740 & 1.2413 & 33.6826 & 2.7378 & 0.0509 \\ \bottomrule\end{tabular}%
  } }
  \caption{\small{\textbf{\texttt{Pose2Sim} model (mean-centered): segment-level metrics.} Segment-level agreement metrics (mean-centered; DTW-aligned) for \texttt{Pose2Sim} \texttt{SMPL} IK vs marker-based MoCap IK; complemented by SPM and Cohen's $d$ over the gait cycle (SPM: Hip flexion--Right significant at 34.9--35.6\% ($p=0.049$, small $d$), 36.1--40.0\% ($p=0.028$, small $d$), and 76.9--88.6\% ($p<.001$, small $d$); all other joints/sides: none).}}
  \label{tab:ik_metrics_models}
\end{subfigure}

\vspace{1.25em}

\begin{subfigure}
  [t]{0.98\linewidth}
  \centering
  \includegraphics[width=1\linewidth]{
    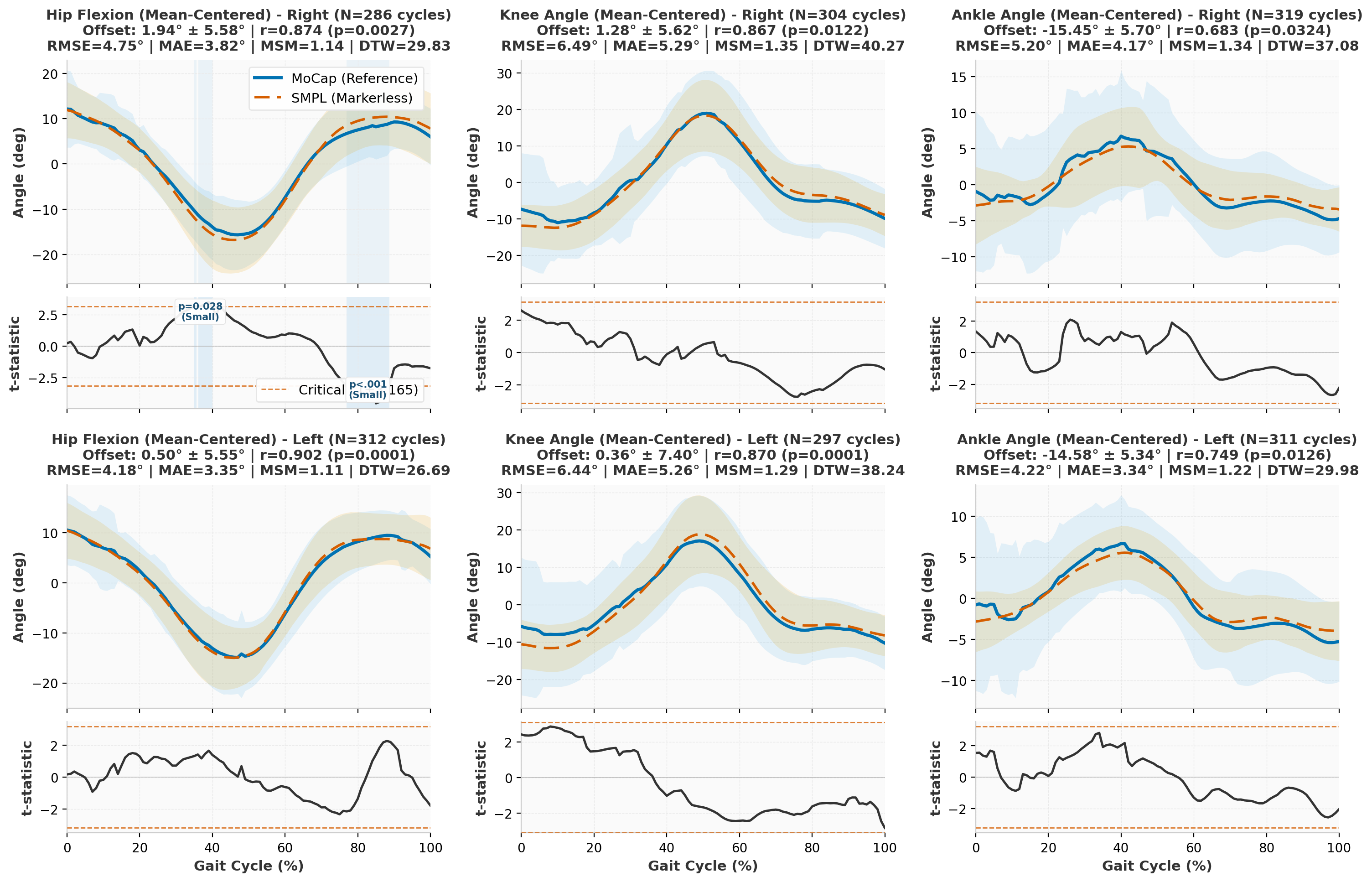
  }
  \caption{\small{\textbf{Pose2Sim model (mean-centered): ensemble-averaged waveforms.} DTW-aligned ensemble-averaged flexion waveforms (markerless vs.~marker-based), with time-continuous SPM significance testing and Cohen's $d$ effect sizes.}}
  \label{fig:ik_p2s_panel}
\end{subfigure}

\caption{\small{\textbf{Figure 3. Markerless IK agreement with marker-based motion capture (Pose2Sim model).} Markerless kinematics: monocular-video \texttt{SMPL} meshes $\rightarrow$ \texttt{SMPL} virtual markers $\rightarrow$ IK in \texttt{OpenSim 4.5}. Marker-based kinematics: physical MoCap markers processed with the same musculoskeletal model and joint definitions. The recordings were trimmed (first/last 10\%) and segmented into gait cycles using extrema-based events (hip minima; knee/ankle maxima), and only matched cycles were retained (\textit{Cycles}, panel~a). Cycles were time-normalized, mean-centered, and DTW-aligned. Panel~\textbf{(a)} reports similarity/error metrics (Pearson $r$/$p$; RMSE/MAE; offset mean $\pm$ SD; MSM/nMSM; DTW distance/$z$/effort). Panel~\textbf{(b)} shows DTW-aligned ensemble-averaged sagittal flexion waveforms (bilateral hip/knee/ankle). SPM with Cohen's $d$ identifies significant gait-cycle regions and effect magnitude.}}
\label{fig:ik_p2s_centred}
\end{figure}

Across \textbf{1,829 matched gait cycles}, the markerless pipeline closely reproduced the
sagittal-plane flexion waveforms (\figref{fig:ik_p2s_centred}).
For mean-centered angles (to deemphasize baseline offsets), the correlations for hip flexion were
$r = 0.874$--$0.902$, with an RMSE of $4.18$--$4.75^{\circ}$, and the correlations for knee
flexion/extension were $r = 0.867$--$0.870$, with an RMSE of $6.44$--$6.49^{\circ}$
(\tabref{tab:ik_metrics_models}). The ankle correlations were
lower ($r = 0.683$--$0.749$), but the mean-centered ankle RMSE ($4.22$--$5.20^{\circ}$) remained small,
indicating preserved within-cycle excursion with greater sensitivity to baseline and/or distal
constraint consistency.

These findings are broadly consistent with prior markerless gait studies that reported stronger
agreement for proximal sagittal-plane kinematics than for ankle motion and non-sagittal degrees of
freedom \cite{KANKO2021110665, Riazi2022, TURNER2024112200, HORSAK2025112986}. Consistent with this,
the DTW warping effort was uniformly low (\textbf{Warp eff.} $\approx 0.05$;
\tabref{tab:ik_metrics_models}), implying close phase matching
after segmentation with only minor temporal adjustment needed.

\textbf{Participant-level aggregation and SPM1D independence:} Because each participant contributed
multiple gait cycles, we additionally computed participant-level summaries (participant medians with
IQRs) and used cluster-robust (CR1) 95\% confidence intervals with the participant as the clustering
unit ($G = 92$). Participant-level median correlations remained high for proximal joints (hip:
$r = 0.911$ [0.636, 0.965]; knee: $r = 0.865$ [0.610, 0.945]) but were lower at the ankle
($r = 0.575$ [0.232, 0.715]), consistent with the proximal--distal trend. To satisfy the Random Field
Theory assumption of independent observations in time-continuous inference, we reran the 1D SPM paired
$t$-tests on participant-averaged trajectories ($N = 92$ independent curves; $df = 91$). Participant
averaging removed spurious suprathreshold regions caused by inflated degrees of freedom: the ankle
suprathreshold differences decreased from 11.9\%--18.8\% (pooled cycles) to 0.0\%--1.0\%
(participant-averaged), and the hip differences decreased to 0.0\%, indicating that the \texttt{Pose2Sim}
and MoCap waveforms are statistically indistinguishable across nearly the entire gait cycle.

\textbf{Ankle static offset stability ($s_{\text{within}}$):} In raw (non-mean-centered) angles, the
dominant discrepancy was a distal static offset concentrated at the ankle (approximately
$-14.6$ to $-15.5^{\circ}$; \tabref{tab:ik_metrics_models}), which
inflated the raw ankle errors (RMSE $16.07$--$17.61^{\circ}$) and increased the overall raw RMSE to
$10.58^{\circ}$. To test whether this offset reflects trial-to-trial instability versus a
participant-specific calibration artifact, we quantified the per-participant offset standard
deviation across independent trials. Across 92 participants, the median within-participant SD was
$1.82^{\circ}$ [IQR: $0.88^{\circ}$, $2.33^{\circ}$]. One-way random effects ANOVA revealed that the
between-participant variance was $> 2.6\times$ greater than the within-participant variance
($F(90,228) = 9.54$, $p < 10^{-15}$), yielding $\mathrm{ICC}(1,1) = 0.710$. Taken together, these
results indicate that the primary residual limitation is a participant-specific ankle
baseline/coordinate-frame offset (driven by anatomical and foot-segment coordinate definitions)
rather than trial-to-trial reconstruction error or failure of sagittal-plane waveform recovery.

\textbf{Musculoskeletal Model Sensitivity Analysis:} To ensure that the observed agreement patterns
were not artifacts of a single musculoskeletal model configuration, we reran the full pipeline with
the \texttt{Rajagopal} generic lower-body gait model \cite{Rajagopal2016}, which differs from
\texttt{Pose2Sim} in terms of segment definitions, joint coordinate conventions, and marker
redundancy/constraint topology (\stabref{tab:ik_metrics_raja_centered};
\sfigref{fig:ik_raja_panel_centered}).

Although \texttt{Pose2Sim} and \texttt{Rajagopal} yielded broadly similar waveform correlations and mean-centered error magnitudes, \texttt{Pose2Sim} was preferred as the primary reporting model because it was more robust and downstream-ready in this markerless setting: it retained a substantially larger number of usable gait cycles (1,829 vs.~1,582), indicating fewer IK failures and fewer segments discarded during matching/segmentation (improving the statistical power and stability of the subject-level summaries), provided a slightly higher overall waveform correlation on the aggregated kinematic set ($r = 0.8241$ vs.~0.8164 on mean-centered signals) with a comparable mean-centered RMSE/MAE (Pose2Sim: 5.22$^{\circ}$/4.21$^{\circ}$; Rajagopal: 5.16$^{\circ}$/4.18$^{\circ}$). These factors made Pose2Sim the most reliable default for the main markerless kinematic analyses, with Rajagopal serving as a principled sensitivity check that supported the same proximal--distal conclusion (hip/knee $\gg$ ankle).

\subsection*{Markerless kinematics enable robust estimation of the resulting ground-reaction force envelopes}

\textbf{Kinematics to GRF Prediction:} Following the markerless inverse kinematics (IK) determination, we evaluated whether these markerless kinematics can support a fully markerless estimate of external loading by using \texttt{GaitDynamics} to infer predicted ground-reaction forces (pGRFs) from the markerless IK solutions and compare them against synchronized instrumented force-plate measurements. Force-plate GRFs from the \texttt{Qualisys} system in the \texttt{SCAI SCI Gait} dataset recordings served as the reference. The analysis included 116 matched subjects and used step-level gait-event detection/segmentation, yielding 232 gait cycles (left/right steps) for paired comparison.

\begin{figure}[H]
\centering

\begin{subfigure}
  [t]{0.98\linewidth}
  \centering
  \resizebox{\linewidth}{!}{%
  \begin{tabular}{l l r c c r r r r r r r r}
    \toprule \textbf{Model}                     & \textbf{Side} & \textbf{Cycles} & \textbf{Static offset (N/kg)} & \textbf{$r$} & \textbf{$p$} & \textbf{RMSE (N/kg)} & \textbf{MAE (N/kg)} & \textbf{MSM} & \textbf{Norm. MSM} & \textbf{DTW} & \textbf{DTW $z$} & \textbf{Warp eff.} \\
    \midrule \multirow{3}{*}{\texttt{Pose2Sim}} & Left          & 116             & $-0.08 \pm 0.99$              & 0.8538       & $<0.001$     & 2.19                 & 1.55                & 90.7792      & 0.8988             & 12.8971      & 2.7521           & 0.0926             \\
                                                & Right         & 116             & $-0.24 \pm 0.93$              & 0.8662       & $<0.001$     & 2.13                 & 1.48                & 88.7788      & 0.8790             & 12.0828      & 2.5818           & 0.0888             \\
                                                & All           & 232             & $-0.16 \pm 0.96$              & 0.8600       & $<0.001$     & 2.16                 & 1.52                & 89.7790      & 0.8889             & 12.4899      & 2.6670           & 0.0907             \\
    \bottomrule
  \end{tabular}%
  }
  \caption{\small{\textbf{Predicted GRFs (Pose2Sim model): step-level agreement metrics.} Agreement between the predicted GRF magnitude (pGRFs; \texttt{GaitDynamics}) and force-plate GRF magnitude, computed on trimmed, bodyweight-normalized gait cycles after DTW alignment of z-scored signals.}}
  \label{tab:pgrf_results}
\end{subfigure}

\vspace{1.0em}

\begin{subfigure}
  [t]{0.98\linewidth}
  \centering
  \includegraphics[width=0.95\linewidth]{
    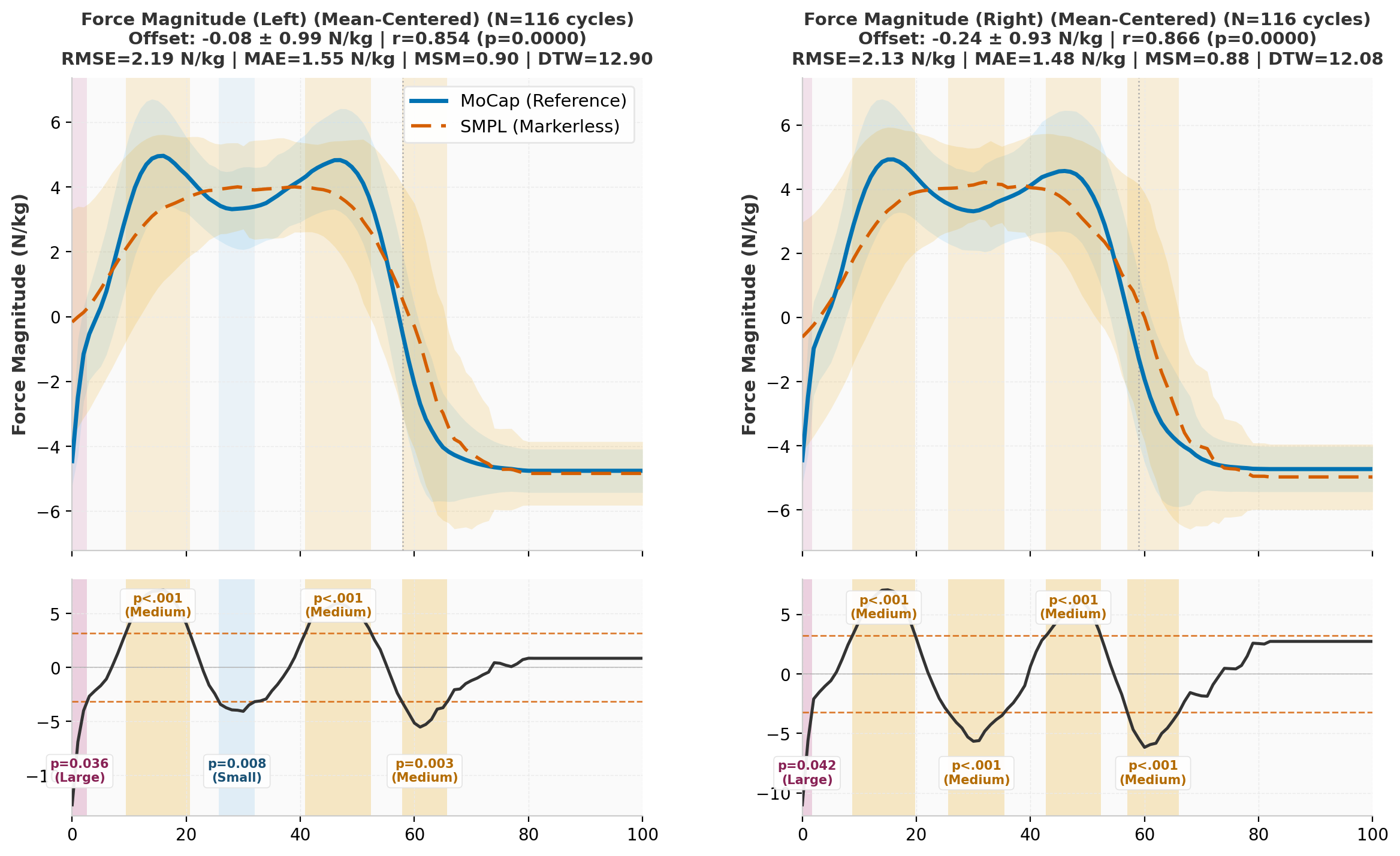
  }
  \caption{\small{\textbf{Averaged Predicted GRFs (Pose2Sim model):} The figure shows the averaged force magnitude of the predicted GRFs (pGRFs; \texttt{GaitDynamics}) against the recorded force-plate GRFs for the \texttt{Pose2Sim} model. Forces are normalized by body weight and trimmed to the recorded GRF duration. The results are reported after DTW alignment of the z-scored pGRF segments to the corresponding z-scored GRF segments. The lower panel reports the time-continuous SPM inference with Cohen's $d$ effect size bands (blue/orange/red denote small/medium/large effects, respectively, as defined in \eqhref{cohend_eqn}). Significance levels: $^{*}\,p<0.05$, $^{**}\,p<0.01$, $^{***}\,p<0.001$.}}
  \label{fig:pgrf_p2s}
\end{subfigure}
\end{figure}

Both the \texttt{Pose2Sim} and \texttt{Rajagopal} model variants were scaled to the participant and, together with the corresponding IK kinematics, provided to \texttt{GaitDynamics}, ensuring that pGRF performance reflects the complete markerless chain rather than a generic template model. Before evaluation, the measured GRFs and pGRFs were normalized by body weight, cropped to each segmented step/stance interval, and reduced to the resultant GRF magnitude (from longitudinal, vertical, and transverse components) to provide a consistent scalar waveform for comparison and to minimize confounding by residual phase differences. The agreement between the pGRFs and force-plate GRFs was quantified using Pearson's correlation (\eqhref{pearson_eqn}), the RMSE/MAE (\eqhref{rmse_eqn}--\eqhref{mae_eqn}), the static offset (\eqhref{offset_eqn}), the DTW cost, the warping efficiency (\eqhref{dtwz_eqn}--\eqhref{warpeff_eqn}), and the shape-sensitive MSM distances (\eqhref{msm_eqn}--\eqhref{msm_norm_eqn}).

Across 232 gait cycles, the \texttt{Pose2Sim}-based pGRF magnitude exhibited a strong step-level association with the force-plate reference. In the mean-centered (main) analysis, the Pearson correlation was high for both limbs (left: $r=0.8538$; right: $r=0.8662$; all: $r=0.8600$; $p<0.001$ for all comparisons), with small static offsets ($0.08\pm0.99$ to $0.24\pm0.93\,\mathrm{N/kg}$; overall $0.16\pm0.96\,\mathrm{N/kg}$), an RMSE of $2.13$--$2.19\,\mathrm{N/kg}$ (all: $2.16\,\mathrm{N/kg}$), an MAE of $1.52\,\mathrm{N/kg}$, and strong normalized MSM agreement ($0.879$--$0.899$; all: $0.8889$; \tabref{tab:pgrf_results}).

In the raw waveform comparison, the correlation was essentially unchanged (left: $r=0.8538$; right: $r=0.8662$), but the amplitude/shape agreement decreased: the RMSE increased to $2.32$--$2.37\,\mathrm{N/kg}$, and the normalized MSM decreased to $0.711$--$0.715$.

Consistent with this, the amplitude-level errors remained meaningful despite minimal systematic bias (no consistent over- or underestimation), reflecting a nontrivial average pointwise mismatch (overall RMSE/MAE above). Temporal misalignment was moderate (overall DTW $=12.49$; DTW $z=2.67$), and the warping efficiency was consistently high ($\eta_{\mathrm{warp}}\approx0.09$; \tabref{tab:pgrf_results}), suggesting that large phase shifts are not the dominant failure mode; rather, residual error is better explained by local waveform shape and amplitude differences that persist even after DTW.

These differences are evident in the ensemble averages (\figref{fig:pgrf_p2s}): the predicted waveform is typically smoother than the force plate signal, with attenuated higher frequency features (impact-related transients and sharper secondary peaks). This pattern indicates that the markerless chain preserves the gross stance-loading pattern (high $r$ for both sides), whereas baseline offsets, global scaling, and residual local shape differences contribute materially to the mismatch in the raw waveform (large raw vs. mean-centered changes in MSM, with only modest changes in DTW/warping metrics). This smoothing is consequential for inverse dynamics because joint moments depend on the within-stance distribution of the external load and its interaction with the segment kinematics.

Overall, Pose2Sim-based pGRFs appear sufficiently consistent to capture between-step and between-side variation in overall loading (high $r$, low bias), making it a reasonable proxy for analyses tolerant of modest amplitude/shape error (gross loading trends, step-to-step comparisons). However, the remaining RMSE/MSM/DTW costs and the observed smoothing suggest that fully markerless inverse dynamics will likely be limited by the loss of high-frequency and peak-detail information in the external-load estimate, motivating our separation of semimarkerless (measured GRFs) and fully markerless (predicted GRFs) kinetic evaluations.

\textbf{Musculoskeletal Model Sensitivity Analysis:} Similar to the kinematic sensitivity analysis, a complementary Rajagopal-based evaluation (mean-centered waveforms) was carried out (see \stabref{tab:pgrf_results_raja_centred}--\sfigref{fig:pgrf_raja_centred}). For the mean-centered condition, the correlations were again strong (left: $r=0.8531$; right: $r=0.8591$; both $p<0.001$), with small static offsets ($0.04\pm0.95$ and $0.10\pm0.88\,\mathrm{N/kg}$), an RMSE of $2.10$--$2.16\,\mathrm{N/kg}$, and a normalized MSM of $0.855$--$0.888$ (\stabref{tab:pgrf_results_raja_centred}). For the raw waveforms, the Pearson correlation remained similar (left: $r=0.8531$; right: $r=0.8591$), but the RMSE increased to $2.27$--$2.33\,\mathrm{N/kg}$, and the normalized MSM decreased to $0.689$--$0.708$. The Rajagopal trends mirrored those of Pose2Sim, where mean centering primarily improved the baseline/scale-dependent shape agreement (MSM) without materially changing the correlation/DTW. This sensitivity check indicated that the pGRF inference is not strongly model-specific within the range of realistic participant-scaled OpenSim parameterizations tested here, strengthening confidence that the reported fully markerless kinetics trends are not an artifact of a single model choice.

A comparison of the models directly revealed that Pose2Sim achieved a slightly greater Pearson association overall ($r=0.8600$ for Pose2Sim vs. $r=0.8531$--$0.8591$ for Rajagopal) with a similarly small bias and similar DTW/warping efficiency, which is consistent with Pose2Sim producing kinematics that better match the force-inference mapping in GaitDynamics (less kinematic noise/soft-tissue artifact in the virtual-marker IK solution and/or closer alignment of segment/joint conventions in the end-to-end pipeline), yielding a more faithful timing/shape coupling between stance kinematics and stance loading.

\subsubsection*{Recursive inverse dynamics reveal failure modes in proximal joint moments under pathological gait}

Inverse dynamics (ID; Nm/kg) were computed in OpenSim 4.5 using the Pose2Sim and Rajagopal models to quantify how markerless kinematics affect net joint moments. In the semimarkerless workflow, SMPL-driven kinematics were aligned to the measured center of pressure (CoP), segmented into gait cycles, and paired with force-plate GRFs across 116 matched trials (232 gait cycles), which is in agreement with the marker-based (MoCap) reference degraded from distal to proximal (\sfigref{fig:id_p2s}--\sfigref{figure:id_raja}; \stabref{tab:semi_markerless_id}), with strong ankle and knee correlations (knee $r\approx0.73$--$0.74$; ankle $r\approx0.84$--$0.87$, all significant) but weaker hip correlations ($r\approx0.51$--$0.58$) and larger waveform errors. This finding indicates that for the SCI gait patterns in this cohort, the OpenSim ID pipeline did not reliably recover whole-limb dynamics even when instrumented force-plate data were available: ankle moments were relatively robust, but errors increased systematically as the analysis moved proximally. Fully markerless kinetics then replaced the force-plate GRFs with the predicted GRFs (pGRFs) using the same ID procedure; 114 of 116 trials produced valid pairs (228 gait cycles), and the agreement decreased at all the joints (\sfigref{fig:pid_p2s}--\sfigref{figure:pid_raja}; \stabref{tab:semi_markerless_id}), yielding low-to-moderate ankle/knee correlations ($r\approx0.38$--$0.53$) and poor hip correlations ($r\approx0.17$--$0.29$).

Beyond pGRF quality, these results suggest additional method-driven error sources under pathologic gait, including potential model--data mismatch in the generic musculoskeletal models (joint-center definitions, segment inertial properties after scaling, or foot--ground contact assumptions) and/or limitations of the OpenSim inverse-dynamics/inverse-solver settings (filtering, frame definitions, or how residuals are handled). Thus, even with measured GRFs, small kinematic and GRF-application inconsistencies can accumulate through the recursive chain; thus, the hip and other proximal ID moments were deemed insufficiently reliable for subsequent conditional-dependence graph analysis because they could reflect method-driven variance rather than clinically meaningful biomechanics.

\subsection*{Graphical models isolate clinical endpoints underlying specific motor impairment phenotypes}

Joint moments were excluded from the conditional-dependence graph because, as shown above, proximal moments were highly sensitive to small inconsistencies between kinematics and external loads. Retaining them would risk recovering dependencies driven by unstable inverse dynamics rather than by clinically meaningful gait structure. The graph feature set comprised eleven markerless waveforms: bilateral hip-flexion, knee-flexion, and ankle-dorsiflexion angles; three pelvic rotations (tilt, list, and rotation); and two vertical ground-reaction-force traces, one per force plate. These eleven signals were consistently available across the cohort and showed the most reliable agreement with their laboratory references.

Graph analysis was performed on the subset of the \texttt{SCAI SCI Gait} dataset with both usable markerless waveforms and complete clinical records: 105 participants contributed 328 peak-anchored gait segments (median 3 segments per participant, IQR 2--4, range 1--8). Because segments from the same participant are not independent, all bootstrap resampling was performed at the participant level rather than at the segment level (see \tabref{tab:graph_cohort}). Two features of that distribution bound the interpretation of the graphs and should be read alongside them: the AIS grade is strongly skewed toward grade D (97/105, 92\%), and injury etiology is spread across twelve sparsely populated categories, eight of which contain four or fewer participants. We therefore interpret the SCIM graph as the primary result, treat the AIS graph as a secondary and severity-restricted comparison, and report the injury-category and etiology graphs descriptively only.

We then performed a graph-based conditional dependence analysis that linked the segmented gait descriptors to clinical function. The preceding analyses determined whether each markerless waveform resembled its MoCap or force plate reference. The graph analysis asked a different question: after all the retained gait descriptors were considered together, which features remained directly associated with a clinical target? We focused primarily on SCIM mobility Total, an established summary measure of functional independence and mobility in patients with SCI, and treated it as a target node in a graph learned from the retained gait waveform principal component (PC) scores.

In this framework, a graph node is either a gait-waveform PC score or a clinical target. An edge does not indicate that the two variables are simply correlated. Instead, an edge indicates that two nodes remain associated after the model has adjusted for all other nodes in the graph. Therefore, the Markov blanket of SCIM mobility Total is the immediate neighborhood of SCIM mobility Total in this conditional-dependence graph: it identifies the gait-derived variables that remain directly informative about functional mobility after indirect associations through other gait components have been accounted for.

\begin{figure}[H]
\centering

\includegraphics[width=\textwidth]{
  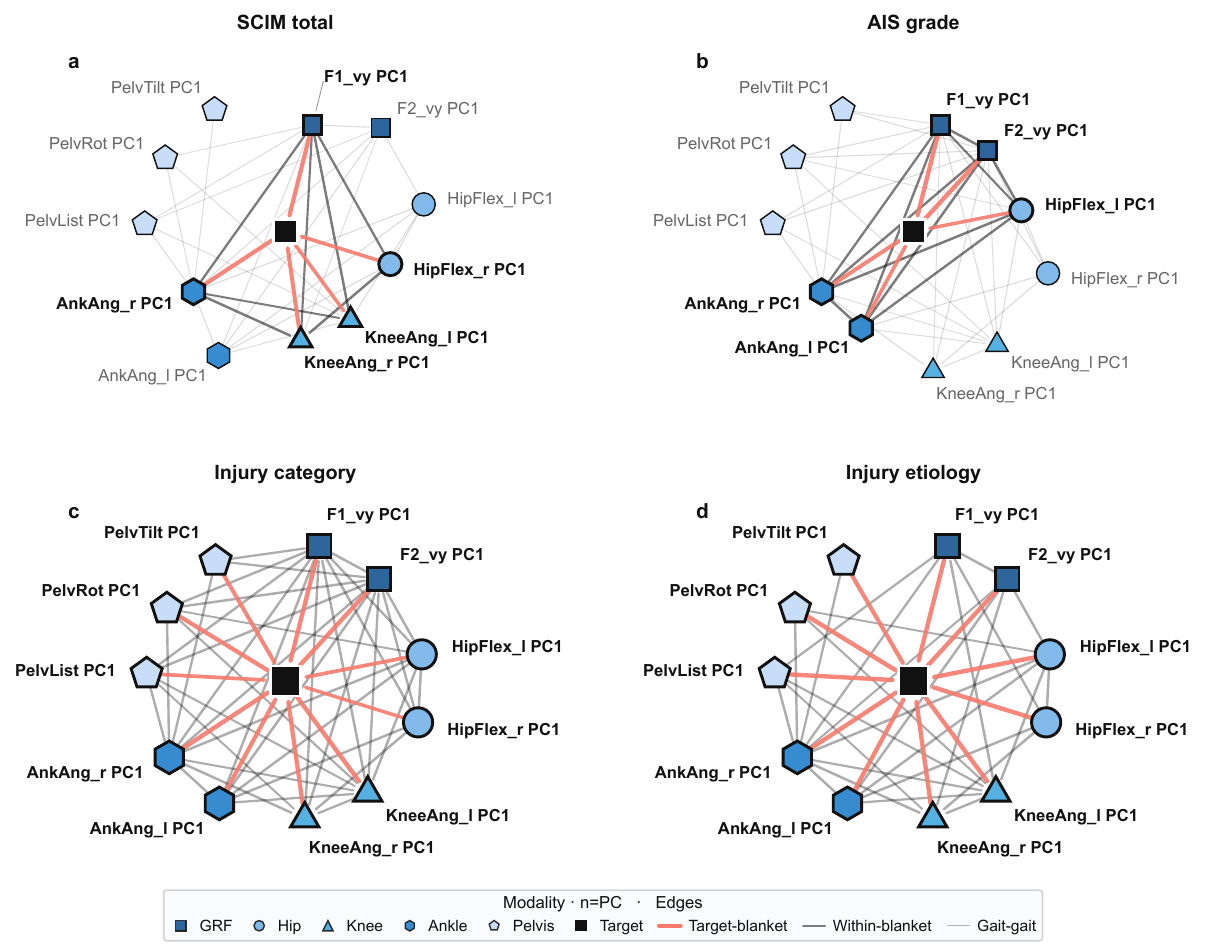
}

\caption{Figure~5. Conditional-dependence graphs linking gait waveform features to clinical targets. Fully learned graphs are shown for (a) SCIM Mobility Total (SCIM total; functional independence/mobility), (b) AIS grade (neurological impairment severity), (c) injury category/subcategory, and (d) injury etiology, with each target’s Markov blanket highlighted. The clinical target (black square) is centered; 55 gait-waveform principal components (11 waveforms $\times$ 5 PCs) are arranged radially by modality (GRFs, hip, knee, ankle, and pelvis), with node digits indicating the PC index; modality is encoded by color and shape.
Markov-blanket nodes lie on an inner radius (heavier outline) and connect to the target via coral edges; darker gray edges denote within-the-blanket dependencies, and faint edges denote remaining stable gait--gait dependencies. Edge width reflects participant-level bootstrap stability within each edge class and is not comparable across classes (\stabref{tab:scim_blanket}).}
\label{fig:scim_mb_main}
\end{figure}

\figref{fig:scim_mb_main} shows the full conditional-dependence graphs for the four clinical targets, with the target-to-Markov-blanket edges highlighted. The full graphs reveal dense interdependency among gait features spanning the GRF-, hip-, knee-, ankle-, and pelvis-related descriptors, which is consistent with gait being a coordinated whole-body task rather than a set of independent joints. Clinically, this matters because it supports the interpretability of the extracted markerless signals: the pipeline retains the expected coupling between limb advancement, pelvic control, and loading/propulsion, which underpins common gait deviations in SCI.

Importantly, compared with the full graph, the highlighted neighborhood of SCIM mobility Total was substantially smaller and more interpretable: 19 of the 55 candidate principal components (see \stabref{tab:scim_blanket}). It comprised bilateral hip flexion (left PC1--3, right PC1--4), right ankle (PC1--4) together with left-ankle PC1, right-knee PC1, pelvic-list PC2, and vertical-GRF modes from both plates (plate 1 PC1/3/5, plate 2 PC1/5). The two most reproducible members were a vertical-GRF loading mode (plate 2 PC5, $\widehat{\pi}=0.99$) and a left-hip-flexion mode (PC3, $\widehat{\pi}=0.93$). The four targets did not share a common blanket: AIS retained 11 of 55 PCs, weighted toward the ankle and loading modes, indicating that neurological severity and functional independence are associated with partly distinct gait signatures. These links align with clinical reasoning: hip-flexion features reflect step initiation and limb advancement, knee features capture stance stability and swing clearance, ankle features are related to distal control and push-off, pelvic list/tilt reflects compensatory strategies such as hip hiking or trunk compensation, and vertical-GRF components capture weight acceptance and limb loading.

This graph-based perspective is directly connected to the benchmarking results. The prominence of hip- and knee-related features is consistent with the strongest markerless agreement observed for hip and knee kinematics, suggesting that the aspects of motion most reliably reconstructed from monocular videos are also those most closely linked to functional mobility.

The inclusion of vertical GRFs and distal ankle-related components is consistent with the finding that distal kinetics were more informative and more faithfully reproduced when measured GRFs were available, whereas hip moments showed greater deviations. Conversely, the reliance of the SCIM neighborhood on loading-related variables helps explain why the fully markerless kinetic condition (with predicted GRFs) was less clinically faithful: errors in external loading estimates propagate through inverse dynamics and can obscure loading signatures that are directly associated with functional independence. Together, these findings suggest a clinically pragmatic pathway: monocular mesh-based kinematics can already capture mobility-relevant gait signatures in SCI, and coupling these kinematics with either measured or substantially improved predicted loading signals will be essential to achieve clinically trustworthy markerless kinetics.

\subsection*{Speed-disentangled gait clustering uncovers clinical impairment subphenotypes}
To stratify individuals with incomplete SCI into interpretable locomotor phenotypes, we performed a two-stage, data-driven clustering analysis in which clusters were not assigned using predefined numerical cutoffs; instead, grouping emerged from the multivariate gait feature space. In Stage~1 (capacity-level stratification), we identified speed-related gait features using a feature--speed correlation threshold $r\ge 0.5$ (including gait speed), reduced this subset to principal components, and applied K-means clustering ($k=2$--$6$) in the resulting PC space. This yielded three primary strata ordered by mean speed and labeled Slow, Medium, and Fast. Because the clustering operates in a multivariate PC space, these strata represent internally coherent patterns of spatiotemporal timing, joint motion, and loading that covary with speed rather than simple ``speed bins.''

In Stage~2 (mechanics-level stratification), we clustered within each primary stratum using the remaining nonspeed-related (residual) features ($r<0.5$ with speed) to recover distinct gait-mechanics phenotypes (reported below). Between-stratum differences were assessed using Kruskal--Wallis tests for continuous variables and chi-squared tests for categorical variables, reporting the mean (SD); within-stratum subcluster comparisons were performed using Mann--Whitney U (continuous) or chi-squared (categorical) tests, with the $p$ values reported in the Supplementary subcluster tables (see \figref{fig:spider_markerless_unified} and \sfigref{fig:supp_spider_markerbased_unified}).

\subsubsection*{Primary cluster analysis}
The primary speed strata exhibited a clear and clinically meaningful gradient in functional independence. In the marker-based cohort, SCIM mobility Total increased from 15.8 (SD 8.5) in the slow cohort to 19.7 (SD 9.7) in the medium cohort and 22.7 (SD 9.2) in the fast cohort (Kruskal--Wallis $p<0.001$; see \stabref{tab:clinical_gait_primaryspeed_markerbased}). The markerless cohort reproduced the same monotonic pattern, with SCIM mobility Total values of 15.3 (SD 8.7) (slow), 19.7 (SD 9.3) (medium), and 23.6 (SD 9.0) (fast) (Kruskal--Wallis $p<0.001$) (see \stabref{tab:clinical_gait_primaryspeed_markerless}). Taken together, these results validate that speed strata encode clinically relevant differences in independence, which is consistent with speed functioning as an integrated proxy for balance, coordination, and confidence in overground ambulation.

Motor impairment patterns further indicate a specific neuromuscular ``rate-limiter'' for speed in ambulatory iSCI. LEMS hip flexors differed strongly across speed strata in both pipelines (both $p<0.001$): marker-based slow walkers averaged 4.4 (SD 0.6), whereas fast walkers approached a ceiling at 4.9 (SD 0.4) (see \stabref{tab:clinical_gait_primaryspeed_markerbased}), and markerless showed the same ordering (slow 4.5 (SD 0.6) vs. fast 4.8 (SD 0.7)) (see \stabref{tab:clinical_gait_primaryspeed_markerless}). In contrast, knee flexors (marker-based $p=0.310$; markerless $p=0.128$) and ankle dorsiflexors (marker-based $p=0.650$; markerless $p=0.457$) showed no statistically detectable between-stratum measurement differences, which is consistent with a ceiling effect (means typically $\approx 4.7$--$4.9$ across strata; see \stabref{tab:clinical_gait_primaryspeed_markerbased} and \stabref{tab:clinical_gait_primaryspeed_markerless}).

This pattern supports the interpretation that among individuals who are already ambulatory, proximal swing-phase control (hip flexor strength for limb advancement and foot clearance) is more discriminative of achievable walking speed than distal strength. Joint mobility provided complementary evidence for speed-dependent mechanical constraints. The hip flexion ROM differed across strata in both pipelines (marker-based $p=0.031$; markerless $p=0.017$). Fast walkers exhibited more preserved hip flexion (marker-based 109.6$^{\circ}$ (SD 25.2$^{\circ}$); markerless 107.1$^{\circ}$ (SD 28.4$^{\circ}$) than slow walkers did (marker-based 96.7$^{\circ}$ (SD 36.1$^{\circ}$); markerless 93.3$^{\circ}$ (SD 39.8$^{\circ}$); see \stabref{tab:clinical_gait_primaryspeed_markerbased} and \stabref{tab:clinical_gait_primaryspeed_markerless}). The knee flexion ROM also increased with speed (marker-based $p=0.029$; markerless $p=0.017$), with marker-based fast walkers reaching 136.0$^{\circ}$ (SD 22.4$^{\circ}$) vs. 117.9$^{\circ}$ (SD 43.7$^{\circ}$) in slow walkers. These between-stratum ROM reductions are clinically consistent with spasticity-related stiffness or contracture limiting swing-phase knee flexion and step length.

As expected, gait speed differed significantly by design ($p<0.001$). Nevertheless, the strata exhibited a coherent multivariate signature consistent with progressively more efficient and stable walking. The cycle time decreased from 1.4\,s (SD 0.1\,s) in slow walkers to 1.1\,s (SD 0.1\,s) in fast walkers (both pipelines; $p<0.001$), whereas the stride length increased from 1.0\,m (SD 0.1\,m) to 1.3\,m (SD 0.1\,m) (both $p<0.001$), respectively. The double-limb support time decreased from 0.4\,s (SD 0.1\,s) to 0.2\,s (SD 0.0\,s) ($p<0.001$), indicating a reduced reliance on bilateral support as dynamic stability improved (see \stabref{tab:clinical_gait_primaryspeed_markerbased} and \stabref{tab:clinical_gait_primaryspeed_markerless}). Stride width was not significant in the marker-based context ($p=0.231$; mean $\approx 0.1\,\mathrm{m}$) but was significant in the markerless context ($p=0.015$), with slow walkers showing a wider base of support (0.13\,m (SD 0.04\,m)) than fast walkers (0.11\,m (SD 0.03\,m)), suggesting improved sensitivity of the markerless pipeline to compensatory frontal-plane stability strategies.

The strata also differed in terms of joint excursions and quality indices. The peak degree of knee flexion increased from 62.6$^{\circ}$ (SD 7.2$^{\circ}$) in slow walkers to 68.2$^{\circ}$ (SD 4.6$^{\circ}$) in fast walkers in the marker-based condition ($p<0.001$), with the markerless condition showing a similar increase (52.8$^{\circ}$ to 59.6$^{\circ}$; $p=0.032$). In contrast, peak hip flexion was not significant in either pipeline ($p>0.95$), implying that speed separation is more tightly linked to the ability to produce and control sufficient knee flexion during swing (and thus foot clearance and step length) than to the peak hip flexion amplitude alone. Peak ankle dorsiflexion was significant in the marker-based pipeline ($p=0.017$ and $p<0.001$ where reported), with slow walkers showing greater values (13.2$^{\circ}$ vs. 10.6$^{\circ}$ in fast walkers), which is consistent with a compensatory hyperdorsiflexion/steppage-like strategy for clearance. 

\begin{figure}[H]
\vspace*{-5em}
\centering
\includegraphics[width=0.83\linewidth]{
  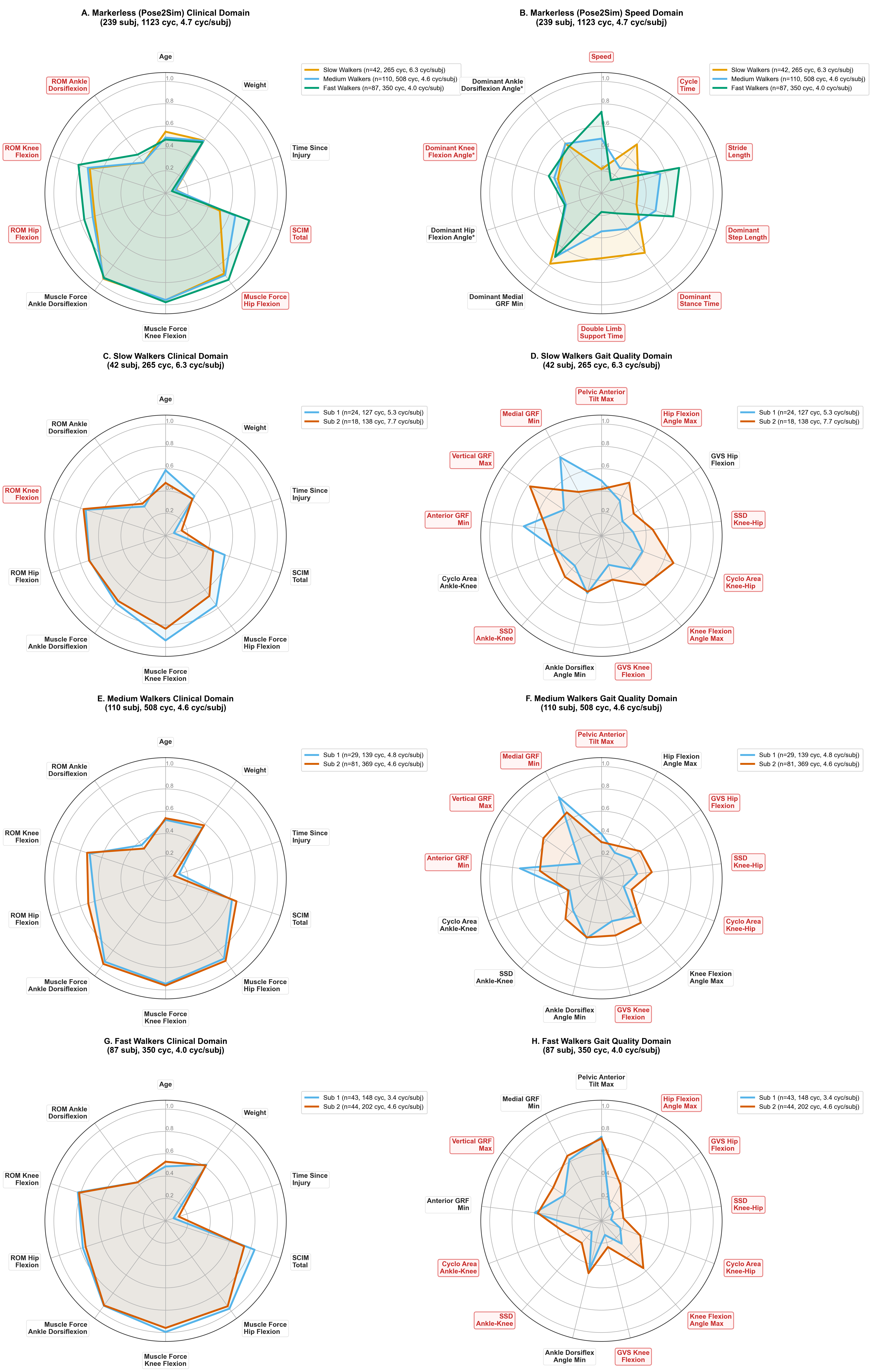
}
\caption{\small{\textbf{Unified spider-plot summary of clustering phenotypes (markerless pipeline).} Radar plots summarize the multivariate cluster profiles on a shared normalized scale (a larger radius indicates a higher normalized value). (A) Primary (speed-strata) clustering: Clinical domain profile across the slow, medium, and fast groups. (B) Primary clustering: speed/spatiotemporal gait-domain profile across the slow, medium, and fast groups. (C--D) Within-stratum subclustering in the slow group: (C) clinical-domain profile and (D) gait-quality-domain profile comparing Subcluster~1 vs. Subcluster~2. (E--F) Within-stratum subclustering in the medium group: (E) clinical-domain profile and (F) gait-quality-domain profile comparing Subcluster~1 vs. Subcluster~2. (G--H) Within-stratum subclustering in the fast group: (G) clinical-domain profile and (H) gait-quality-domain profile comparing Subcluster~1 vs. Subcluster~2.}}
\label{fig:spider_markerless_unified}
\end{figure}

The pelvic anterior tilt differed between the marker-based condition (slow walkers 11.6$^{\circ}$ vs. fast walkers 8.1$^{\circ}$; $p=0.013$) but not the markerless condition ($p=0.972$), which is consistent with the greater susceptibility of video-based pelvis tracking to soft-tissue and clothing artifacts.

Finally, the gait quality indices provided a high-level summary of the same progression. All GVS parameters differed across speed strata (marker-based $p<0.001$; markerless $p<0.05$), with GVS knee flexion decreasing from 11.7 (SD 3.2) in slow walkers to 6.5 (SD 2.8) in fast walkers in the marker-based condition, indicating closer-to-normative kinematics in faster-walking individuals. Coordination metrics revealed concordant changes: the SSD of the knee--hip decreased ($p<0.001$), whereas the cyclogram of the knee--hip area increased ($p<0.001$) from 1250.0 to 1674.2 in the marker-based condition and from 966.1 to 1414.1 in the markerless condition, which is consistent with the expansion and stabilization of interjoint coordination loops as gait improved. Kinetically, the anterior GRF minimum (braking) was highly significant in both pipelines ($p<0.001$), with fast walkers producing larger braking magnitudes ($-0.1$~BW vs. $-0.0$~BW). The maximum vertical GRFs were highly significant in the marker-based condition ($p<0.001$; slow 0.9~BW vs. fast 1.2~BW) but not in the markerless condition ($p=0.101$), which is consistent with the markerless pipeline capturing gross loading trends while underresolving the peak vertical impact features.

\figref{fig:spider_markerless_unified} provides an ``at-a-glance'' visual synthesis of the primary (speed-strata) clustering results. Panels A--B confirm that the primary clustering step recovers a coherent capacity axis: across slow medium fast, the clinical domain expands radially for mobility/strength/ROM variables, whereas the spatiotemporal/gait domain contracts for timing-related variables (cycle time and double support) and expands for speed-linked variables (step/stride length). In other words, the primary clusters are not merely ``speed bins'' but reflect coupled clinical--biomechanical progression.

\subsubsection*{Within-stratum subcluster analysis (gait quality phenotypes)}

\textbf{Clinical homogeneity but biomechanical separation.} Across strata and pipelines, clinical variables typically did not differ significantly between subclusters (generally $p>0.05$), indicating that Stage~2 did not simply rediscover gross clinical severity. In contrast, gait variables frequently showed statistically significant and clinically interpretable subcluster separation (often $p<0.05$ and sometimes $p<0.01$), demonstrating that individuals with comparable capacity (speed) can exhibit distinct coordination and compensation phenotypes. This within-speed separation provides evidence that there are identifiable patterns between groups beyond speed itself, and it motivates a therapy-oriented interpretation: if a subcluster is characterized by restricted swing-phase knee motion, reduced coordination-loop area, or elevated deviation indices, these features represent candidate modifiable targets, and improving them may be more actionable than attempting to change speed directly.

\textbf{Slow stratum: restricted/compromised mechanics vs. relatively preserved mechanics at low speed.} With respect to markerless slow walkers, subclusters were highly significantly separated by swing-phase knee mechanics and coordination: the maximum degree of knee flexion differed (Mann--Whitney $p=0.005$; 47.3$^{\circ}$ vs. 60.2$^{\circ}$), as did the degree of GVS knee flexion ($p=0.009$; 26.2 vs. 29.2), the cyclogram knee--hip area ($p=0.002$; 757.4 vs. 1244.5), and the maximum vertical GRFs ($p<0.001$; 0.7~BW vs. 1.0~BW) (see \stabref{tab:supp_subclusters_slow_ml}). These differences imply that, even at similarly low speeds, one subgroup exhibits markedly reduced knee excursion and a contracted coordination loop (consistent with global mechanical limitation), whereas the other maintains greater excursion/coordination and higher loading capacity, which is consistent with a ``better quality'' gait pattern that is nonetheless performed cautiously or under nonmechanical constraints.

\textbf{Middle stratum: coordinated ``stiff'' strategy vs. compensatory high-excursion strategy.} In marker-based medium walkers, subclusters were distinguished by hip kinematics and associated deviation: the maximum degree of hip flexion differed strongly ($p<0.001$; 30.6$^{\circ}$ vs. 41.9$^{\circ}$) together with the degree of GVS hip flexion ($p<0.001$; 5.8 vs. 11.9) (see \stabref{tab:supp_subclusters_medium_mb}). This pattern suggests two mechanically distinct solutions to achieve medium walking speed: one subgroup walks with smaller hip excursions but relatively closer-to-normative hip kinematics (lower GVS), whereas the other achieves comparable speed via greater hip excursion but with more deviant coordination (higher GVS), which is consistent with compensation through proximal motion.

\textbf{Fast stratum: high-excursion phenotype vs. closer-to-normative phenotype.} In fast walkers, subclusters were again separated primarily by sagittal-plane excursions and gait-quality indices. In the marker-based pipeline, Subcluster~1 had a greater maximum degree of hip flexion ($p=0.001$; 44.4$^{\circ}$ vs. 33.0$^{\circ}$) and greater GVS hip flexion ($p=0.001$; 12.9 vs. 4.8) and GVS knee flexion ($p=0.018$; 8.1 vs. 5.9) (see \stabref{tab:supp_subclusters_fast_mb}). In the markerless pipeline, the subclusters exhibited broader separation that additionally involved distal joints and coordination: Subcluster~2 had larger hip, knee, and ankle excursions (all $p=0.001$; hip 44.5$^{\circ}$ vs. 33.0$^{\circ}$, knee 69.5$^{\circ}$ vs. 49.4$^{\circ}$, and ankle 26.9$^{\circ}$ vs. 20.3$^{\circ}$), a larger cyclogram knee--hip area ($p=0.001$; 1734.4 vs. 1087.3), and greater vertical maximum GRFs ($p=0.009$; 1.0~BW vs. 0.9~BW) (see \stabref{tab:supp_subclusters_fast_ml}). Together, these results suggest that, even among high-capacity walkers, one phenotype achieves fast speed with a more constrained, closer-to-normative kinematic profile, whereas the other relies on amplified excursions (and, in markerless condition, higher loading/coordination amplitude) with a concomitantly greater deviation.

\textbf{The residual speed catching is limited.} Because speed was used only to define the primary strata, the within-stratum gait speed did not differ significantly between subclusters (marker-based slow $p=0.492$; marker-based medium $p=0.740$), supporting the interpretation that this separation reflects residual mechanics rather than remaining speed differences.

\textbf{Methodological equivalence across pipelines.} Both marker-based and markerless analyses revealed subclusters primarily through differences in joint excursions and gait quality indices (GVS/SSD/cyclograms), with the markerless pipeline recovering the same directionality of key separations (knee flexion and cyclogram area differences in slow subclusters). This finding supports the clinical resolution of markerless gait analysis for detecting fine-grained coordination phenotypes beyond simple-speed stratification.

\subsubsection*{Cross-speed-combined analysis}
To test whether the within-stratum subcluster structure is preserved when walking speed is no longer used to separate participants, we additionally pooled the Slow/Medium/Fast cohorts onto a shared normalized feature axis and recomputed the cross-subcluster similarity structure on the combined dataset (see \sfigref{fig:supp_cross_speed_combined_markerless} and \sfigref{fig:supp_cross_speed_combined_markerbased}).

\section*{Discussion}

\subsubsection*{Scope and primary contributions}
This study evaluated monocular, calibration-free video pipelines for gait biomechanics with the goal of defining an end-to-end workflow suitable for individuals with spinal cord injury (SCI). Rather than focusing only on pose estimation accuracy, we tested whether video-derived outputs could support downstream biomechanical inference, clinically meaningful lower-limb kinematics, force-plate-free estimates of external loading, and multivariate descriptors that preserve functionally relevant structure in a heterogeneous neurological cohort, with the longer-term goal of enabling therapy decisions when it is difficult to predict who will improve in gait speed. We benchmarked a monocular \texttt{CameraHMR}-based, \texttt{SMPL}-driven pipeline against synchronized marker-based motion capture (MoCap) and force-plate data in a large SCAI SCI Gait dataset. On a 9-keypoint lower-body set, \texttt{CameraHMR} achieved an overall PA-MPJPE of 83.0~mm (\tabref{tab:combined_benchmarking}). A prior report on the same dataset cited an approximate PA-MPJPE of 90.7~mm using a 7-keypoint configuration and a different valid-frame subset; thus, the values are not strictly comparable \cite{Natraj2025npjDM}.

\subsubsection*{Markerless kinematics with clinical fidelity in SCI}
Downstream kinematics also provided the clearest validation that monocular 3D-HMR can recover clinically informative gait structure in SCI patients. Overall correlation coefficients ranged from $r=0.68$ to $0.90$ ($p<0.05$; \eqhref{pearson_eqn}), and the RMSE ranged from approximately 4.18$^{\circ}$--6.49$^{\circ}$ (\eqhref{rmse_eqn}). Despite occasional MoCap foot-marker placement over footwear rather than the anatomical foot, the markerless pipeline showed high sagittal plane agreement at the ankle and very high agreement at the hip and knee (\figref{fig:ik_p2s_centred}, \tabref{tab:ik_metrics_models}). Although distal segments are typically more sensitive to occlusion, the overall waveform fidelity supports the conclusion that this system can recover properly scaled lower-body kinematics suitable for OpenSim modeling and subsequent kinematic analyses in pathological gait populations \cite{Wade2022, PETERS20101, GORTON2009398}. Compared with existing methods, our per-frame mesh pipeline achieved a 3.4$^{\circ}$--5.3$^{\circ}$ mean absolute error at the hip and knee in a neurological cohort without optimization (\eqhref{mae_eqn}). This is comparable to the 4.8$^{\circ}$ mean absolute error reported for OpenCap Monocular in healthy adults after a two-stage optimization \cite{OpenCapMonocular2026}. Together, these results suggest that the anatomical representation provided to downstream reconstruction can materially affect the fidelity of the resulting biomechanical model (see \stabref{tab:system_comparison}).

\subsubsection*{Force-plate-free loading with predicted GRF (pGRF)}
The predicted GRF magnitude was also strongly associated with the force-plate reference ($r\approx0.85$--0.87; \eqhref{pearson_eqn}; see \figref{fig:pgrf_p2s}, \tabref{tab:pgrf_results}), indicating that the kinematics-driven model recovered the dominant stance-phase loading envelope and captured broad between-step variation. Because GRFs are typically the key missing input when force plates are unavailable, recovering a credible proxy from video-derived motion substantially improves the scalability of gait assessment beyond specialized laboratories.

\subsubsection*{Multivariate phenotyping using conditional dependence and clustering}
The causal graph conditional-dependence analysis extended the interpretation of the waveform-level validation by examining gait as an interconnected multivariate system rather than a collection of independent joint trajectories. The SCIM total score graph provided the most interpretable clinical result, where the Markov blanket contained 19 of the 55 candidate components and included bilateral hip flexion, both ankle variables, right knee features, the pelvic list, and the vertical GRFs from both force plates (\figref{fig:scim_mb_main}, \stabref{tab:scim_blanket}). The composition of this neighborhood indicated that functional independence was associated with a distributed locomotor configuration encompassing proximal limb advancement, distal control, pelvic organization, and external loading, which was consistent with the functional demands of walking in SCI, where successful locomotion depended on coordinated control of limb progression, foot clearance, weight acceptance, propulsion, and compensatory trunk or pelvic strategies. Clinically, such minimal sets of conditionally relevant features can help prioritize which measurable patterns to target in therapy and which signals to track when evaluating whether a patient is likely to improve in speed or is instead adopting a different compensatory solution. This graph reinforced the argument that clinically meaningful gait impairment could not be reduced to an abnormality in a single joint.

The clustering analyses provided a complementary demonstration that markerless biomechanics could preserve clinically meaningful interindividual structure even in a heterogeneous SCI dataset. The primary clustering identified a coherent slow, medium, and fast stratification aligned with locomotor capacity variables, including step/stride length, gait cycle timing, and double-support characteristics (\figref{fig:spider_markerless_unified}), whereas the within-speed-stratum subclustering was arguably more clinically informative because participants who achieved comparable walking speeds could nevertheless be separated into distinct gait-quality phenotypes driven by residual joint excursions, coordination measures such as SSD and cyclogram area, and deviation measures such as GVS (\stabref{tab:supp_subclusters_slow_ml}, \stabref{tab:supp_subclusters_medium_mb}). These findings demonstrate that walking speed represents only one dimension of neurological gait: two individuals can achieve similar speeds while relying on different coordination patterns and compensations, implying that the same speed outcome may reflect different limiting mechanisms. From a translational standpoint, this directly addresses the clinical question of whether we can observe repeatable patterns between patient groups and, if so, which pattern(s) are most promising to target and improve with therapy. Clinically, this supports a phenotype-guided therapy rationale in which the goal is not only to increase speed but to shift a patient toward a more efficient, less deviant, and more stable strategy within their capacity stratum by targeting the subcluster-defining features, and to track whether that targeted pattern changes over time. In this sense, cluster membership and the associated waveform features provide a structured way to generate testable hypotheses about which impairments are most modifiable for a given subgroup and to design rehabilitation that is more specific than speed-focused progression alone\cite{Werner2023, Hulleck2022}. The fact that similar subcluster drivers emerged across marker-based and markerless representations strengthened the interpretation, suggesting that the phenotypic structure was not solely created by errors in the monocular reconstruction pipeline and that a substantial component of relevant variability was preserved despite the transition from laboratory measurements to monocular video-derived biomechanics, supporting the broader proposition that markerless systems might be able to provide scalable phenotype-aware gait characterization without requiring all of the instrumentation associated with conventional gait laboratories.

\subsubsection*{Limitations with Inverse Dynamics}
We also examined kinetics by pairing markerless kinematics with measured force-plate GRFs (semimarkerless kinetics). Inverse dynamics showed substantial agreement with MoCap at the distal joints (ankle: $r\approx0.84$--0.87; knee: $r\approx0.73$--0.74; see \sfigref{fig:id_p2s}--\sfigref{figure:id_raja} and \stabref{tab:semi_markerless_id}), but agreement progressively degraded toward the hip, consistent with distal-to-proximal propagation of uncertainty in linked-segment inverse dynamics \cite{RIEMER2008578, BENEDETTI2013934}. Notably, this proximal degradation was driven primarily by numerical sensitivity and modeling assumptions in the standard OpenSim inverse dynamics formulation \cite{OpenSim}, rather than by camera-based reconstruction errors; the solver produced plausible but systematically biased proximal moments even when the force-plate GRFs were available, and these sensitivities became substantially more limiting under fully markerless conditions by perturbing effective moment arms and magnifying residual errors (\sfigref{fig:pid_p2s}--\sfigref{figure:pid_raja} and \stabref{tab:semi_markerless_id}).

\subsubsection*{Future directions}
Future progress toward robust fully markerless kinetics will likely require data-driven inverse-dynamics approaches such as ImDy, which uses imitation learning and physics-based simulation to predict internal and external joint moments from kinematics alone \cite{liu2025imdy}; by jointly optimizing the relationship between motion, external loading, and internal joint loading, such learned dynamics models could reduce the sequential error propagation inherent to conventional inverse dynamics (which estimates kinematics and GRFs separately and then inserts them into a deterministic solver) and may be especially effective when the training distribution explicitly represents neurological gait. Beyond kinetics, extending the workflow to estimate muscle activation patterns and additional patient descriptors via a clinical dashboard could increase clinical richness by transforming monocular video from a passive measurement modality into an interactive clinical assessment tool.

\subsubsection*{Conclusion}
Taken together, these results establish a realistic and differentiated performance profile for monocular, calibration-free, and SMPL-based gait biomechanics in patients with SCI, and support monocular gait analysis as a practical option for at-home and scalable assessment. The proposed pipeline recovers clinically meaningful sagittal-plane joint-angle waveform structure from a single camera, with hip, knee, and ankle sagittal-plane MAEs remaining close to 5$^{\circ}$, a commonly used threshold for clinical acceptability \cite{MCGINLEY2009360}, and with waveform fidelity sufficient for many gait-cycle and pattern-based analyses. Beyond kinematics, video-driven GRF magnitude reproduces the dominant stance-phase loading envelope (with correlations), yielding interpretable coarse loading descriptors when force plates are unavailable. The graph and clustering results further indicate that the monocular biomechanical representation is clinically richer than trajectory agreement alone: the SCIM Markov blanket identifies a distributed set spanning hip, knee, ankle, pelvic, and vertical-loading variables, and clustering shows that individuals with similar walking speeds can nonetheless exhibit distinct coordination and gait-quality phenotypes.

Collectively, these findings suggest that monocular markerless kinematics are sufficiently mature to complement (and partially substitute for) conventional MoCap-based sagittal-plane gait assessment, particularly where access to specialized gait laboratories is limited and repeated in-lab testing is challenging. The calibration-free workflow improves feasibility for repeated assessment across locations and over longer periods where repeated measurement can track whether an individual is moving toward a more efficient pattern within their phenotype and help stratify responders versus non-responders when speed alone does not explain improvement. 

\section*{Methods}
\subsection*{Dataset}

This study used the \texttt{SCAI SCI Gait} dataset, a custom clinical gait dataset collected from \textbf{239} individuals with \texttt{SCI} at the \texttt{Swiss Paraplegic Center (SPZ)} in Nottwil, Switzerland. Data were acquired during routine instrumented gait analysis sessions conducted between \textbf{2015 and 2022}, spanning a broad age range (\textbf{18--83} years). All participants provided written informed consent, and data collection and processing were approved by the Ethics Committee of Northwest and Central Switzerland (Project-ID \textbf{2022-00935}). The cohort comprised \textbf{176 males} (\textbf{74\%}) and \textbf{63 females} (\textbf{26\%}), with mean (\,\textpm\,SD) anthropometrics at examination of $\mathbf{46.62 \pm 15.52}$ years, $\mathbf{1.76 \pm 0.10}$ m, and $\mathbf{75.27 \pm 15.69}$ kg for age, height, and bodyweight, respectively. Injury presentation was classified as tetraplegia (\textbf{49\%}, \textit{n}=110) or paraplegia (\textbf{51\%}, \textit{n}=115), with neurological level of injury distributed across cervical (C1--C8, \textbf{48\%}), thoracic (T1--T12, \textbf{34\%}), lumbar (L1--L5, \textbf{17\%}), and sacral (S1--S3, \textbf{1\%}) regions. Time since injury was highly variable, reflecting both acute and chronic cases (mean: $\mathbf{47.2 \pm 97.5}$ months; range: \textbf{0--554} months).

Clinical gait analyses were performed on a \textbf{10 m} level walkway under standardized
laboratory conditions using a synchronized setup comprising marker-based
motion capture (MoCap), \textbf{two} instrumented force plates for ground-reaction
force (GRF) measurement, and \textbf{two} RGB cameras mounted at approximately
\textbf{1.5 m} height to provide sagittal and coronal views. The acquisition
system was installed at \texttt{SPZ} by \texttt{Qualisys}, and all modalities were
recorded as part of the routine clinical protocol. MoCap marker trajectories and GRF were
sampled at \textbf{100 Hz} or \textbf{200 Hz}, while video was recorded at either
\textbf{1920$\times$1080, 50 FPS} (high-resolution) or \textbf{720$\times$576,
25 FPS} (low-resolution). 

Across all recorded trials, the \texttt{SCAI SCI Gait}
dataset contains \textbf{49,838} video frames, of which \textbf{46,717} frames
yielded valid human detections suitable for downstream \texttt{SMPL} mesh
reconstruction (\textbf{36,925} high-resolution and \textbf{9,792} low-resolution
frames). Cohort demographics and evaluation-set composition are summarized (see
\figref{fig:dataset_summary}).

\subsection*{Monocular Human Mesh Recovery (HMR)}

For each monocular RGB frame, the HMR estimates the \texttt{SMPL} parameters (pose
and shape) together with the camera parameters that map the recovered body to
the image plane. We evaluated five monocular \texttt{SMPL}-based estimators (\texttt{CameraHMR}\cite{patel2024camerahmr},
\texttt{WHAM}\cite{shin2024wham}, \texttt{Multi-HMR}\cite{multi-hmr2024},
\texttt{W-HMR}\cite{yao2024whmr}, and \texttt{PLIKS}\cite{Shetty_2023_CVPR})
on the \texttt{SCAI SCI Gait} dataset (see
\tabref{tab:combined_benchmarking}). Each
method was run on complete trial videos to obtain a temporally ordered sequence of meshes/poses; therefore, any temporal consistency was method-dependent (\texttt{WHAM} refines estimates over a video window, whereas most others are primarily frame-based). Because these approaches differ in output conventions (axis definitions, camera parameterization, whether joints are reported as regressed \texttt{SMPL} joints or auxiliary keypoint sets, and whether results are expressed in metres or millimeters), all outputs were converted to a common representation before evaluation.

For every method and frame, we extracted a shared 9-joint lower-body keypoint set (pelvis/root, and bilateral hip, knee, ankle, and toe). When a method provided \texttt{SMPL} parameters and vertices, joints were computed using the standard \texttt{SMPL} joint regressor and then restricted to the 9-joint subset.

When a method directly returned joints/keypoints, we mapped the outputs to the same semantic joint definitions. All predicted 3D joint coordinates were expressed in a consistent right-handed coordinate convention and converted to
millimeters for scoring. The \texttt{SCAI SCI Gait} recordings provide synchronized monocular video and MoCap trajectories. For each trial, we used the motion-capture stream as a reference and derived a corresponding 9-joint 3D trajectory in the laboratory/world coordinate system. Specifically, MoCap marker trajectories were converted from \texttt{.c3d} to \texttt{.trc} using the same conversion pipeline used elsewhere in this work \cite{Bagheri_2024}, which also maps trajectories into the \texttt{OpenSim} coordinate convention. The reference 3D joint centers were then obtained by fitting the musculoskeletal model to the MoCap markers (inverse kinematics) and exporting the resulting model joint center positions for the pelvis,
hips, knees, ankles, and toes. Finally, these reference joint trajectories were temporally aligned to the video frame times by interpolation using dataset synchronization, ensuring a one-to-one frame correspondence for per-frame error computation.

We quantified 3D reconstruction accuracy using the Procrustes-aligned mean per-joint
position error (PA-MPJPE) over the 9-keypoint set (\eqhref{eq:pa_mpjpe_align}--\eqhref{eq:pa_mpjpe}).
Let $\hat{\mathbf{x}}_{j} \in \mathbb{R}^{3}$ denote the predicted 3D position
of joint $j$ and $\mathbf{x}_{j} \in \mathbb{R}^{3}$ the corresponding
reference position. For each frame we estimate the optimal similarity transform
(rotation $\mathbf{R}\in SO(3)$, scale $s>0$, translation
$\mathbf{t}\in \mathbb{R}^{3}$) by solving the orthogonal Procrustes problem:

\begin{figure}[H]
\vspace*{-2.0em}
\centering

\begin{subfigure}
  {0.95\linewidth}
  \centering
  \begin{tabular}{ll}
    \hline
    \textbf{Dataset}          & \texttt{SCAI SCI Gait}                                 \\
    \hline
    \textbf{Population}       & SCI                                       \\
    \textbf{Subjects}         & 239                                                \\
    \textbf{Gender Ratio}     & 176 males, 63 females                              \\
    \textbf{Mean Age}         & $46.62 \pm 15.52$ years (18--83)                   \\
    \textbf{Mean Height}      & $1.76 \pm 0.10$ m                                  \\
    \textbf{Mean Weight}      & $75.27 \pm 15.69$ kg                               \\
    \textbf{Trials}           & 307 total (1--3 per subject; single sagittal POV)  \\
    \textbf{Video Resolution} & 1920$\times$1080 (50 FPS), 720$\times$576 (25 FPS) \\
    \textbf{Total Frames}     & 49,838 total; 46,717 w/ humans (36,925 high-res., 9,792 low-res.) \\
    \hline
  \end{tabular}
  \subcaption{\small{\textbf{\texttt{SCAI SCI Gait} Dataset Distribution:} This dataset, comprising individuals with SCI, provided the monocular video, marker-based motion capture, and GRF data, which were used for \texttt{SMPL} mesh reconstruction and subsequent musculoskeletal analyses. The video data include both high-resolution (1920$\times$1080 at 50~FPS) and low-resolution (720$\times$576 at 25~FPS) sequences.}}
  \label{tab:id_dataset}
\end{subfigure}

\vspace{0.75em}

\begin{subfigure}
  {0.95\linewidth}
  \centering
  \includegraphics[width=0.8\linewidth]{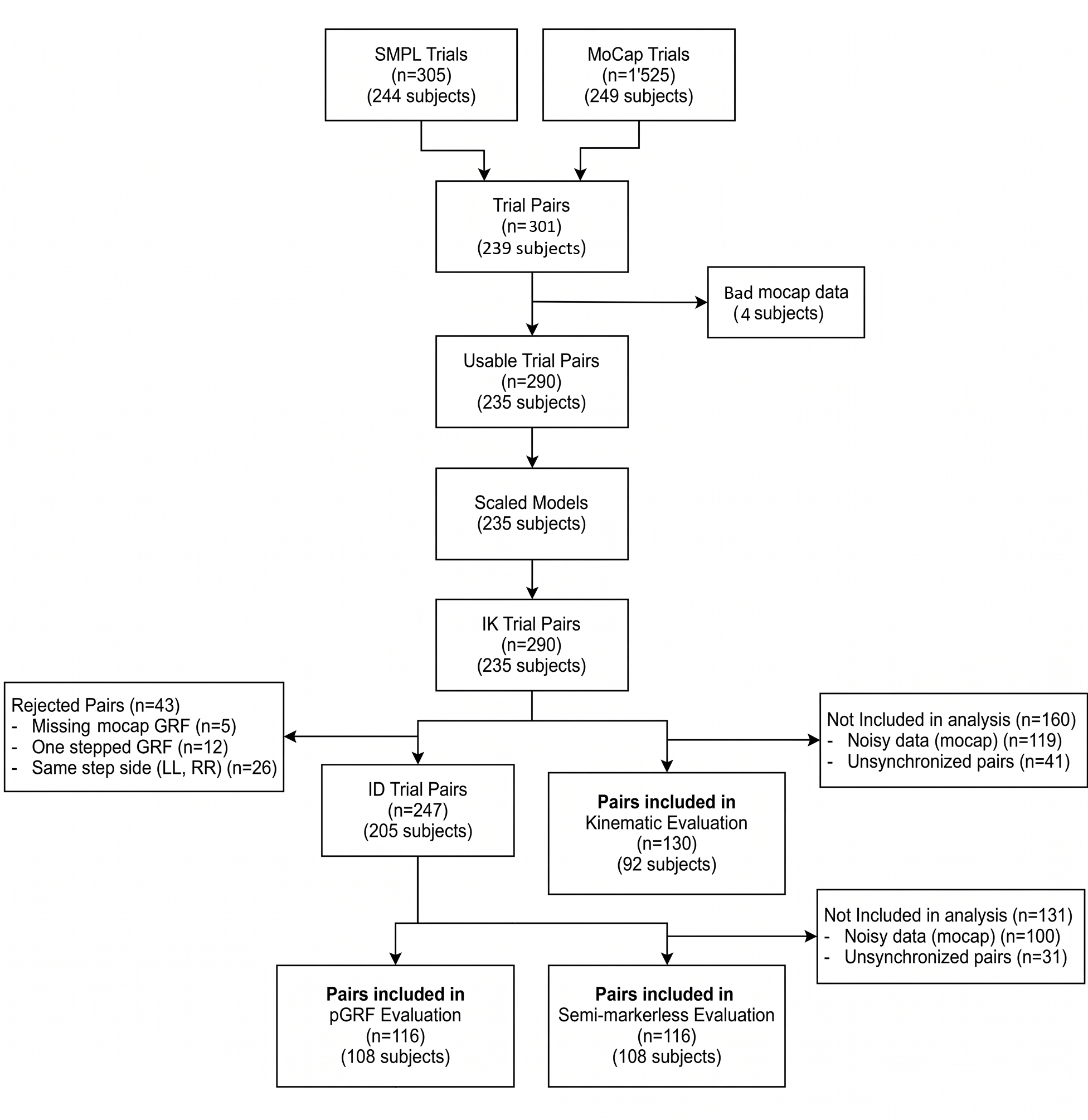}
  \subcaption{\small{\textbf{Overview of SCAI SCI Gait Trials Included.} The given figure shows the number of \texttt{SCAI SCI Gait} trials, and corresponding number of SCI subjects, included for each evaluation. In the \textit{markerless \textbf{kinematic}} evaluation, 130 trials from 92 subjects were included. \textit{Semi-markerless \textbf{kinetics}} evaluation and the evaluation of \texttt{GaitDynamics} prediction both included 116 trials from 108 subjects. Fully \textit{markerless \textbf{kinetics}} were evaluated on the same cohort as semi-markerless kinetics, for the exception of two trials from a single subject.}}
  \label{fig:trials_included}
\end{subfigure}

\caption{\small{\textbf{\texttt{SCAI SCI Gait} cohort summary.} Overview of the \,\texttt{SCAI SCI Gait}\, dataset used throughout this study. \textbf{(a)} \emph{Dataset-level demographics and acquisition characteristics}: summary statistics describing the cohort composition (sample size, age distribution, sex, body size and other available anthropometrics, and relevant clinical descriptors) together with key recording metadata (acquisition site(s), hardware/sensor configuration, sampling rate, protocol/track conditions, and any preprocessing or quality-control criteria applied prior to inclusion). \textbf{(b)} \emph{Evaluation set composition}: number of unique subjects and total trials contributing to each reported evaluation, after applying all inclusion/exclusion rules and any train/validation/test partitioning. Counts are shown to make explicit which portions of the cohort support each experiment and to clarify the balance of subjects versus repeated trials within each evaluation.}}%
\label{fig:dataset_summary}
\end{figure}

\begin{equation}
\label{eq:pa_mpjpe_align}(\mathbf{R}^{*}, s^{*}, \mathbf{t}^{*}) = \arg\min_{\mathbf{R},s,\mathbf{t}}
\sum_{j=1}^{J}\left\| s\,\mathbf{R}\,\hat{\mathbf{x}}_{j} + \mathbf{t}- \mathbf{x}
_{j} \right\|_{2}^{2},
\end{equation}
with $J=9$. The \textit{PA-MPJPE} (mm) is then computed as
\begin{equation}
\label{eq:pa_mpjpe}\mathrm{PA\text{-}MPJPE}= \frac{1}{J}\sum_{j=1}^{J}\left\|
s^{*}\,\mathbf{R}^{*}\,\hat{\mathbf{x}}_{j} + \mathbf{t}^{*} - \mathbf{x}_{j}
\right\|_{2}.
\end{equation}

We used \textit{PA-MPJPE} because monocular, calibrationless HMR has
unavoidable global ambiguities (unknown camera-to-world placement and absolute
scale). By removing global translation, orientation, and scale, \textit{PA-MPJPE}
yields a view- and scale-invariant measure of relative 3D pose accuracy, enabling
fair comparison across recordings with heterogeneous viewpoints and resolutions
and directly reflecting the geometric consistency of the recovered lower-body pose.
In addition to \textit{PA-MPJPE}, we report the number of \textit{valid frames}
per method. A frame was considered valid if the method returned a complete 3D
reconstruction from which all 9 benchmarking keypoints (and the required \texttt{SMPL}
surface vertices for marker extraction) could be obtained without missing values.
This count serves as a practical robustness metric under occlusion, motion blur,
truncation at image boundaries, and assistive-device interactions. For
comparability with downstream gait segmentation and to reduce trial initiation/termination
artifacts, we trimmed 5\% of frames from the beginning and end of each
recording prior to evaluation.

\subsection*{Mesh recovery to virtual markers and \texttt{OpenSim} inputs}
For musculoskeletal analyses, temporally ordered \texttt{SMPL} meshes were reconstructed from monocular \texttt{SCAI SCI Gait} videos using \texttt{CameraHMR} \cite{patel2024camerahmr} (best-performing method for this SCI cohort; see \tabref{tab:combined_benchmarking}), using the sagittal camera view to maximize full-body visibility and minimize occlusions (including assistive-device interactions). To obtain clinically interpretable trajectories for \texttt{OpenSim}, each mesh sequence was converted to \emph{virtual markers} by sampling \textbf{105} predefined \texttt{SMPL} surface vertices approximating a full-body clinical markerset (see \sfigref{fig:smpl_grid_full_p2s}--\sfigref{fig:smpl_grid_parts}); per-frame vertex coordinates were exported as \texttt{.npy}, assembled into trajectories with custom Python scripts built on \texttt{AddBiomechanics} \cite{AddBiomech_lib}, and written to \texttt{OpenSim}-compatible \texttt{.trc} files (marker labels, frame indices, timestamps, units). These \texttt{.trc} files were used to scale the \texttt{OpenSim} model (\texttt{Scale Tool}) and compute joint kinematics (\texttt{IK Tool}). Because \texttt{CameraHMR} is camera-centred (no explicit world frame/ground plane), we applied (i) a static vertical offset (pelvis translation) to place the feet near ground level and (ii) a dynamic global alignment adjusting horizontal translation to match the measured GRF center of pressure (CoP), yielding physically plausible foot--ground contact and consistency with force-plate data. All time series (MoCap markers, \texttt{SMPL} virtual markers, recorded GRF, predicted GRF) were low-pass filtered (4th-order Butterworth, 6~Hz) and resampled by linear interpolation to 100~Hz (see \figref{fig:osim_pipeline_overview}). \texttt{SMPL} trajectories were then linearly time-warped to the MoCap trial duration, converted from metres to millimetres, and bilateral marker labels were swapped to correct a systematic left/right inversion (excluding midline markers). This inversion arose because participants traversed the walkway in both directions (right-to-left in some videos and left-to-right in others, on the return path). Accordingly, trials recorded with the opposite walking direction were handled by mirroring (\textit{168 trials}: videos horizontally mirrored prior to reconstruction) to restore consistent directionality/laterality with MoCap and GRF.

\subsubsection*{Subject Specific Scaling}

Participant-specific scaling was performed in \texttt{OpenSim 4.5} using the \texttt{Scale Tool} under two parallel input conditions: \textit{marker-based scaling} driven by recorded MoCap marker trajectories (\texttt{.c3d}$\rightarrow$\texttt{.trc}; \textbf{20} total markers, \textbf{18} retained for scaling) and \textit{markerless scaling} driven by \texttt{SMPL}-derived virtual-marker trajectories exported from mesh recovery (\textbf{105} vertices for the \texttt{Pose2Sim} configuration and \textbf{77} vertices for the Rajagopal configuration, with \textbf{68} and \textbf{50} retained for scaling, respectively; see \stabref{tab:markers_tools}). Marker-based scaling served as the reference personalization (clinical MoCap workflow), while virtual-marker scaling quantified how well mesh-derived markers could personalize the same musculoskeletal model.

For each participant and model variant, we generated a dedicated \texttt{Scale Tool} setup file (\texttt{.xml}) specifying the input \texttt{.trc} file, the corresponding \texttt{OpenSim} marker-set definition, the segments to scale, segment-specific scale-factor initialisation/constraints, and marker weights. Segment lengths were inferred from explicit bilateral marker-pair definitions where available (pelvis/hip, femur, tibia); segments lacking reliable marker-pair constraints were assigned fixed multiplicative scale factors (see \stabref{tab:scale_factors}). During scaling, \texttt{OpenSim} optimizes segment scale factors (and, where enabled, marker offsets) to minimize \emph{weighted} marker misfit in the scaling pose while updating inertial properties consistently with the scaled geometry (see \figref{fig:scaling}). Model mass was set from clinical bodyweight records; when unavailable, sex-specific cohort means were used (\textbf{male: 79.59 kg}; \textbf{female: 62.25~kg}); if sex was missing, the overall cohort mean was used (\textbf{75.09~kg}).

Scaling choices balanced anatomical plausibility with robustness to marker variability in the clinical dataset. Because \texttt{SCAI SCI Gait} provides only lower-limb MoCap markers, upper-body segments were left at default scale in the marker-based condition. Foot dimensions were held fixed (not scaled from marker pairs) due to heterogeneous footwear and inconsistent heel-marker placement. A bilateral MoCap marker pair near the thigh/greater-trochanter region was excluded for both models because placement varied substantially across trials (see \sfigref{fig:mocap_markerset}). Otherwise, consistent lower-limb anatomical landmarks were used across modalities: bony landmarks were up-weighted (\textbf{weight=10}) relative to soft-tissue markers (\textbf{weight=1--5}); in the \texttt{SMPL}-based configuration, head/adipose-region markers were excluded when unstable and pelvis/shoulder markers were moderately up-weighted (\textbf{weight=2}) to stabilize global alignment (see \stabref{tab:scale_factors}--\stabref{tab:markers_tools}). For the Rajagopal configuration, upper-limb virtual markers were excluded to match model topology (see \sfigref{fig:smpl_grid_full_raja}).

\subsubsection*{Inverse Kinematics}
Inverse kinematics (IK) was performed for every trial using the \texttt{OpenSim} \texttt{IK Tool} on the participant-specific scaled model from the corresponding scaling condition. For each trial, we ran \textbf{two} IK solves with identical model coordinates, joint constraints, solver settings, and analysis time window, differing only in the experimental marker trajectories: a reference solve driven by filtered/resampled MoCap markers written to \texttt{.trc} (converted from \texttt{.c3d}) and a markerless solve driven by filtered/resampled \texttt{SMPL}-derived virtual markers written to \texttt{.trc} (assembled from fixed \texttt{SMPL} surface vertices). Trials were retained for kinematic benchmarking only when \emph{both} sources provided complete trajectories for all required markers over a common time span and IK returned a full coordinate solution without missing values, enabling like-for-like paired comparisons (see \figref{fig:trials_included}).

Marker sets and inclusion/exclusion rules matched those used during scaling for each modality/model configuration (see \stabref{tab:markers_tools}), ensuring that kinematic differences reflect measurement modality rather than altered anatomical constraints. Bony landmarks were up-weighted (\textbf{weight=10}) relative to soft-tissue markers (\textbf{weight=1--5}). For \texttt{SMPL}-driven IK, head markers were excluded due to lower tracking reliability; adipose-region markers were retained but strongly down-weighted (\textbf{weight=0.1}); pelvis/shoulder markers were moderately up-weighted (\textbf{weight=2}) to stabilize global pose; and remaining markers used uniform weights (\textbf{weight=1}) (see \stabref{tab:scale_factors}). For the Rajagopal configuration, upper-limb virtual markers were excluded to match model topology (see \sfigref{fig:smpl_grid_full_raja}).

For kinematic evaluation, \textit{SMPL-driven IK} joint-angle time series were compared with the paired \textit{MoCap-driven IK} solution within the same \texttt{OpenSim} model and coordinate definitions. To mitigate small sampling-rate differences and cumulative phase drift, each trial was trimmed by \textbf{10 }\% at the beginning and end to reduce initiation/termination artifacts (see \figref{fig:kinematic_pipeline}), after which a robust subset of comparable step/cycle segments was extracted using an extrema-based event definition designed for atypical SCI gait and occasional tracking dropouts. Candidate extrema were detected via a three-point test and amplitude-filtered using an interquartile-range criterion (retaining only events exceeding $median + 0.25\,\mathrm{IQR}$ of the candidate distribution) to suppress low-salience oscillations and noise; to avoid clustered detections from high-frequency jitter or multi-peak patterns, we enforced a minimum temporal separation of \textbf{0.75~s}, retaining the most prominent event within a \textbf{5 }\% tolerance when spacing was violated. Consistent with typical sagittal gait signatures, segmentation used local maxima for knee and ankle angles and local minima for hip flexion; centered analysis windows with joint-specific margins were then defined around each retained event (hip $>$ ankle $>$ knee), constructed independently for \texttt{SMPL} and MoCap to avoid cross-source temporal assumptions.

Because the number of retained events can differ between modalities within a trial, segments were paired by index after selecting an integer shift that minimized the median absolute time difference between matched extrema. Paired segments were trimmed to identical durations, exported as per-segment \texttt{.mot} files (time reset to 0), pooled into per-source \texttt{.csv} tables, time-normalised, and temporally aligned using DTW prior to waveform-level scoring.

\subsubsection*{Prediction of Ground Reaction Forces}
For pGRF inference, IK coordinate trajectories were exported, checked for missing values, and resampled to the \textbf{100~Hz} cadence expected by \texttt{GaitDynamics}\cite{tan2025gaitdynamics}. For benchmarking against force plates, we retained only trials with valid IK solutions over the force-plate contact interval, non-zero and clearly segmented force-plate contacts (vertical GRF $>20$~N) for \textbf{two consecutive} steps (\textit{Left--Right}
or \textit{Right--Left}), and sufficient surrounding frames to extract full \textbf{1.5~s} windows centered on each retained step; trials failing any of these checks were excluded from pGRF evaluation (see \figref{fig:trials_included}).

To examine the sensitivity of pGRF to musculoskeletal-model parametrization in our markerless setting, we generated kinematics using two scaled model variants (Pose2Sim and Rajagopal) and performed pGRF inference for both. For each retained trial, \texttt{GaitDynamics} was provided with: the participant-specific scaled \texttt{.osim} model, the corresponding IK kinematics (\texttt{.mot}), participant height and bodyweight, and a walking-speed parameter. Following the default overground configuration in \texttt{GaitDynamics}, we set the walking speed to 0~m/s (the model infers gait speed implicitly from the kinematics rather than prescribing treadmill speed). Predicted forces were exported and synchronized to the corresponding force-plate contact windows for direct comparison to measured GRF and for downstream use of pGRF as the external-load input to inverse dynamics (see \sfigref{fig:kinetic_pipeline}).

Kinetic quantities (joint moments) were computed with \texttt{OpenSim 4.5} inverse dynamics (ID) under two configurations: \textbf{semi-markerless} (\texttt{SMPL}-driven kinematics with \textbf{measured} force-plate GRF) and \textbf{fully markerless} (\texttt{SMPL}-driven kinematics with \textbf{predicted GRF (pGRF)} from \texttt{GaitDynamics}). Semi-markerless ID additionally required CoP-consistent alignment of the reconstructed global translation so that the measured GRF wrench was applied with a mechanically plausible line of action (see \sfigref{fig:kinetic_pipeline} and \figref{fig:osim_pipeline_overview}; trial inclusion/segmentation is summarised in \figref{fig:trials_included}).

\subsection*{Graph-based conditional-dependence analysis}

\subsubsection*{Purpose and feature scope}

The purpose of the graph analysis was to determine whether the markerless gait
descriptors formed clinically interpretable multivariate patterns. This analysis
was performed after waveform benchmarking and used the same segmented gait representation
described above. Therefore, it was not a separate data-acquisition stream.
Instead, it was a secondary analysis of the markerless kinematic and loading-related
descriptors produced by the main pipeline. Joint-moment features were excluded
because the inverse-dynamics analyses showed that these variables, especially proximal
moments, were sensitive to kinematic and external-load inconsistencies.

The graph should be interpreted as a conditional association map. A standard correlation
asks whether two variables vary together or not. A conditional dependence
graph asks a stricter question: Do two variables remain associated after the model
has been adjusted for all other variables in the graph? This distinction is
important for gait data because many waveform descriptors are correlated. For example,
an ankle feature and a clinical score may appear to be correlated only because
both are related to walking speed or pelvic motion. Graph analysis attempts to
separate such indirect relationships from associations that remain after the rest
of the gait feature set is considered.

\subsubsection*{Waveform-to-node representation}

The first step was to convert each time-varying gait waveform into a small
number of scalar descriptors. For each segmented gait waveform $x_{iq}(t)$,
where $i$ indexes the segmented trial, $q$ indexes the gait feature, and
$t\in[0,1]$ denotes normalised gait-cycle time, we used functional principal component
analysis (FPCA). FPCA represents each waveform as a mean trajectory plus a
weighted combination of principal modes of variation,
\begin{equation}
x_{iq}(t) = \mu_{q}(t) + \sum_{k=1}^{K_q}\xi_{iqk}\,\phi_{qk}(t), \label{eq:graph_fpca}
\end{equation}
where $\mu_{q}(t)$ is the mean waveform for feature $q$, $\phi_{qk}(t)$ is the
$k$-th principal mode, and $\xi_{iqk}$ is the FPCA score for trial $i$. In
practical terms, each score $\xi_{iqk}$ summarises how strongly one trial expresses
a specific waveform pattern, such as greater hip-flexion amplitude, altered knee
timing, or a vertical-GRF loading pattern. For the reported graphs, we
retained up to the first five PC scores per waveform feature ($K_{q}\le5$). The
graph nodes therefore consisted of the retained gait-component scores $\{\xi_{iqk}
\}$ together with one clinical target node $m$, such as SCIM mobility Total, AIS grade,
injury category/subcategory, or injury etiology.

Before model fitting, gait-feature scores were median-imputed when necessary
and standardized to zero mean and unit variance. Continuous clinical variables,
including SCIM mobility Total and ordinal AIS grade, were also standardized; AIS grades
were encoded in their clinical order. Categorical clinical variables were encoded
as class labels for logistic or multinomial modelling. Rare categorical levels
were pooled to avoid unstable class estimates, and the categorical targets were
restricted to up to eight of the most frequent classes, with the remaining
levels assigned to an \texttt{Other} category.

\subsubsection*{Sparse graph estimation}

Let $\mathbf{z}_{i} = (z_{i1},\dots,z_{ip})^{\top}$ denote the node vector for
segmented trial $i$. The first $p-1$ entries correspond to gait-feature PC scores,
and the final entry corresponds to the selected clinical target. We estimated a
sparse graph using nodewise $\ell_{1}$-regularised neighbourhood selection.
This procedure fits one regression model per node. For a given node $v$, the
model tries to predict $v$ from all other nodes. The $\ell_{1}$ penalty shrinks
many coefficients exactly to zero, so only a small set of predictors remains
as candidate neighbours of node $v$. Formally, for each node $v$, we solved
\begin{equation}
\widehat{\boldsymbol{\beta}}^{(v)}= \arg\min_{\boldsymbol{\beta}}\left[ \frac{1}{n}
\sum_{i=1}^{n}\ell_{v}\!\left(z_{iv},\, \beta_{0} + \sum_{u\neq v}\beta_{u} z
_{iu}\right) + \lambda_{v} \|\boldsymbol{\beta}\|_{1} \right], \label{eq:graph_neighborhood}
\end{equation}
where $\ell_{v}(\cdot)$ denotes the node-appropriate loss function and $\lambda
_{v}$ controls the degree of sparsity. The sparsity parameter was selected
separately for each node by cross-validation. Continuous nodes, including all
gait-feature PC scores and continuous clinical targets, were modelled using squared-error
loss. Binary clinical targets were modelled using binomial logistic loss, and multiclass
clinical targets were modelled using multinomial logistic loss.

Neighbourhood selection produces two possible coefficients for each pair of nodes:
the coefficient for node $u$ when predicting node $v$, and the coefficient for
node $v$ when predicting node $u$. Because the final graph was intended to be
undirected, we combined these two directions into a single symmetric edge weight,
\begin{equation}
\widehat{w}_{uv}= \max\!\left( \left|\widehat{\beta}^{(v)}_{u}\right|, \left|
\widehat{\beta}^{(u)}_{v}\right| \right), \qquad u\neq v. \label{eq:graph_sym}
\end{equation}
An undirected edge between nodes $u$ and $v$ was therefore interpreted as a
direct conditional association that remained after adjusting for all other
graph nodes. The edge should not be interpreted as evidence that one variable causes
the other.

\subsubsection*{Bootstrap stability}

To reduce dependence on a single fitted graph and to account for repeated
trials from the same participant, we assessed graph stability by subject-level
cluster bootstrap resampling. In each bootstrap replicate, participants,
rather than individual trials, were sampled with replacement. All available
trials from a selected participant were retained together. The graph was then re-estimated
and the presence of each edge was recorded. With $B=300$ bootstrap replicates,
the edge-stability score was defined as
\begin{equation}
\widehat{\pi}_{uv}= \frac{1}{B}\sum_{b=1}^{B}\mathbf{1}\!\left(\widehat{w}^{(b)}
_{uv}>0\right), \label{eq:graph_stability}
\end{equation}
where $\mathbf{1}(\cdot)$ is the indicator function. We retained only edges
whose bootstrap stability exceeded the predefined threshold $\tau=0.70$. Thus,
an edge shown in the graph had to reappear in at least 70\% of the subject-level
bootstrap fits. Edge width in the plotted graphs was proportional to
$\widehat{\pi}_{uv}$, so thicker edges indicate more reproducible conditional associations.

\subsubsection*{Markov blanket extraction and visualisation}

For a given clinical metric node $m$, the Markov blanket was defined as the set
of its direct neighbours in the retained undirected graph,
\begin{equation}
\mathrm{MB}(m) = \{u\neq m : \widehat{\pi}_{um}\ge \tau\}. \label{eq:graph_mb}
\end{equation}
In an undirected graphical model, this set is the immediate neighbourhood that
makes the target conditionally independent of the remaining nodes. For readers
less familiar with graphical models, the Markov blanket can be understood as the
smallest set of gait descriptors that still carries direct information about the
clinical target once all other retained descriptors have been considered. We
therefore used the Markov blanket as the clinically interpretable summary of
the full graph.

The figures display the fully retained graph with the Markov blanket
highlighted. The clinical target is the black square at the center of the ring,
and gait-feature nodes are arranged around it in sectors by domain (GRF, hip, knee,
ankle, pelvis), with blanket members drawn on an inner radius such that proximity
to the center encodes blanket membership. The domain is encoded redundantly by
color (a monotone blue ladder ordered proximal to distal) and by node shape, so
the figures remain readable under color-vision deficiency; the digit inside
each node is the principal-component index. Coral edges identify target-to-blanket
connections, darker gray edges show dependencies among the selected neighbors,
and faint background edges show the remaining stable gait--gait dependencies, limited
to the most stable edges for readability. The edge width is proportional to
the bootstrap stability within each of these three classes; however, the width-to-stability
mapping differs between classes and should not be compared across them. Figures
were prepared at their final printed size, following the laboratory figure
standard. These visualizations are intended to summarize the multivariate association
structure, not to infer causal pathways.

\subsection*{Unsupervised Clustering analysis}
We used an unsupervised strategy to derive \emph{data-driven} locomotor phenotypes
from the multivariate gait feature space. Clustering served two purposes: (i) to
summarize heterogeneous neurological gait in an interpretable manner for
rehabilitation planning, and (ii) to avoid relying on a single summary measure
(most commonly walking speed), which captures \emph{capacity} but only partially
reflects \emph{movement strategy}. Clusters were \textbf{not} defined by
pre-specified numerical cutoffs; instead, they were inferred from the data using
a workflow that explicitly addresses the confounding effect of speed.

\subsubsection*{Feature matrix, scaling, and speed filtering.}
Let $\mathbf{X}\in\mathbb{R}^{n\times p}$ denote the participant-by-feature matrix
(rows = participants, columns = gait features) and let $\mathbf{v}\in\mathbb{R}^{n}$
denote gait speed. Prior to any distance-based method, each feature $j$ was
standardised to a $z$-score,
\begin{equation}
\label{eq:zscore}Z_{ij}=\frac{X_{ij}-\mu_{j}}{\sigma_{j}},
\end{equation}
where $\mu_{j}$ and $\sigma_{j}$ are the across-participant mean and standard deviation
of feature $j$, preventing variables with large physical units from dominating
Euclidean distances.

To quantify speed dependence, we computed the Pearson correlation between each
standardised feature and speed,
\begin{equation}
\label{eq:pearson_clust}r_{j}=\mathrm{corr}(\mathbf{z}_{:j},\mathbf{v})=\frac{\sum_{i=1}^{n}(Z_{ij}-\bar{Z}_{:j})(v_{i}-\bar{v})}{\sqrt{\sum_{i=1}^{n}(Z_{ij}-\bar{Z}_{:j})^{2}}\sqrt{\sum_{i=1}^{n}(v_{i}-\bar{v})^{2}}},
\end{equation}
then defined a speed-related set $\mathcal{S}=\{j:|r_{j}|\ge 0.5\}$ and a residual
(speed-silent) set $\mathcal{R}=\{j:|r_{j}|<0.5\}$. This split operationalises the
clinical observation that speed drives large, largely monotonic changes in many
gait variables.

\subsubsection*{Dimensionality reduction (PCA) and variance retention}
Within each subset ($\mathcal{S}$ and $\mathcal{R}$), we applied PCA to the
standardised feature matrix. Let $\mathbf{T}=\mathbf{Z}\mathbf{W}$ denote the
PC-score matrix (columns are PCs). We retained the smallest number $m$ of PCs such
that the cumulative explained variance exceeded 70\%,
\begin{equation}
\label{eq:pca_70}\frac{\sum_{k=1}^{m}\lambda_{k}}{\sum_{k=1}^{p}\lambda_{k}}\ge 0.70,
\end{equation}
where $\lambda_{k}$ is the eigenvalue associated with PC $k$.

\subsubsection*{Primary cluster analysis: speed strata (K-means)}
Participants were first grouped using the PC scores derived from the speed-related
set $\mathcal{S}$ together with gait speed. K-means partitions the PC-score vectors
$\{\mathbf{t}_{i}\}_{i=1}^{n}$ into $k$ groups by minimising the within-cluster sum
of squares,
\begin{equation}
\label{eq:kmeans_obj}\min_{\{C_1,\dots,C_k\}}\sum_{\ell=1}^{k}\sum_{i\in C_\ell}\lVert\mathbf{t}_{i}-\boldsymbol{\mu}_{\ell}\rVert_{2}^{2},
\end{equation}
where $\boldsymbol{\mu}_{\ell}$ is the centroid of cluster $C_{\ell}$. We used $k=2-6$ and resulted in best configuration of $k=3$; the resulting three groups were ordered by mean gait speed and labelled \textbf{Slow},
\textbf{Medium}, and \textbf{Fast}.

For continuous variables compared across the three primary clusters
(Slow/Medium/Fast), we used the Kruskal--Wallis rank test. Let $x_{ij}$ denote
observations in cluster $j\in\{1,2,3\}$ with sample size $n_{j}$, and let
$N=\sum_{j} n_{j}$. After ranking all values jointly (average ranks for ties), the
Kruskal--Wallis statistic is
\begin{equation}
\label{eq:kruskal_wallis}H = \frac{12}{N(N+1)}\sum_{j=1}^{3}n_{j}\left(\bar{R}_{j}-\frac{N+1}{2}\right)^{2},
\end{equation}
where $\bar{R}_{j}$ is the mean rank in cluster $j$. Under the null hypothesis of
identical distributions, $H$ is approximately $\chi^{2}$-distributed with 2 degrees
of freedom.

For categorical variables, we used Pearson's chi-squared test. With observed counts
$O_{jk}$ and expected counts $E_{jk}$ in cluster $j$ and category $k$, the statistic
is
\begin{equation}
\label{eq:chi_square}\chi^{2} = \sum_{j}\sum_{k}\frac{(O_{jk}-E_{jk})^{2}}{E_{jk}}.
\end{equation}

\subsubsection*{Sub-cluster analysis}

Within each speed stratum, we removed the speed-related variables and clustered
participants using only residual features $\mathcal{R}$. We used Ward's
agglomerative hierarchical clustering, which recursively merges the pair of clusters
that yields the smallest increase in within-cluster variance (equivalently,
minimises the increase in \emph{error sum of squares}) at each merge.

For candidate cuts with $k\in\{2,3,\dots\}$ sub-clusters within a stratum, we
computed silhouette scores based on Euclidean distances in residual-PC space. For
participant $i$, the silhouette value is
\begin{equation}
\label{eq:silhouette}s(i)=\frac{b(i)-a(i)}{\max\{a(i),b(i)\}},
\end{equation}
where $a(i)$ is the mean distance from $i$ to all points in its own cluster and
$b(i)$ is the minimum mean distance from $i$ to points in any other cluster. The
cut maximising the mean silhouette consistently supported \textbf{two} sub-clusters
per speed stratum (\textbf{Sub-cluster~1} and \textbf{Sub-cluster~2}).

To enforce clinically tight phenotypes, we computed the Pearson correlation between
each participant's residual-feature vector $\mathbf{z}^{(\mathcal{R})}_{i}$ and the
corresponding cluster centroid $\bar{\mathbf{z}}^{(\mathcal{R})}_{C(i)}$. Participants
with centroid similarity $<0.6$ were labelled \textit{unclassified} for downstream
descriptive statistics.

Overall, this speed-disentangled design enforces that (i) primary clusters
represent capacity-level walking speed strata, while (ii) sub-clusters reflect
speed-independent gait mechanics (coordination, joint excursions, and loading
patterns) that would otherwise be masked by the global speed effect.

After clusters were assigned, we characterised each group using two complementary
descriptor sets: (i) a \textbf{clinical dataset} (demographics, injury descriptors,
SCIM mobility, LEMS, spasticity, ROM) and (ii) a \textbf{gait dataset}
(spatiotemporal variables, kinematics, gait-quality indices such as GVS/SSD/cyclograms,
and GRF-derived loading descriptors). Cluster labels were treated as grouping
variables; all hypothesis tests were therefore \emph{post hoc descriptive}
comparisons intended to summarise which clinical and biomechanical factors differed
across phenotypes.

Within each speed stratum, continuous variables were compared between Sub-cluster~1
and Sub-cluster~2 using the Mann--Whitney U test. If $R_{1}$ is the sum of ranks in
Sub-cluster~1 (size $n_{1}$) after joint ranking of the two groups (total
$N=n_{1}+n_{2}$), then
\begin{equation}
\label{eq:mann_whitney_u}U_{1} = R_{1} - \frac{n_{1}(n_{1}+1)}{2}, \qquad U_{2} = n_{1}n_{2} - U_{1},
\end{equation}
with the reported $p$-value based on the exact or normal-approximation distribution
depending on sample size and ties. Categorical comparisons within strata again used
\eqhref{eq:chi_square}.

Continuous variables were reported as means (SD), and categorical variables are
reported as percentages. We report two-sided $p$-values and use a nominal
significance threshold of $\alpha=0.05$. Because the purpose of this section is
\emph{phenotype description} rather than confirmatory inference on a single
pre-registered endpoint, we do not interpret individual $p$-values in isolation;
instead, we emphasise consistent, biomechanically coherent patterns across related
metrics (double-support time, GVS/SSD, cyclogram areas, and GRF peaks).

\paragraph*{Cross-speed cluster analysis}
To qualitatively compare phenotype ``fingerprints'' across strata, we generated
cross-speed-combined summaries by placing cluster-aggregated, $z$-scored features
(\eqhref{eq:zscore}) on a shared normalised axis. This layer is intended for
interpretability only (see \sfigref{fig:supp_cross_speed_combined_markerless}--\sfigref{fig:supp_cross_speed_combined_markerbased}).

\section*{Data and Code Availability}
All data were obtained following informed consent by the participants and processed in accordance with approval from the ethics committee of northwest and central Switzerland (EKNZ Project-ID 2022-00935). We release an anonymized version of the dataset alongside this paper. The released dataset consists of \texttt{CameraHMR} re-renders: each sequence is re-rendered using the best-performing \texttt{CameraHMR} model from our benchmark with a black background and a generic \texttt{SMPL} head, and is provided together with the de-identified medical metadata used in this study. The dataset and accompanying scripts are available at \url{https://github.com/SCAI-Lab/SCAI-SCI-Gait-Dataset}.

\section*{Acknowledgments}
We sincerely thank all participants in the \texttt{SPZ} Gait Analysis Study and our colleagues for their valuable contributions to data collection. We also acknowledge Lorena M\"uller (scientific assistant at Swiss Paraplegic Research) and Kathrin Becker (physiotherapist at the Swiss Paraplegic Centre) for their substantial effort and contributions during the early phase of the project. Special thanks are extended to Sabrina Amrein for her assistance with data collection and annotation of recording settings. The project is supported by the AO research opportunity award from the AO and ETH Zurich foundation. This study was partially funded by the Schweizer Paraplegiker Stiftung and the ETH Zürich Foundation (2021-HS-348), within the Digital Transformation in Personalized Healthcare initiative for individuals with spinal cord injury, and the Innosuisse innovation project number: 115.290 IP-ICT.

\section*{Author Contributions}
S.N. and D.P. conceptualised the framework. S.N. developed the human mesh recovery (HMR) models and performed the core data analyses. M.R. carried out the OpenSim-related work (including musculoskeletal modelling and processing). M.R. and S.N. implemented the software pipeline and contributed to data processing. Y.L. performed the graph-based analysis of conditional relationships. M.R. and S.N. wrote the first draft of the manuscript. R.R. and D.P. managed the project. All authors reviewed, modified, and approved the final manuscript.

\section*{Competing Interests}
No competing interest to declare.

\bibliography{sample}

\newpage

\renewcommand{\thefigure}{S\arabic{figure}}
\setcounter{figure}{0}
\renewcommand{\thetable}{S\arabic{table}}
\setcounter{table}{0}

\section*{Supplementary Information}

\subsection*{Comparison to related monocular and multi-camera biomechanics systems}

\begin{table}[H]
\centering
{\small
\setlength{\tabcolsep}{3.5pt}
\renewcommand{\arraystretch}{1.25}

\begin{tabularx}{\linewidth}{>{\bfseries\raggedright\arraybackslash}p{2.9cm} *{5}{>{\raggedright\arraybackslash}X}}
\toprule
& \multicolumn{1}{c}{\textbf{OpenCap}\cite{opencap}}
& \multicolumn{1}{c}{\textbf{\shortstack{OpenCap\\Monocular}}\cite{OpenCapMonocular2026}}
& \multicolumn{1}{c}{\textbf{SAM4Dcap}\cite{sam4dcap}}
& \multicolumn{1}{c}{\textbf{\shortstack{CameraHMR +\\OpenSim gait}}\cite{HORSAK2025112986,patel2024camerahmr}}
& \multicolumn{1}{c}{\textbf{This study}} \\
\midrule
Cameras & 2+ smartphones (synchronized) & 1 smartphone & 1 camera & 1 camera & 1 sagittal view \\
Validation sample size\newline (as reported) & $n=10$ + $n=100$ & $n=10$ & $n=1$ (single-case) & $n=19$ (validity) $+\,n=19$ (reliability) & $\mathbf{n=239}$ (SCI)\newline 49{,}838 frames (46{,}717 valid) \\
Lower-body\newline joint-angle error
& \textbf{MAE:} 4.5$^{\circ}$ (18 rotational DoFs)\newline \textbf{RMSE:} 2.0--10.2$^{\circ}$ (lower-extremity DoFs)
& \textbf{MAE:} 4.8$^{\circ}$ (18 rotational DoFs)\newline \textbf{RMSE:} NR
& \textbf{MAE/RMSE:} NR\newline Peak hip flexion error (drop-jump): $22.2^{\circ}$ (L), $20.6^{\circ}$ (R)
& \textbf{RMSD:} $5.5\pm 1.1^{\circ}$ (validity; trunk + lower limb)\newline \textbf{RMSD:} $3.0\pm 1.0^{\circ}$
& \textbf{MAE:} 3.4--5.3$^{\circ}$ (hip/knee)\newline \textbf{RMSE:} 4.18--6.49$^{\circ}$ \\
Error metric\newline reported & MAE; RMSE & MAE & Peak-angle error & Waveform RMSD & MAE; RMSE \\
Intended use & General (field/clinic) & General/pathology & General & Healthy + simulated pathology & \textbf{Real SCI cohort} \\
Vision / 3D\newline representation & Multi-view 2D keypoints $\rightarrow$ 3D (triangulation) & WHAM\cite{shin2024wham}: world-grounded SMPL\cite{Loper2015} & MHR (monocular; training-free) & CameraHMR\cite{patel2024camerahmr}: perspective-aware SMPL\cite{Loper2015} & \textbf{CameraHMR}\cite{patel2024camerahmr} $+$ \textbf{SMPL}\cite{Loper2015} \\
GRF & Yes$^{\dagger}$ & Yes$^{\dagger}$ & Reported$^{\dagger}$ & No & \textbf{Yes$^{\dagger}$} \\
Clinical phenotyping & Demonstrated & Emerging & No & No & \textbf{Yes} \\
SCI validation & No & No & No & No & \textbf{Yes} \\
\bottomrule
\end{tabularx}

\par\vspace{3pt}

\noindent
\parbox{0.98\linewidth}{
\footnotesize
\textit{Notes:}
MAE = mean absolute error; RMSE = root mean squared error;
RMSD = root mean square deviation; NR = not reported.
Reported errors are reproduced using the metric and scope defined by each source. For OpenCap, the 4.5$^{\circ}$ MAE represents the mean error across 18 rotational degrees of freedom, whereas the reported 2.0--10.2$^{\circ}$ RMSE range corresponds to lower-extremity degrees of freedom. OpenCap Monocular reports a 4.8$^{\circ}$ MAE across its 18 rotational degrees of freedom; a lower-body-only RMSE was not reported. CameraHMR reports waveform RMSD rather than MAE or RMSE, and therefore RMSD is retained here rather than relabeled. Metrics are consequently not directly interchangeable across studies. Differences in activities, anatomical degrees of freedom, reference systems, participant populations, and error definitions should be considered when interpreting the reported values. $^{\dagger}$GRF indicates ground-reaction-force estimation without force plates, i.e., inferred from kinematics using learned and/or physics-based approaches.
}
}
\caption{\small
\textbf{Comparison of representative markerless biomechanics systems.} The comparison summarizes capture requirements, validation cohorts, lower-body joint-angle accuracy, three-dimensional representation, and availability of force and kinetic estimates. Error metrics are reported using the definitions provided by the respective studies; therefore, numerical values should be interpreted as indicative rather
than as directly comparable performance measures.
}
\label{tab:system_comparison}
\end{table}

\hyperref[tab:system_comparison]{Supplementary Table~\ref{tab:system_comparison}} summarizes key design choices and claimed capabilities of representative markerless biomechanics pipelines, spanning multi-camera keypoint-based approaches (e.g., OpenCap) and emerging single-camera mesh-based approaches (e.g., WHAM/OpenCap Monocular, SAM4Dcap, and CameraHMR-derived workflows). The \emph{Cameras} and \emph{Vision / 3D representation} rows distinguish whether 3D motion is obtained by multi-view triangulation of 2D keypoints versus monocular recovery of a full-body parametric mesh (SMPL), which can improve anatomical plausibility under occlusion but still inherits single-view depth ambiguity. The \emph{Participants (reported validation)} row provides the sample sizes reported in the corresponding primary validation studies (when available), highlighting that many systems are validated predominantly in healthy participants and/or simulated pathological gaits rather than neurological cohorts. The \emph{GRF} and \emph{Joint kinetics} rows indicate whether a pipeline reports force-plate-free ground-reaction-force estimation and derived net joint moments (inverse dynamics outputs); as noted in the table footnote, ``GRF'' here does not imply instrumented force plates and different systems use different learned and/or physics-based constraints. Finally, the \emph{Clinical phenotyping} and \emph{SCI validation} rows summarize whether outputs have been used for clinically oriented subgrouping/characterization and whether results have been explicitly evaluated in spinal cord injury (SCI). In this context, the present study uniquely targets a real SCI cohort from a single sagittal view, performs clinically oriented phenotyping, and explicitly characterizes failure modes in joint-moment estimates rather than reporting aggregate accuracy alone.

\subsection*{SMPL-constrained mesh recovery and reduced \textit{PA-MPJPE} under occlusion}

\begin{figure}[H]
  \centering
  \includegraphics[width=1\linewidth]{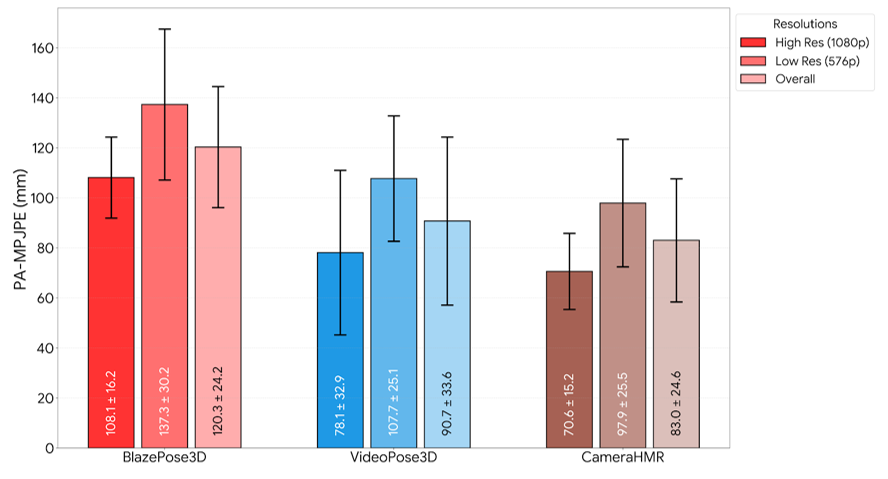}%
  \caption{\small{\textbf{PA-MPJPE comparison across pose-only and \texttt{SMPL}-based mesh recovery methods.} \textit{PA-MPJPE} is reported for two representative pose-only pipelines (\texttt{BlazePose3D} and \texttt{VideoPose3D}) and an \texttt{SMPL}-based HMR method (\texttt{CameraHMR}). \texttt{CameraHMR} achieves the lowest error, consistent with \texttt{SMPL} acting as a structural prior that couples joints through a coherent body model (fixed limb lengths, constrained joint rotations, and temporally stable surface correspondences), thereby reducing implausible configurations and error spikes that can occur in pose-only reconstructions under occlusion, truncation, or motion blur.}}
  \label{fig:supp_pampjpe_resolution}
\end{figure}

\sfigref{fig:supp_pampjpe_resolution} contrasts pose-only 3D estimators (\texttt{BlazePose3D} and \texttt{VideoPose3D}) with an \texttt{SMPL}-based human-mesh-recovery (HMR) method (\texttt{CameraHMR}) on the \texttt{SCAI Gait} dataset. Although all methods operate on the same monocular recordings, the mesh-recovery approach achieves consistently lower \textit{PA-MPJPE}, particularly when image quality degrades (low resolution, motion blur) and when gait-relevant joints are intermittently occluded by assistive devices, clothing, or self-occlusion during swing.

A key reason for this robustness is that \texttt{SMPL}-based HMR does not predict each joint independently. Instead, it optimizes (or regresses) a low-dimensional set of pose and shape parameters that must correspond to a physically plausible human body. This acts as a strong structural prior: limb lengths remain consistent over time, joint rotations are constrained to anatomically feasible ranges, and left--right symmetry and global body proportions are implicitly regularized by the \texttt{SMPL} parameterization. As a consequence, when the image evidence for a distal joint (e.g., the ankle or toe) becomes unreliable, the estimate can be stabilized by the rest of the kinematic chain and by the learned pose manifold, yielding a more coherent lower-body geometry after Procrustes alignment.

In contrast, keypoint-based pipelines often behave like a collection of local detectors followed by a lifting step. When some joints are missing or corrupted, the lifting model may still produce a numerically valid 3D output, but the inferred configuration can drift toward implausible limb placements (e.g., unrealistically large knee flexion excursions, abrupt inter-frame depth flips, or transient limb-length changes). These failure modes may be visually subtle in 2D overlays but become prominent in 3D, where inconsistent inter-joint distances and joint-angle discontinuities directly increase \textit{PA-MPJPE}. This effect is amplified in clinical gait because atypical postures, compensatory trunk lean, and assistive-device interactions increase the likelihood of partial visibility and introduce appearance patterns that differ from the distribution of many generic in-the-wild pose datasets.

From a gait biomechanics perspective, \texttt{SMPL} constraints are particularly beneficial because they promote temporally stable segment orientations and smoother joint trajectories across stance--swing transitions. Even when the swing leg is partially occluded (e.g., by the contralateral limb or a walking aid), the mesh-recovery model tends to preserve a globally consistent pelvis and femur configuration and infer distal pose in a manner that remains compatible with the overall body motion. This regularization helps prevent ``abnormal'' or non-physiological articulations (such as sudden ankle inversions or knee hyperflexion spikes) that would otherwise degrade downstream gait-cycle segmentation, temporal alignment, and musculoskeletal inverse kinematics. Empirically, this manifests as lower \textit{PA-MPJPE} and, qualitatively, as a cleaner and more interpretable gait pattern with fewer implausible outliers.

Finally, it is important to emphasize what \textit{PA-MPJPE} captures (and does not capture). Because Procrustes alignment removes global translation, rotation, and scale, improvements in \textit{PA-MPJPE} primarily indicate better \emph{relative} pose geometry---i.e., more accurate inter-joint relationships and more coherent limb configurations. This is precisely where \texttt{SMPL} provides leverage: the body model couples joints through a shared kinematic and shape representation, which can ``fill in'' missing evidence and maintain plausible gait mechanics under occlusion. For this reason, \texttt{SMPL}-constrained mesh recovery forms a particularly strong foundation for monocular clinical gait analysis, where robustness and physiological plausibility are as critical as raw pointwise accuracy.

\subsection*{Virtual Biomarker Extraction \& Biomechanical Model Development}

To convert monocular videos into biomechanically interpretable time-series, we used an \texttt{SMPL}-constrained reconstruction--to--\texttt{OpenSim} workflow consistent with the main manuscript. In this pipeline, the primary interface between human mesh recovery and musculoskeletal modelling is a marker representation: each reconstructed \texttt{SMPL} mesh is converted into a set of anatomically motivated \emph{virtual markers} defined as fixed mesh vertices and/or landmark points in the \texttt{SMPL} template space. Because these template locations are constant, the same marker index always corresponds to the same anatomical region, providing temporally stable marker correspondence across the entire sequence. Exporting these per-frame marker coordinates yields a standard OpenSim marker trajectory time series, enabling the markerless output to be processed using the same tools used for optical motion capture.

\begin{table}[H]
\centering

\begin{minipage}[t]{0.56\linewidth}
  \vspace{0pt}
  \centering
  \begin{subtable}[t]{\linewidth}
    \centering
    {\fontsize{6}{7}\selectfont
    \setlength{\tabcolsep}{3pt}
    \renewcommand{\arraystretch}{1.1}
    \resizebox{\linewidth}{!}{%
    \begin{tabular}{lll}
    \toprule
    \textbf{Body Part} & \textbf{MoCap Marker Pairs} & \textbf{SMPL Marker Pairs} \\
    \midrule
    Pelvis     & All Hip Markers                 & All Hip Markers \\
    Sacrum     & All Hip Markers                 & All Hip Markers \\
    Femur      & Front Hip, Knee                 & Front Hip, Knee \\
    Patella    & Front Hip, Knee                 & Front Hip, Knee \\
    Tibia      & Knee, Ankle                     & Knee, Ankle \\
    Talus      & 0.85                            & 0.85 \\
    Calcaneus  & 1                               & 1 \\
    Toes       & 0.85                            & 0.85 \\
    Head       & 0.8                             & 0.8 \\
    Abdomen    & 1                               & 1 \\
    Lumbar     & 1                               & Shoulder, Front Hip \\
    Torso      & 0.8                             & Shoulder, Front Hip \\
    Humerus    & 1                               & Shoulder, Elbow \\
    Ulna       & 1                               & Elbow, Wrist \\
    Radius     & 1                               & Elbow, Wrist \\
    Hand       & 1                               & Elbow, Wrist \\
    \bottomrule
    \end{tabular}%
    }
    }
    \caption{\small{\textbf{Segment scaling definitions for the \texttt{Scale} tool.} Marker pairs denote bilateral (left/right) landmarks; scalar entries indicate a fixed multiplicative scale factor.}}
    \label{tab:scale_factors}
  \end{subtable}
\end{minipage}\hfill
\begin{minipage}[t]{0.41\linewidth}
  \vspace{0pt}
  \centering
  \begin{subtable}[t]{\linewidth}
    \centering
    {\scriptsize
    \setlength{\tabcolsep}{3pt}
    \renewcommand{\arraystretch}{1.1}
    \resizebox{\linewidth}{!}{%
    \begin{tabular}{llll}
    \toprule
    \textbf{Tool} & \textbf{Data Source} & \textbf{Total} & \textbf{Used} \\
    \midrule
    \multirow{3}{*}{\texttt{Scale}}
    & MoCap                             & 20  & 18 \\
    & \texttt{SMPL} (\texttt{Pose2Sim})  & 105 & 68 \\
    & \texttt{SMPL} (Rajagopal)          & 77  & 50 \\
    \cmidrule(lr){1-4}
    \multirow{3}{*}{\texttt{IK}}
    & MoCap                             & 20  & 18 \\
    & \texttt{SMPL} (\texttt{Pose2Sim})  & 105 & 96 \\
    & \texttt{SMPL} (Rajagopal)          & 77  & 70 \\
    \bottomrule
    \end{tabular}%
    }
    }
    \caption{\small{\textbf{Marker availability and selection in \texttt{OpenSim}.} Total markers in each source versus the subset used for \texttt{Scale} and inverse kinematics (IK).}}
    \label{tab:markers_tools}
  \end{subtable}

  \vspace{0.7em}

  \begin{subtable}[t]{\linewidth}
    \centering
    {\fontsize{9}{10}\selectfont
    \setlength{\tabcolsep}{6pt}
    \renewcommand{\arraystretch}{1.15}
    \begin{tabular}{lll}
    \toprule
    \textbf{Eval.} & \textbf{Win.} & \textbf{Thresh.} \\
    \midrule
    Kinematic                 & 0.7, 0.7, 0.3 & 1.5 \\
    Kinetic (Semi-Markerless) & 0.3           & 1.5 \\
    Predicted GRF             & 0.6           & 1.5 \\
    Kinetic (Markerless)      & 0.3           & 1.5 \\
    \bottomrule
    \end{tabular}
    }
    \caption{\small{\textbf{DTW temporal-alignment hyperparameters.} Window and threshold factors (fractions of series length) used in evaluation; joint-specific values are ordered hip/knee/ankle.}}
    \label{tab:dtw_settings}
  \end{subtable}
\end{minipage}

\caption{\small{\textbf{Settings for \texttt{SMPL}-to-\texttt{OpenSim} processing and evaluation.} (a) Segment-level scaling definitions used to personalize the musculoskeletal model from either physical MoCap markers (MB-MoCap) or \texttt{SMPL}-derived virtual markers. Marker pairs indicate bilateral landmarks used to infer segment lengths, whereas scalar entries represent fixed multiplicative factors applied when direct marker-pair definitions were not used. (b) Marker-set size and the subset retained for the \texttt{OpenSim} \texttt{Scale} and inverse-kinematics (IK) tools, reported for each data source and model variant (\texttt{Pose2Sim} and Rajagopal). (c) Dynamic time warping (DTW) alignment parameters used to compare joint-angle waveforms across gait cycles: the Sakoe--Chiba bandwidth (window factor) and heuristic pruning threshold (threshold factor), expressed as fractions of the time-series length; where three values are given, they correspond to hip, knee, and ankle joints (in that order).}}
\label{tab:scaling_marker_settings}
\end{table}

\begin{figure}[H]
  \centering
  \begin{subfigure}[t]{0.49\textwidth}
    \centering
    \begin{tabular}{cc}
      \includegraphics[width=0.48\textwidth]{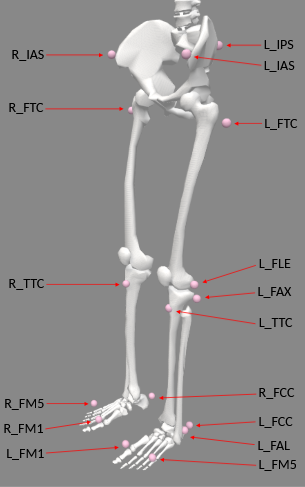} & \includegraphics[width=0.48\textwidth]{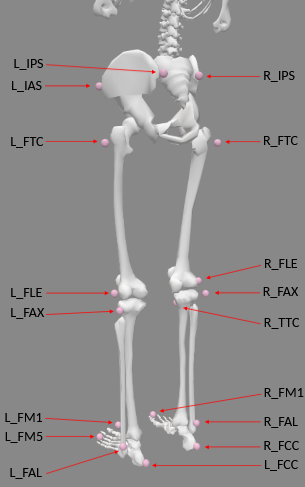} \\[-2pt]
      \includegraphics[width=0.48\textwidth]{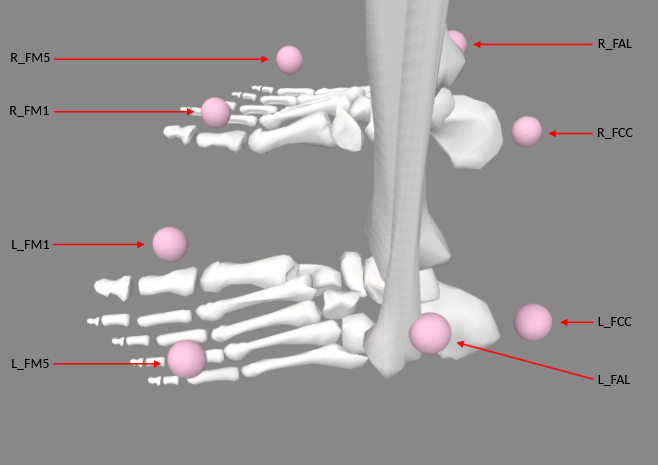}   & \includegraphics[width=0.48\textwidth]{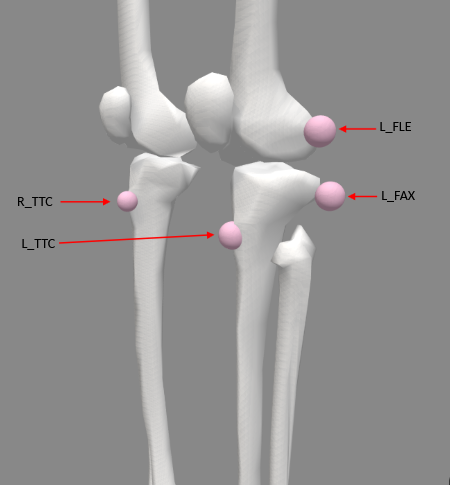} \\
    \end{tabular}
    \subcaption{\small{\textbf{MoCap markerset (lower body): reference clinical placement.} Whole-body anterior/posterior views (top row) and distal close-ups (bottom row) showing the physical marker distribution used to scale the musculoskeletal model and to constrain inverse kinematics, with emphasis on pelvis, shank, knee-line, and foot markers that define segment lengths and distal coordinate frames.}}
    \label{fig:mocap_markerset}
  \end{subfigure}
  \hfill
  \begin{subfigure}[t]{0.49\textwidth}
    \centering
    \setlength{\tabcolsep}{2pt}
    \begin{tabular}{cc}
      \includegraphics[width=0.48\textwidth]{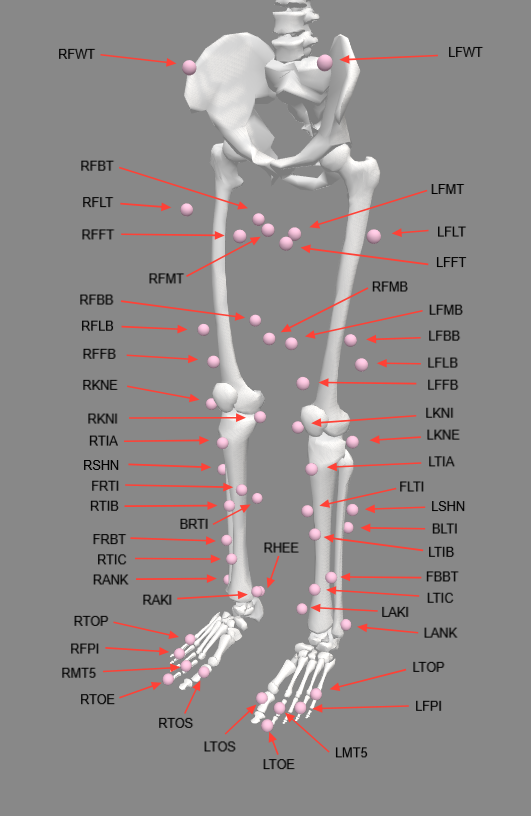} &
      \includegraphics[width=0.48\textwidth]{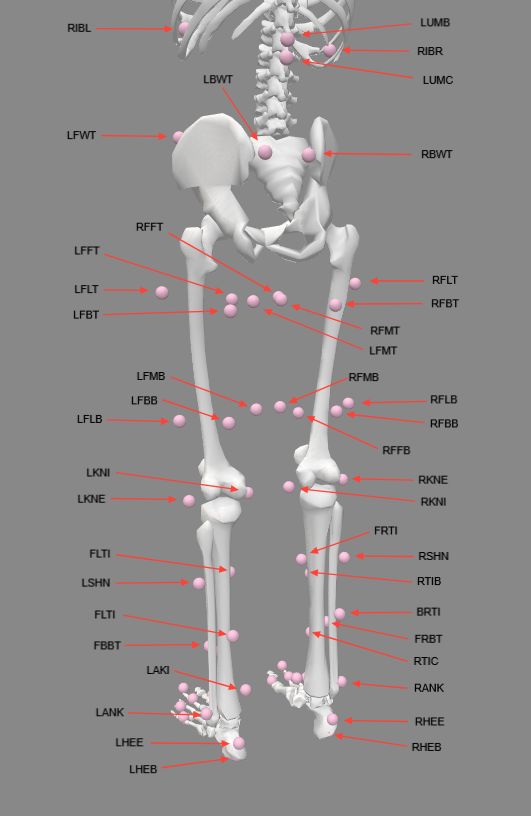} \\[4pt]
      \includegraphics[width=0.48\textwidth]{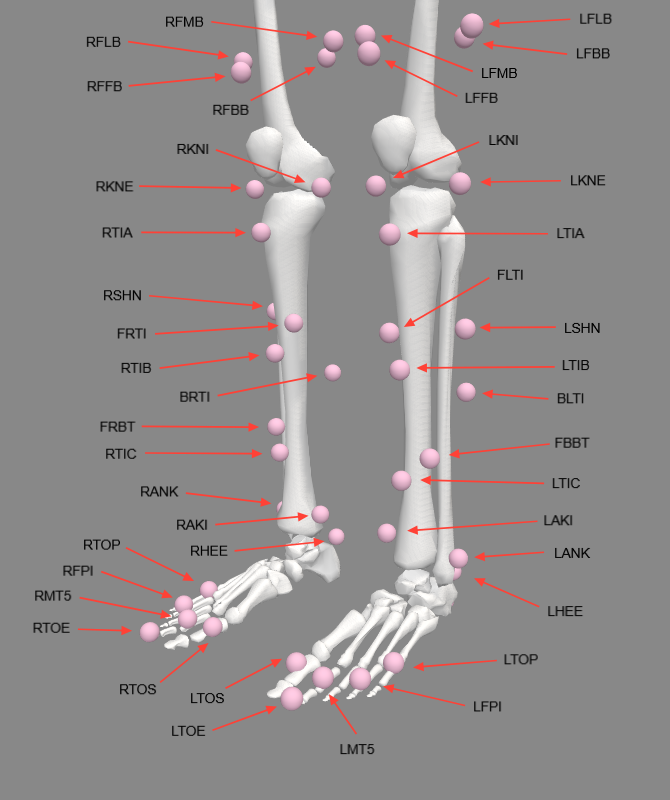} &
      \includegraphics[width=0.48\textwidth]{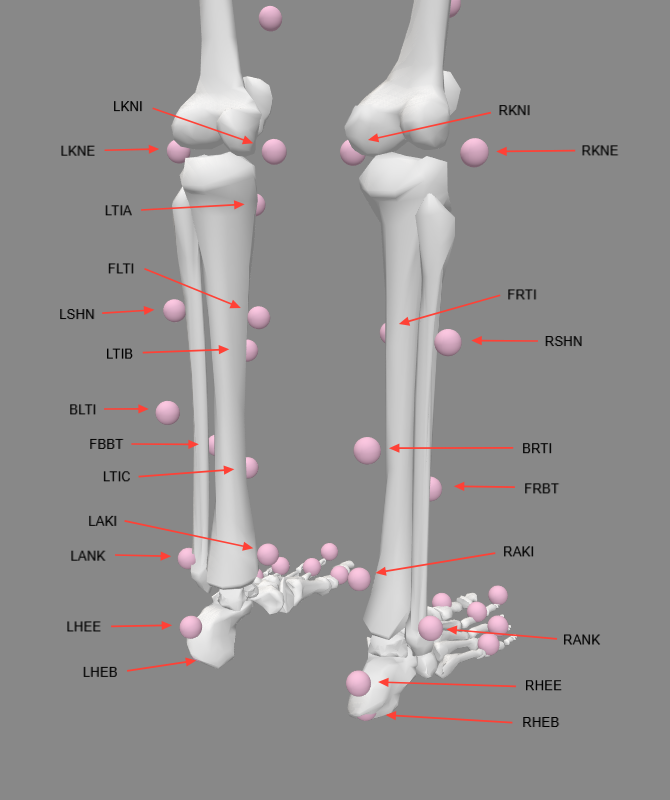} \\
    \end{tabular}
    \subcaption{\small{\textbf{\texttt{SMPL} virtual markerset (\texttt{Pose2Sim}, lower body): template-anchored correspondences.} Virtual markers sampled from fixed \texttt{SMPL} surface vertices over the pelvis, thighs, shanks, and feet, including distal close-ups used to stabilise ankle/foot and knee constraints during scaling and IK (temporally consistent across frames by construction).}}
    \label{fig:smpl_grid_parts}
  \end{subfigure}

  \caption{\small{\textbf{Lower-body marker sets used for scaling and inverse kinematics.} (a) MoCap lower-body marker set used in the clinical reference pipeline. (b) \texttt{SMPL}-derived virtual markers used for the \texttt{Pose2Sim} markerless pipeline (lower-body views).}}
\end{figure}

The physical reference marker placement is illustrated in \sfigref{fig:mocap_markerset}. The whole-body front and rear views show how the marker set spans the pelvis, thighs, shanks, and feet to provide constraints on segment position and orientation. The pelvis/hip-region markers define the pelvic coordinate frame and provide robust bilateral landmarks for scaling. The thigh and shank markers provide additional constraints for femur and tibia tracking, helping to resolve segment rotations during gait when distal landmarks may be partially occluded. The close-up panels of the knee and foot clarify how distal markers cluster around the knee joint line and around the ankle/foot complex, which is important for defining tibial length, ankle location, and the forefoot--hindfoot orientation used by the musculoskeletal model.

\begin{figure}[H]
  \centering
  \begin{subfigure}[t]{0.49\textwidth}
    \centering
    \setlength{\tabcolsep}{2pt}
    \begin{tabular}{c}
      \includegraphics[width=\textwidth]{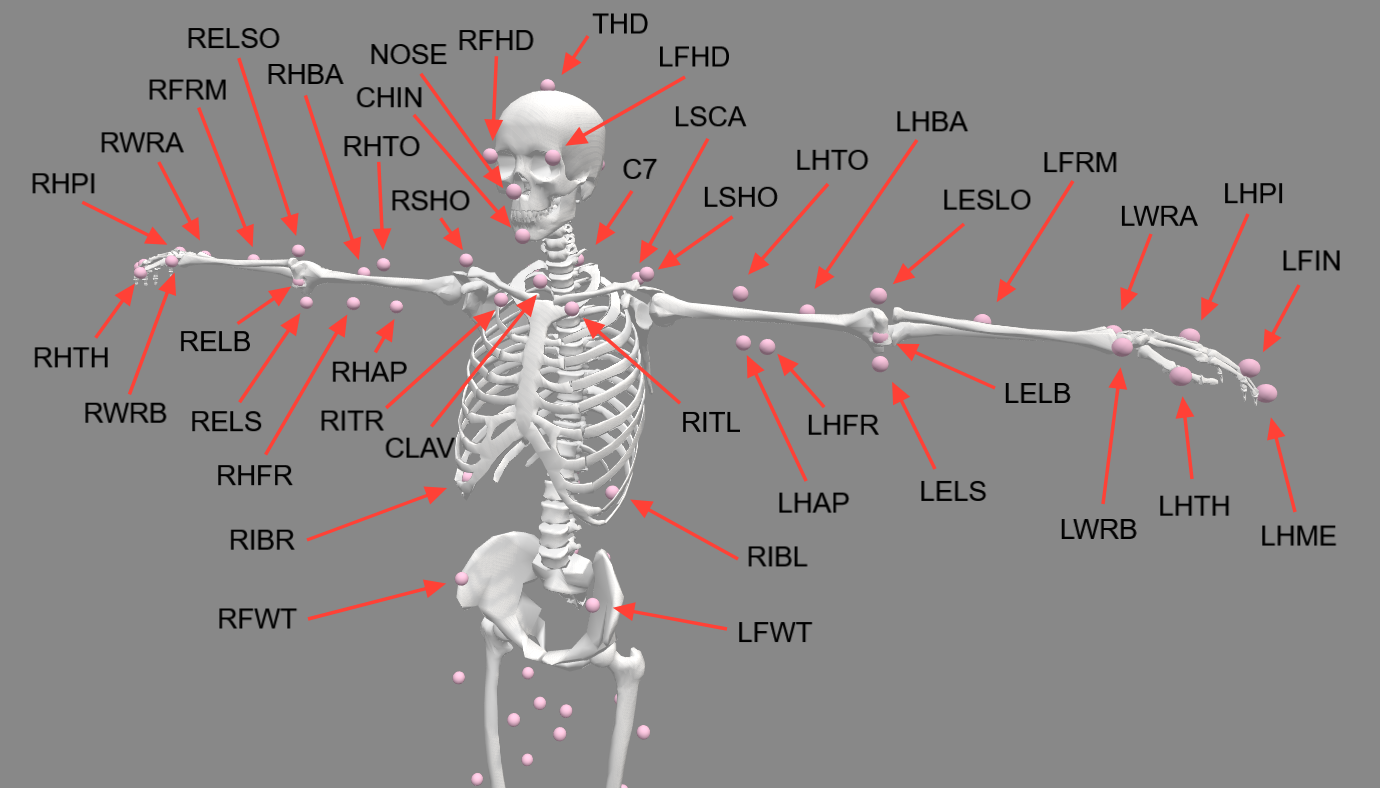} \\[2pt]
      \includegraphics[width=\textwidth]{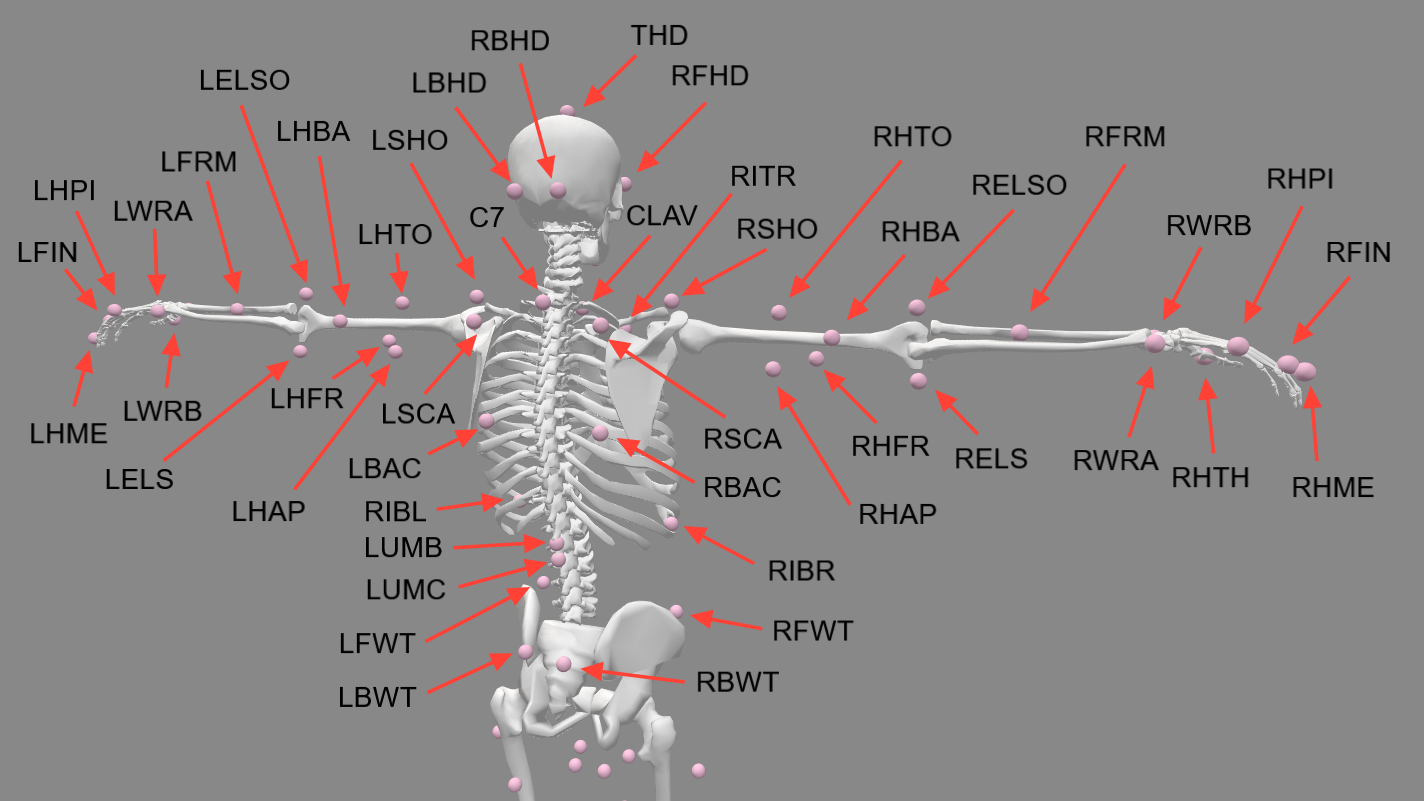} \\
    \end{tabular}
    \subcaption{\small{\textbf{\texttt{SMPL} virtual markerset (\texttt{Pose2Sim}, upper body): trunk anchoring for scaling/IK.} Dense shoulder, thorax, and pelvis-adjacent virtual markers that provide long-lever constraints to stabilise torso orientation and the pelvis--trunk relationship during scaling and inverse kinematics.}}
    \label{fig:smpl_grid_full_p2s}
  \end{subfigure}
  \hfill
  \begin{subfigure}[t]{0.49\textwidth}
    \centering
    \setlength{\tabcolsep}{2pt}
    \begin{tabular}{cc}
      \includegraphics[width=0.48\textwidth]{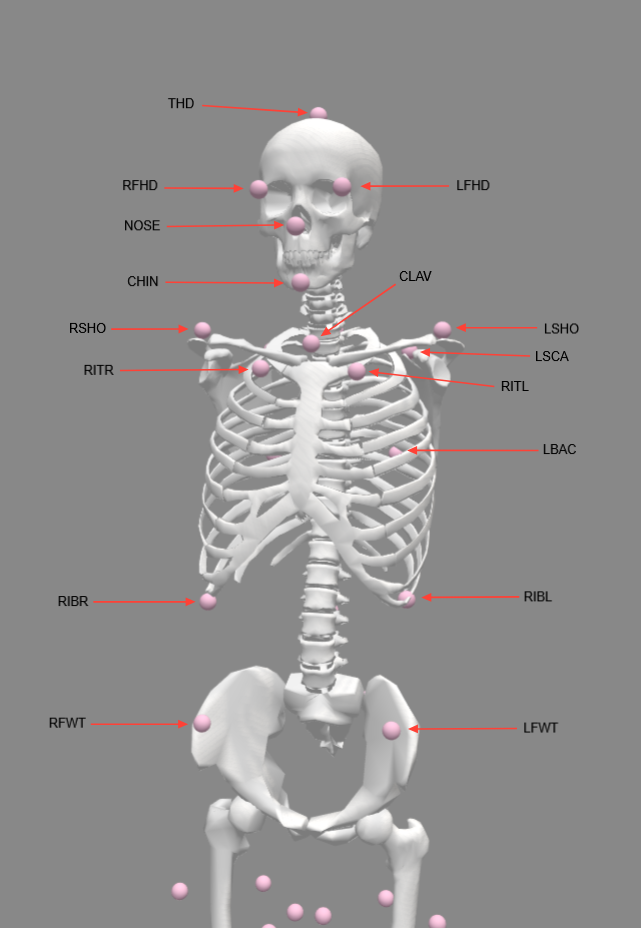} &
      \includegraphics[width=0.48\textwidth]{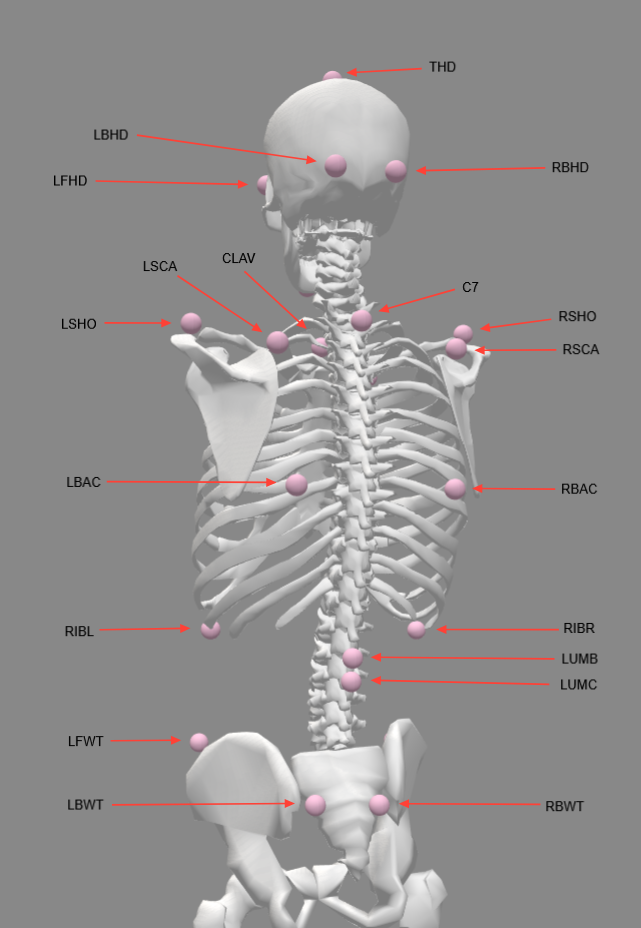} \\
    \end{tabular}
    \subcaption{\small{\textbf{\texttt{SMPL} virtual markerset (Rajagopal, upper body): reduced set matching model topology.} Upper-body virtual markers retained for the armless Rajagopal variant (as used by \texttt{GaitDynamics}), providing sufficient trunk/pelvis anchoring while maintaining compatibility with the simplified segment set.}}
    \label{fig:smpl_grid_full_raja}
  \end{subfigure}

  \caption{\small{\textbf{Upper-body marker sets used for markerless scaling and model stabilisation.} (a) \texttt{SMPL}-derived upper-body markers for the \texttt{Pose2Sim} model. (b) \texttt{SMPL}-derived upper-body markers for the Rajagopal model variant.}}
\end{figure}

The corresponding \texttt{SMPL}-derived virtual marker placements are shown for the \texttt{Pose2Sim} configuration in \sfigref{fig:smpl_grid_full_p2s} and \sfigref{fig:smpl_grid_parts}. The upper-body views highlight trunk and shoulder-region markers that complement the lower-limb set by providing additional long-lever constraints for torso scaling and by stabilizing the pelvis--trunk relationship. The lower-body views show virtual markers distributed across the pelvis, thighs, shanks, and feet, with close-ups that make the intended correspondence to common gait-laboratory landmarks explicit. In particular, the distal foot and knee close-ups help interpret how the virtual marker cloud samples the ankle/foot region and the knee region, which directly affects segment-length estimation during scaling and the quality of distal kinematic constraints.

For the Rajagopal configuration, the virtual-marker set shown in \sfigref{fig:smpl_grid_full_raja} reflects the reduced upper-body representation used in that model variant. The retained trunk/shoulder markers provide sufficient upper-body anchoring for scaling and inverse kinematics, whereas the omission of upper-limb markers ensures compatibility with the model topology and avoids introducing unconstrained degrees of freedom from segments that are not represented.

\stabref{tab:scaling_marker_settings} describes how the marker sets in \sfigref{fig:mocap_markerset} and \sfigref{fig:smpl_grid_parts}--\sfigref{fig:smpl_grid_full_raja} are operationalised in OpenSim. It specifies which segment lengths are inferred from bilateral marker-pair distances and which are assigned fixed multiplicative scale factors, and it reports which markers are available versus which markers are actually retained for the OpenSim \texttt{Scale Tool} and inverse kinematics. This distinction is important because the retained subset determines which anatomical regions contribute constraints during model personalization and kinematic fitting. The table also records the dynamic time warping settings used later when comparing marker-driven outputs across steps, allowing the full virtual-biomarker pipeline to be traced from marker definition, through scaling, to quantitative evaluation. After virtual-marker extraction, trajectories were filtered and resampled to match the marker-based acquisition where needed, and checked for consistent units and coordinate conventions prior to scaling and IK (see \figref{fig:osim_pipeline_overview}).

Biomechanical model development followed standard \texttt{OpenSim 4.5} practice and was applied identically to marker-based and \texttt{SMPL}-based inputs to enable a like-for-like comparison of the results. Guided by the workflow overviews in \figref{fig:osim_pipeline_overview} and \figref{fig:scaling}, we evaluated two musculoskeletal model configurations (\texttt{Pose2Sim} and Rajagopal) under a shared scaling and inverse-kinematics (IK) protocol. Subject-specific scaling used the same marker-pair definitions and segment-wise scale factors for both data sources (\stabref{tab:scaling_marker_settings}), and identical marker-selection rules were applied in \texttt{OpenSim} \texttt{Scale} and IK to ensure that differences arose from the input modality rather than model-setting discrepancies (\stabref{tab:scaling_marker_settings}b). After scaling, IK was solved from either physical MoCap marker trajectories or \texttt{SMPL}-derived virtual-marker trajectories (\figref{fig:scaling}), yielding sagittal-plane joint-angle time series (bilateral hip flexion, knee angle, and ankle angle) that served as the primary \emph{virtual biomarkers} for subsequent agreement analysis.

\subsection*{Evaluation of Markerless Kinematics \& pGRF vs. MoCap \& GRF plate
Time-Series}
To validate the similarity of the \texttt{SMPL}-derived segmented time series signatures,
we evaluated them against equivalently segmented ground truth data. Comparisons
were organized into three conditions: \textit{marker-based}, \textit{semi-markerless},
and \textit{markerless}.
\figref{fig:trials_included} provides a
comprehensive overview of the trials included in each evaluation.

For kinematics, joint angles obtained from \texttt{SMPL}-driven (markerless)
kinematics were compared against MoCap joint angles (marker-based). For kinetics,
joint moments obtained from MoCap with measured GRF were treated as ground
truth and compared against semi-markerless solutions (markerless with measured
GRF) and fully markerless solutions (markerless with pGRF). This evaluation framework
was intended to verify that segmentation preserved salient gait features and waveform
signatures relevant for gait analysis, while attenuating noise and artifacts
related to the camera coordinate frame.

The similarity between the ground truth and (semi-)markerless time series was
quantified using \texttt{DTW} to handle residual phase shifts and timing misalignments
between segments. \texttt{Pearson correlation coefficient} was used to characterise
the linear association between biomechanical estimates and frame-wise \texttt{RMSE/MAE}
to evaluate amplitude agreement. To distinguish statistically meaningful time-continuous
waveform differences from potential oversensitivity of scalar distance metrics
to small offsets, we additionally used \texttt{Statistical Parametric Mapping
(SPM1D)} over the gait cycle together with \texttt{Cohen's $d$} effect-size
estimation (see \hyperref[spm_t_eqn]{Equations~\ref*{spm_t_eqn}--\ref*{cohend_eqn}}).

\subsubsection*{Dynamic Time Warping}
\texttt{DTW} is an established method for aligning temporal sequences that exhibit
variability in local timing or speed, a common challenge in gait analysis. In this
work, \texttt{DTW} was applied to determine an optimal, potentially non-linear,
temporal alignment between two time series by minimizing the cumulative distance
over paired samples. The \texttt{DTW distance} between two time series $X = (x_{1}
, x_{2}, \dots, x_{n})$ and $Y = (y_{1}, y_{2}, \dots, y_{m})$ was given by:
\begin{equation}
\label{dtw_eqn}D_{\text{DTW}}(X, Y) = \min_{\pi}\sum_{(i,j) \in \pi}d(x_{i},
y_{j})
\end{equation}

where $\pi$ denotes a warping path that specifies the alignment between the indices
of the time series $X$ and $Y$. The term $d(x_{i}, y_{j})$ denotes the local
distance between frames $x_{i}$ and $y_{j}$ (e.g., the Euclidean distance). A lower
\texttt{DTW distance} corresponds to higher waveform similarity between two
time series.

Before applying \texttt{DTW}, each segment was \texttt{z-scored} by subtracting
its mean and dividing by its standard deviation. This normalization ensured
that we primarily captured differences in waveform signature and timing, rather
than differences in absolute magnitude or offset.

\subsubsection*{Pearson Correlation Analysis}
\texttt{Pearson correlation analysis} was used to quantify the linear
association between the \textit{marker-based} and the potentially \texttt{DTW
aligned} \textit{(semi-)markerless} time series. The Pearson correlation
coefficient $r$ was defined as:
\begin{equation}
\label{pearson_eqn}r = \frac{\sum_{i=1}^{n}(x_{i} - \bar{x})(y_{i} - \bar{y})}{\sqrt{\sum_{i=1}^{n}(x_{i}
- \bar{x})^{2} \sum_{i=1}^{n}(y_{i} - \bar{y})^{2}}}
\end{equation}

where $x_{i}$ and $y_{i}$ denote, respectively, the \textit{marker-based} and \textit{(semi-)markerless}
values at frame $i$ for a given biomechanical variable. The terms $\bar{x}$ and
$\bar{y}$ denote the corresponding sample means over the segments. The
coefficient satisfies $r \in [-1, 1]$, with values close to $1$ or $-1$ indicating
a strong positive or negative linear relationship, respectively, and values near
$0$ indicating a weak linear association.

\subsubsection*{Error Quantification}
Beyond waveform similarity, we evaluated absolute magnitude agreement using
both \texttt{Root Mean Square Error (RMSE)} and \texttt{Mean Absolute Error (MAE)},
which provide complementary scalar summaries of frame-wise amplitude differences
between the \textit{marker-based} ground truth and the potentially \texttt{DTW
aligned} \textit{(semi-)markerless} segments. For two time series $\{x_{i}\}_{i=1}
^{N}$ (marker-based) and $\{\hat{x}_{i}\}_{i=1}^{N}$ ((semi-)markerless), RMSE
and MAE were defined as:
\begin{equation}
\label{rmse_eqn}RMSE = \sqrt{\frac{1}{N}\sum_{i=1}^{N}(x_{i} - \hat{x}_{i})^{2}}
\end{equation}
\begin{equation}
\label{mae_eqn}MAE = \frac{1}{N}\sum_{i=1}^{N}\left|x_{i} - \hat{x}_{i}\right
|
\end{equation}

Smaller RMSE/MAE values indicate closer amplitude agreement (with RMSE penalizing
larger deviations more strongly).

Because baseline offsets can dominate amplitude errors while leaving waveform
shape largely intact, we also report the \texttt{static offset} as the segment-wise
mean signed difference (reported as mean $\pm$ SD across segments):
\begin{equation}
\label{offset_eqn}\Delta_{\mathrm{offset}}= \frac{1}{N}\sum_{i=1}^{N}(\hat{x}
_{i} - x_{i}).
\end{equation}

\subsubsection*{Waveform and warping distances}
In addition to pointwise errors and correlation, we report two complementary families
of sequence distances: (i) \textbf{DTW-based warping costs}, which explicitly
quantify the cost of temporally aligning two sequences, and (ii) \textbf{MSM
distances}, which quantify waveform dissimilarity while penalising local insertions/deletions.

\paragraph*{DTW (alignment cost and warping effort).}
For each segment pair, we compute the \texttt{DTW distance} $D_{\mathrm{DTW}}$
as the cumulative local cost along the optimal monotone warping path $\pi$ (Equation~\ref{dtw_eqn}).
Because absolute DTW costs depend on the variable and cohort variability, we
additionally report a within-set standardisation to facilitate comparison
across joints/conditions:
\begin{equation}
\label{dtwz_eqn}z_{\mathrm{DTW}}= \frac{D_{\mathrm{DTW}}- \mu_{D}}{\sigma_{D}}
,
\end{equation}
where $\mu_{D}$ and $\sigma_{D}$ are the mean and standard deviation of DTW distances
over the relevant evaluated set of segment pairs.

To characterise \emph{how much} non-linear time deformation DTW required (as reported
in the results as \textit{Warp eff.}), we compute \texttt{warping efficiency} as
the proportion of diagonal (one-to-one) steps along the optimal path $\pi$:
\begin{equation}
\label{warpeff_eqn}\eta_{\mathrm{warp}}= \frac{n_{\mathrm{diag}}}{|\pi|},
\end{equation}
where $|\pi|$ is the path length and $n_{\mathrm{diag}}$ is the number of
diagonal moves. Values closer to 1 indicate that the two sequences were already
closely phase-matched (DTW makes mostly one-to-one matches), whereas lower
values indicate heavier many-to-one / one-to-many warping.

\paragraph*{MSM (shape distance with split/merge penalties).}
We report the \texttt{Move--Split--Merge (MSM)} distance as a complementary
waveform distance that is less explicitly tied to timing warps than DTW. MSM is
defined via the dynamic-programming recursion
\begin{equation}
\label{msm_eqn}MSM(i,j) = \min\!\Big\{ MSM(i-1,j-1) + |x_{i}-y_{j}|,\; MSM(i-
1,j) + C(x_{i}, x_{i-1}, y_{j}),\; MSM(i,j-1) + C(y_{j}, x_{i}, y_{j-1}) \Big
\},
\end{equation}
with boundary conditions $MSM(0,0)=0$,
$MSM(i,0)=MSM(i-1,0)+C(x_{i},x_{i-1},y_{1})$, and $MSM(0,j)=MSM(0,j-1)+C(y_{j},
x_{1},y_{j-1})$. Here $x_{i}$ and $y_{j}$ denote samples from the two
sequences and $C(\cdot)$ is the MSM split/merge cost that penalises locally
repeating a value to match an extra sample.

Following the standard MSM definition, the split/merge cost uses a non-negative
penalty parameter $c$:
\begin{equation}
\label{msm_cost_eqn}C(a,b,c) =
\begin{cases}
  c,                                 & \text{if }(b\le a\le c)\ \text{or}\ (b\ge a\ge c), \\
  c + \min\bigl(|a-b|,\,|a-c|\bigr), & \text{otherwise.}
\end{cases}
\end{equation}
Intuitively, when $a$ lies between the neighbouring values $b$ and $c$, a
split/merge is treated as a relatively small structural edit (cost $c$); otherwise
it incurs an additional amplitude-dependent penalty.

To improve interpretability across variables and offsets, we additionally
report \texttt{normalised MSM}:
\begin{equation}
\label{msm_norm_eqn}MSM_{\mathrm{norm}}= \frac{MSM}{\frac{1}{N}\sum_{i=1}^{N}|x_{i}|
+ \epsilon},
\end{equation}
where $N$ is the segment length and $\epsilon$ is a small constant to avoid
division by zero.

\subsubsection*{Time-continuous SPM1D inference and Cohen's $d$ effect sizes}
Scalar agreement metrics (RMSE/MAE, DTW/MSM) provide compact summaries but can
be \emph{oversensitive} to small, systematic offsets or short-lived misalignments,
and they do not localise \emph{where} in the gait cycle two waveforms differ.
To complement these metrics, we performed \texttt{SPM1D} hypothesis testing
across the time-normalised gait cycle, treating each gait-cycle waveform as a one-dimensional
random field. This enables detection of temporally contiguous regions of
statistically significant differences while controlling the family-wise error
rate across the full trajectory.

For each joint/side and model, we compared markerless and marker-based waveforms
using a paired design. Let $x_{k}(t)$ denote the marker-based waveform and
$y_{k}(t)$ the markerless waveform for matched cycle $k\in\{1,\dots,K\}$ at
time-normalised gait phase $t\in[0,1]$. Define the difference field $\delta_{k}
(t)=y_{k}(t)-x_{k}(t)$. The SPM test statistic was computed pointwise as a one-sample
(paired) $t$-field:
\begin{equation}
\label{spm_t_eqn}SPM\{t\}(t) = \frac{\bar{\delta}(t)}{s_{\delta}(t)/\sqrt{K}}
,
\end{equation}
where $\bar{\delta}(t)$ and $s_{\delta}(t)$ are the mean and standard deviation
of $\{\delta_{k}(t)\}$ across cycles.

Significant differences were identified as supra-threshold clusters where $|SPM
\{t\}(t)|$ exceeded the random-field-theory critical threshold $t^{*}$ corresponding
to a two-sided family-wise error rate $\alpha$:
\begin{equation}
\label{spm_thresh_eqn}\mathcal{C}= \left\{ t\in[0,1] : \left|SPM\{t\}(t)\right
| > t^{*}(\alpha) \right\}.
\end{equation}
We report the resulting significant time regions (clusters) alongside ensemble-average
curves to indicate where markerless and marker-based kinematics differ in a
statistically reliable manner.

Because statistical significance depends on sample size, we additionally report
time-continuous \texttt{Cohen's $d$} to quantify the \emph{magnitude} of differences
independent of $p$-values. For the paired design, Cohen's $d$ was computed from
the difference field as:
\begin{equation}
\label{cohend_eqn}d(t) = \frac{\bar{\delta}(t)}{s_{\delta}(t)}.
\end{equation}
This paired-effect formulation expresses the mean markerless--marker-based difference
in units of the across-cycle variability. In all SPM figures, the shaded effect-size
bands summarise \emph{absolute} Cohen's $d$ using conventional thresholds: \textbf{blue}
denotes small effects ($0.2\le |d|<0.5$), \textbf{orange} denotes medium
effects ($0.5\le |d|<0.8$), and \textbf{red} denotes large effects ($|d|\ge 0.8$);
values with $|d|<0.2$ are treated as negligible. Presenting SPM1D significance
together with Cohen's $d$ therefore helps distinguish (i) statistically
meaningful dissimilarities with non-trivial magnitude from (ii) small but
detectable differences that may arise from baseline offsets, coordinate conventions,
or minor temporal inconsistencies.

\subsection*{Kinematics Analysis}
For downstream biomarker extraction and evaluation, joint-angle trajectories were trimmed (first/last 10\% of each recording), segmented into gait cycles using extrema-based event definitions, and time-normalized per cycle. To mitigate residual phase offsets introduced by sampling-rate mismatch and upstream processing, cycles were temporally aligned using dynamic time warping (DTW; \eqhref{dtw_eqn}), with fixed Sakoe--Chiba bandwidth and heuristic threshold factors per setting (\stabref{tab:dtw_settings}). Agreement was quantified on both raw (non-mean-centered) and mean-centered waveforms using complementary association, error, and distance measures: Pearson correlation (\eqhref{pearson_eqn}), RMSE and MAE (\eqhref{rmse_eqn}--\eqhref{mae_eqn}), MSM, and DTW-based distances (\eqhref{msm_eqn}--\eqhref{warpeff_eqn}). Finally, to test whether deviations were time-localized and statistically meaningful across the gait cycle (rather than metric sensitivity to small timing offsets), we applied Statistical Parametric Mapping using paired \texttt{SPM\{t\}} (\eqhref{spm_t_eqn}) alongside Cohen's $d$ effect-size trajectories (\eqhref{cohend_eqn}).

\subsubsection*{\texttt{Rajagopal} Model (Raw \& Mean Centered Values)}
\begin{figure}[H]
  \centering

  \begin{subfigure}[t]{0.98\linewidth}
    \centering
    {\scriptsize
    \setlength{\tabcolsep}{3pt}
    \renewcommand{\arraystretch}{1.15}
    \resizebox{\linewidth}{!}{%
    \begin{tabular}{llcccccccccccc}
      \toprule
      \multicolumn{14}{c}{\texttt{\textbf{Rajagopal: Marker-based vs Markerless SCI Flexion Angles (Mean-Centered)}}} \\
      \midrule
      \textbf{Model} & \textbf{Joint} & \textbf{Side} & \textbf{Cycles} & \textbf{Static offset ($^\circ$)} & \textbf{$r$} & \textbf{$p$} & \textbf{RMSE ($^\circ$)} & \textbf{MAE ($^\circ$)} & \textbf{MSM} & \textbf{Norm. MSM} & \textbf{DTW} & \textbf{DTW $z$} & \textbf{Warp eff.} \\
      \midrule
      \multirow{9}{*}{\texttt{Rajagopal}}
      & Hip flexion  & Right & 196 & $\phantom{-}1.24^\circ\,\pm\,5.57^\circ$ & 0.8852 & 0.0084 & 4.85 & 3.84 & 109.9354 & 1.0885 & 30.9176 & 2.1139 & 0.0431 \\
      & Hip flexion  & Left  & 226 & $-0.87^\circ\,\pm\,5.91^\circ$            & 0.8521 & 0.0101 & 5.17 & 4.29 & 117.8510 & 1.1668 & 30.0130 & 2.2730 & 0.0534 \\
      & Knee angle   & Right & 276 & $\phantom{-}1.01^\circ\,\pm\,6.18^\circ$ & 0.8347 & 0.0111 & 6.97 & 5.68 & 137.8484 & 1.3648 & 45.5915 & 2.5891 & 0.0488 \\
      & Knee angle   & Left  & 284 & $\phantom{-}0.07^\circ\,\pm\,6.42^\circ$ & 0.8592 & 0.0008 & 5.97 & 4.85 & 129.2577 & 1.2798 & 34.8236 & 2.3298 & 0.0451 \\
      & Ankle angle  & Right & 296 & $-11.47^\circ\,\pm\,4.71^\circ$           & 0.7092 & 0.0328 & 4.28 & 3.44 & 126.8099 & 1.2555 & 29.8045 & 3.8744 & 0.0599 \\
      & Ankle angle  & Left  & 304 & $-10.65^\circ\,\pm\,4.59^\circ$           & 0.7582 & 0.0033 & 3.70 & 2.97 & 114.5041 & 1.1337 & 23.9580 & 3.3431 & 0.0547 \\
      \cmidrule(lr){2-14}
      & \multirow{3}{*}{\textit{Overall}} & Right & 768  & $-3.07^\circ\,\pm\,5.49^\circ$ & 0.8097 & 0.0174 & 5.37 & 4.32 & 124.8646 & 1.2363 & 35.4379 & 2.8591 & 0.0506 \\
      &                                & Left  & 814  & $-3.82^\circ\,\pm\,5.64^\circ$ & 0.8232 & 0.0047 & 4.95 & 4.04 & 120.5376 & 1.1934 & 29.5982 & 2.6486 & 0.0511 \\
      &                                & All   & 1582 & $-3.45^\circ\,\pm\,5.56^\circ$ & 0.8164 & 0.0111 & 5.16 & 4.18 & 122.7011 & 1.2149 & 32.5180 & 2.7539 & 0.0508 \\
      \bottomrule
    \end{tabular}%
    }
    }
\caption{\small{\textbf{\texttt{Rajagopal} model (mean-centered): segment-level metrics.} Segment-level agreement metrics (mean-centered; DTW-aligned) for \texttt{Rajagopal} \texttt{SMPL} IK vs marker-based MoCap IK; complemented by SPM and Cohen's $d$ over the gait cycle.}}
    \label{tab:ik_metrics_raja_centered}
  \end{subfigure}

  \vspace{1.25em}

  \begin{subfigure}[t]{0.98\linewidth}
    \centering
    \includegraphics[width=\linewidth]{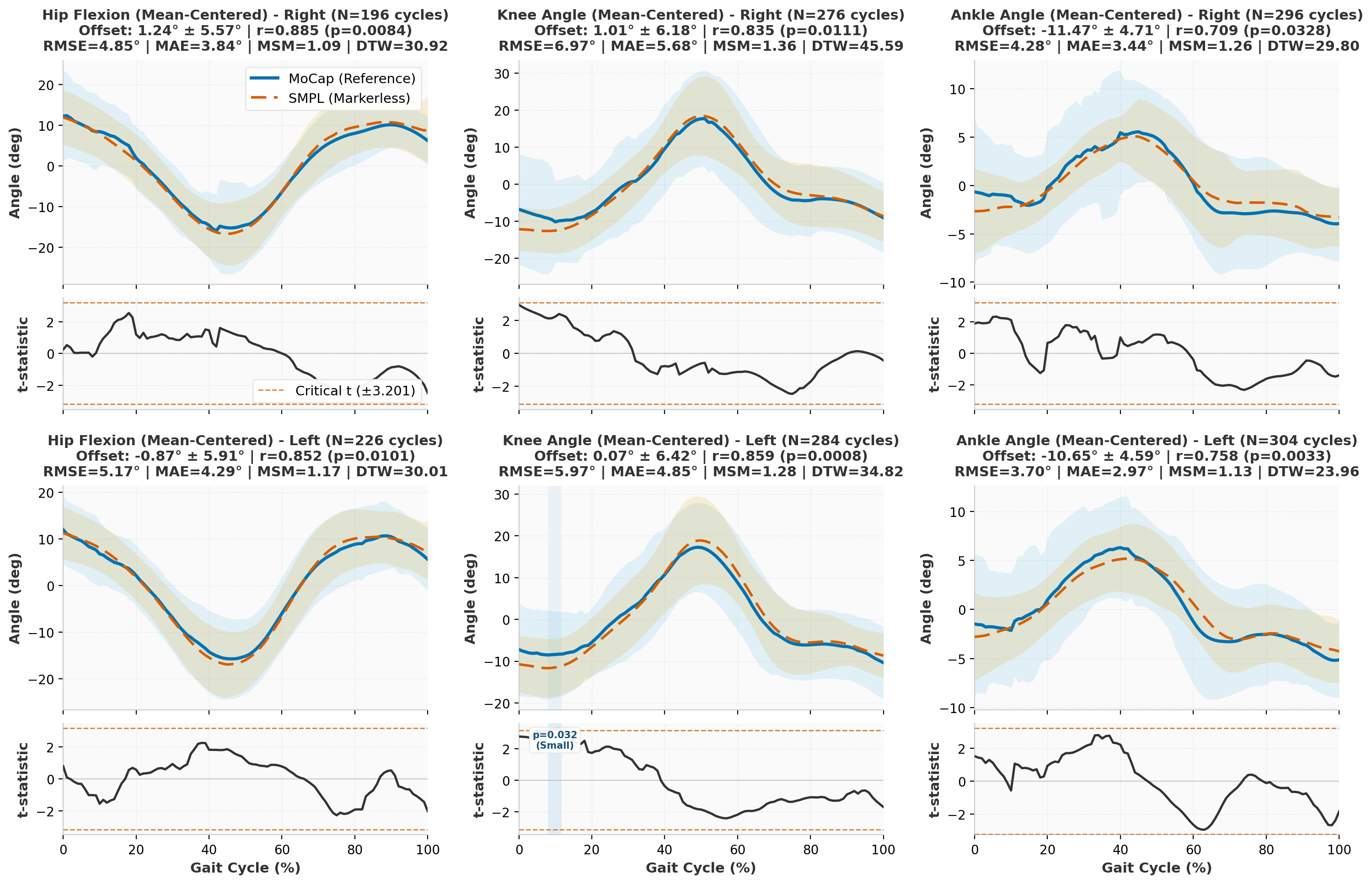}
    \caption{\small{\textbf{\texttt{Rajagopal} model (mean-centered): ensemble-average waveforms.} DTW-aligned ensemble-average flexion waveforms (markerless vs marker-based) with time-continuous SPM testing and Cohen's $d$ effect sizes.}}
    \label{fig:ik_raja_panel_centered}
  \end{subfigure}

  \caption{\small{\textbf{Markerless inverse-kinematics (IK) agreement with marker-based motion capture (\texttt{Rajagopal} model; mean-centered signals).} Markerless kinematics were obtained by reconstructing \texttt{SMPL} meshes from monocular videos, extracting \texttt{SMPL}-based virtual-marker trajectories, and solving IK in \texttt{OpenSim 4.5}; marker-based kinematics were obtained from physical MoCap markers using the same musculoskeletal model and joint definitions. Recordings were trimmed (first/last 10\%), segmented into gait cycles using extrema-based event definitions (hip: local minima; knee/ankle: local maxima), time-normalized, and temporally aligned with dynamic time warping (DTW). Panel\textbf{(a)} reports segment-level similarity/error metrics by joint and side on the \emph{mean-centered} angles. Panel~\textbf{(b)} shows DTW-aligned ensemble-average sagittal-plane flexion waveforms for bilateral hip, knee, and ankle, with Statistical Parametric Mapping (SPM) and Cohen's $d$ effect sizes used to assess time-continuous significance and magnitude of differences.}}
  \label{figure:ik_raja_centered}
\end{figure}

The \texttt{Rajagopal} model was evaluated as a sensitivity check using the same preprocessing, scaling/marker settings, IK solver, and DTW-based evaluation protocol (\figref{fig:osim_pipeline_overview}; \stabref{tab:scaling_marker_settings}; \stabref{tab:dtw_settings}). Segment-level metrics are reported for both mean-centered signals (\sfigref{figure:ik_raja_centered}, panel~\textbf{a}; \stabref{tab:ik_metrics_raja_centered}). On mean-centered angles (1,582 total segments), overall agreement was $r=0.8164$ ($p=0.0111$) with RMSE $=5.16^{\circ}$ and MAE $=4.18^{\circ}$ (overall DTW $=32.5180$, DTW $z=2.7539$, warping efficiency $=0.0508$; \stabref{tab:ik_metrics_raja_centered}). Because mean-centering removes constant offsets, the remaining disagreement is dominated by differences in \emph{waveform shape} (relative excursions) and any residual phase-dependent mismatch after DTW alignment. Proximally, hip correlations were high (Right: $r=0.8852$, Left: $r=0.8521$) with low errors (RMSE $=4.85^{\circ}$--$5.17^{\circ}$; MAE $=3.84^{\circ}$--$4.29^{\circ}$). Knee agreement was slightly weaker but still strong (Right: $r=0.8347$, Left: $r=0.8592$) with RMSE $=5.97^{\circ}$--$6.97^{\circ}$ (MAE $=4.85^{\circ}$--$5.68^{\circ}$). Notably, the ankle showed the lowest correlations (Right: $r=0.7092$, Left: $r=0.7582$), even though its mean-centered absolute errors were comparatively small (RMSE $=3.70^{\circ}$--$4.28^{\circ}$; MAE $=2.97^{\circ}$--$3.44^{\circ}$), indicating that ankle discrepancies were more related to localized waveform-shape/phase differences than large-amplitude deviations. Although offsets are expected to be minimized under mean-centering, the reported static-offset summary terms remained negative at the ankle ($-11.47^{\circ}$ Right; $-10.65^{\circ}$ Left), consistent with a distal baseline bias that would re-emerge when comparing raw angles (\stabref{tab:ik_metrics_raja_centered}).

Across both representations, the Rajagopal sensitivity analysis indicates that the dominant source of disagreement is \emph{not} gross timing mismatch (warping efficiency is essentially unchanged between mean-centered and raw: $0.0508$ in both cases), but rather a combination of (i) a systematic distal baseline bias that chiefly affects absolute angle comparisons, and (ii) residual ankle-specific waveform differences that persist even after mean-centering. The clearest evidence for the baseline-bias component is the large reduction in aggregate error when switching from raw to mean-centered angles (overall RMSE from $9.27^{\circ}$ to $5.16^{\circ}$; MAE from $8.19^{\circ}$ to $4.18^{\circ}$), while the overall correlation remains identical ($r=0.8164$ in both), implying that the \emph{shape association} between markerless and marker-based kinematics is broadly stable and that much of the raw-angle penalty is attributable to constant offsets rather than fundamentally different gait cycle patterns. Joint-wise, the hip and knee behaved as expected for a well-tracked proximal chain: they preserved high correlations (hip $r=0.8521$--$0.8852$, knee $r=0.8347$--$0.8592$) and showed moderate raw-angle errors, suggesting that once phase-aligned, the markerless solution reproduces the primary flexion--extension excursions of these joints with relatively small baseline disagreement. 

However, the ankle emerges as the bottleneck for both \emph{absolute fidelity} and \emph{shape robustness}: its correlation is consistently the lowest (down to $r=0.7092$), and its raw-angle RMSE inflates markedly (to $12.05^{\circ}$--$13.04^{\circ}$) in a way that is consistent with its strongly negative static offsets ($-10.65^{\circ}$ to $-11.47^{\circ}$). Taken together, these patterns suggest that the markerless-to-marker-based gap is driven primarily by distal segment orientation/joint-definition sensitivity (and any accumulated upstream uncertainty); therefore, improvements to ankle/foot modeling, coordinate-system harmonization, or post-hoc offset correction would be expected to yield the largest gains in absolute-angle agreement without necessarily changing the overall waveform correlation.

\subsection*{Predicted ground reaction forces (pGRF) with \texttt{GaitDynamics}}

pGRF was estimated from kinematics and a musculoskeletal model and is therefore sensitive to the upstream pose/mesh reconstruction, inverse kinematics (IK) solution, and the model segment/joint definitions. To support a like-for-like comparison across trials and participants, forces were normalized by body weight (reported as $\mathrm{N/kg}$) and evaluated on step-level segments trimmed to the force plate recording window. Because even small errors in event timing or stance phase duration can dominate waveform comparisons, each step segment was time-normalized and then temporally aligned using dynamic time warping (DTW) prior to computing agreement metrics. We report complementary metrics capturing (i) global association (Pearson $r$ with corresponding $p$-values), (ii) average magnitude error (RMSE and MAE), and (iii) waveform/shape similarity (MSM and normalized MSM), together with DTW distance measures (DTW and DTW $z$) and the resulting warping efficiency (fractional time-warping required).

\begin{figure}[H]
  \centering

  \begin{subfigure}[t]{0.98\linewidth}
    \centering
    \resizebox{\linewidth}{!}{%
      \begin{tabular}{l l r c c r r r r r r r r}
        \toprule
        \textbf{Model} & \textbf{Side} & \textbf{Cycles} & \textbf{Static offset (N/kg)} & \textbf{$r$} & \textbf{$p$} & \textbf{RMSE (N/kg)} & \textbf{MAE (N/kg)} & \textbf{MSM} & \textbf{Norm. MSM} & \textbf{DTW} & \textbf{DTW $z$} & \textbf{Warp eff.} \\
        \midrule
        \multirow{2}{*}{\texttt{Rajagopal}}
        & Left  & 116 & $0.04 \pm 0.95$  & 0.8531 & 0.001 & 2.16 & 1.51 & 89.6821 & 0.8879 & 12.9061 & 2.8237 & 0.0866 \\
        & Right & 116 & $-0.10 \pm 0.88$ & 0.8591 & 0.001 & 2.10 & 1.45 & 86.3859 & 0.8553 & 11.8962 & 2.6549 & 0.0909 \\
        \bottomrule
      \end{tabular}%
    }
    \caption{\small{\textbf{Predicted GRF (Rajagopal model): step-level agreement metrics.} Agreement between predicted GRF magnitude (pGRF; \texttt{GaitDynamics}) and force-plate GRF magnitude, computed on trimmed, bodyweight-normalized, mean-centered step segments after DTW alignment of z-scored signals.}}
    \label{tab:pgrf_results_raja_centred}
  \end{subfigure}

  \vspace{1.0em}

  \begin{subfigure}[t]{0.98\linewidth}
    \centering
    \includegraphics[width=0.95\linewidth]{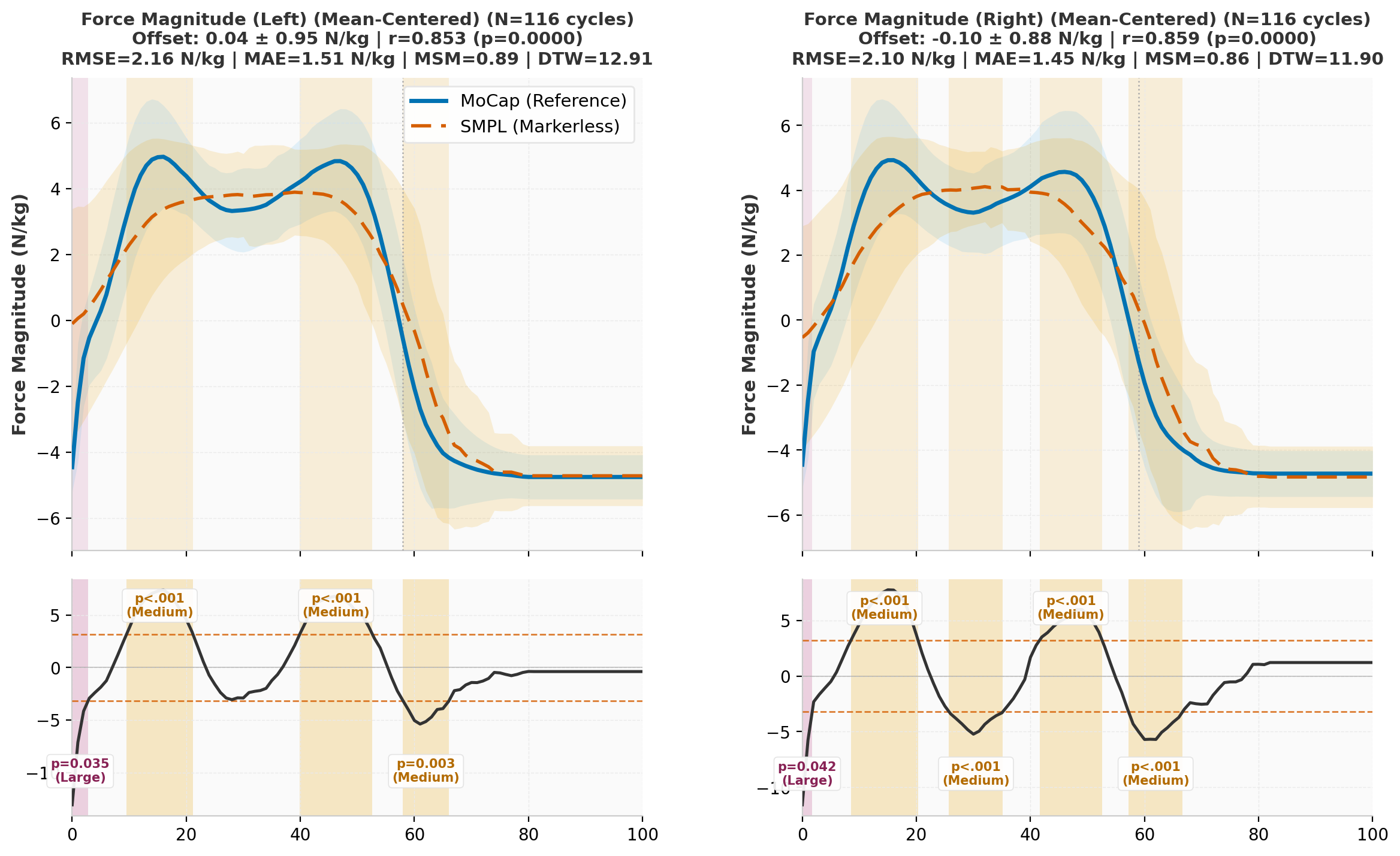}
    \caption{\small{\textbf{Averaged Predicted GRF (Rajagopal model):} The figure shows the averaged \emph{force magnitude} of the predicted GRF (pGRF; \texttt{GaitDynamics}) against the recorded force-plate GRF for the \texttt{Rajagopal} model (mean-centered). Forces are normalized by bodyweight and trimmed to the recorded GRF duration. Results are reported after \texttt{DTW} alignment of the \texttt{z-scored} pGRF segments to the corresponding \texttt{z-scored} GRF segments. The lower panel reports time-continuous SPM inference with Cohen's $d$ effect-size bands (blue/orange/red denote small/medium/large effects as defined in Equation~\ref{cohend_eqn}). Significance levels: $^{*}p<0.05$, $^{**}p<0.01$, $^{***}p<0.001$.}}
    \label{fig:pgrf_raja_centred}
  \end{subfigure}

\end{figure}

\subsubsection*{Pose2Sim model}
For the \texttt{Pose2Sim} pipeline, pGRF was generated using markerless lower-limb kinematics obtained from the \texttt{Pose2Sim} model paired with the \texttt{SMPL}-derived virtual-marker inverse-kinematics (IK) workflow. Specifically, the reconstructed \texttt{SMPL} mesh sequence was converted into time-varying 3D virtual-marker trajectories, which were then used to participant-specifically scale the \texttt{Pose2Sim} \texttt{OpenSim} model using the same segment definitions and marker-selection rules as the marker-based reference. IK was subsequently solved on the scaled model to recover dynamically interpretable sagittal-plane coordinate trajectories (bilateral hip, knee, and ankle flexion/extension) expressed in the \texttt{Pose2Sim}/\texttt{OpenSim} coordinate conventions. These joint-angle time series, together with the subject anthropometrics and model mass properties implied by scaling, were then provided to \texttt{GaitDynamics} to estimate the step-wise predicted GRF magnitude (pGRF), enabling a force-plate-free external-loading input for downstream inverse dynamics while preserving model-specific kinematic and inertial consistency.

The analysis included $116$ step segments for each limb. After DTW alignment, predicted and measured GRF magnitudes exhibited strong waveform association on both sides (left: $r=0.8538$, $p<0.001$; right: $r=0.8662$, $p<0.001$), indicating that the step-to-step loading pattern was largely preserved by the kinematics-driven prediction pipeline.

Bias between pGRF and force-plate GRF magnitude was small: the static offset was $-0.08\pm0.99\,\mathrm{N/kg}$ on the left and $-0.24\pm0.93\,\mathrm{N/kg}$ on the right. Consistent with these small mean offsets, the average magnitude errors were moderate and highly symmetric (left: RMSE $2.37\,\mathrm{N/kg}$, MAE $1.47\,\mathrm{N/kg}$; right: RMSE $2.32\,\mathrm{N/kg}$, MAE $1.43\,\mathrm{N/kg}$). Shape similarity showed similar performance across sides (left: MSM $72.2600$, normalized MSM $0.7154$; right: MSM $71.8375$, normalized MSM $0.7113$). Finally, DTW distances (left: DTW $12.6696$, DTW $z=2.7521$; right: DTW $11.4121$, DTW $z=2.5818$) coupled with low warping efficiencies (left: $0.0926$; right: $0.0888$) indicate that only limited temporal re-parameterization is required; thus, residual disagreement is more consistent with differences in amplitude and local waveform shape than with gross stance-phase timing errors.

\subsubsection*{Rajagopal model}
For the \texttt{Rajagopal} model, the same pGRF workflow was applied using the Rajagopal musculoskeletal model and its segment geometry/joint coordinate conventions. For completeness, we report both mean-centered segments (\stabref{tab:pgrf_results_raja_centred}, \sfigref{fig:pgrf_raja_centred}) and raw (non-mean-centered) segments; each condition again contains $116$ step segments per side.

At the level of global association, the performance closely matched that of the \texttt{Pose2Sim} case: correlations were $r=0.8531$ (Left, $p<0.001$) and $r=0.8591$ (Right, $p<0.001$). The mean bias was also minimal, with static offsets of $0.04\pm0.95\,\mathrm{N/kg}$ (Left) and $-0.10\pm0.88\,\mathrm{N/kg}$ (Right).

For the \emph{mean-centered} comparison, magnitude errors were low (left: RMSE $2.16\,\mathrm{N/kg}$, MAE $1.51\,\mathrm{N/kg}$; right: RMSE $2.10\,\mathrm{N/kg}$, MAE $1.45\,\mathrm{N/kg}$) and MSM-based shape agreement was high (left: MSM $89.6821$, normalized MSM $0.8879$; right: MSM $86.3859$, normalized MSM $0.8553$). DTW metrics remained comparable to those observed with \texttt{Pose2Sim} (left: DTW $12.9061$, DTW $z=2.8237$, warp efficiency $0.0866$; right: DTW $11.8962$, DTW $z=2.6549$, warp efficiency $0.0909$).

For the \emph{raw-waveform} comparison, the errors were slightly higher (Left: RMSE $2.33\,\mathrm{N/kg}$, MAE $1.43\,\mathrm{N/kg}$; Right: RMSE $2.27\,\mathrm{N/kg}$, MAE $1.39\,\mathrm{N/kg}$), and the MSM agreement decreased substantially (Left: MSM $71.4739$, normalized MSM $0.7077$; Right: MSM $69.5528$, normalized MSM $0.6886$). Importantly, the DTW distance and warping efficiency values remained similar (Left: DTW $12.4774$, DTW $z=2.8237$, warp efficiency $0.0866$; Right: DTW $11.0397$, DTW $z=2.6549$, warp efficiency $0.0909$), supporting the interpretation that mean-centering primarily affects baseline/scale-dependent shape agreement, rather than reducing the amount of temporal warping required to align stance-phase dynamics.

Across models, the pGRF magnitude tracked the force-plate GRF magnitude with consistently strong correlations ( $r\approx0.85$--$0.87$ on both sides) and small mean biases (static offsets within $\pm0.24\,\mathrm{N/kg}$), indicating that the overall stance-loading pattern was captured reasonably well by the kinematics-driven pGRF pipeline. The comparatively low and stable warping efficiencies ($\approx0.087$--$0.093$) further suggest that large timing discrepancies were not the dominant failure mode; instead, most residual disagreement was explained by amplitude/shape differences.

Two further deductions follow from the subtable/figure. First, within the \texttt{Rajagopal} model, the strong improvement in MSM/normalized MSM with mean-centering (normalized MSM from $0.7077$ to $0.8879$ on the left (see \stabref{tab:pgrf_results_raja_centred}) together with only modest changes in DTW and warping efficiency implies that baseline offsets and global scaling contribute materially to raw-waveform shape mismatch, whereas the phase structure of the stance waveform is comparatively consistent. Second, because the present evaluation was performed on the \emph{resultant GRF magnitude} rather than the full 3D vector (and does not assess the center-of-pressure location), these results should be interpreted as a \textbf{best-case external-load check}: good magnitude agreement supports plausible \emph{gross loading} but does not guarantee accuracy of the anterior--posterior/medial--lateral components or moment arms that drive joint-moment estimation.

Overall, the pGRF subtable patterns support the use of DTW-aligned, normalized pGRF magnitude as a viable surrogate for force-plate GRF magnitude at the step level in this dataset because the prediction preserves the dominant stance-phase loading envelope and yields a strong step-wise association after timing normalization. However, the ensemble comparisons also revealed structured, biomechanics-relevant deviations that limit the direct interchangeability of pGRF with force-plate GRF when the signal is used as an input to inverse dynamics rather than as an endpoint metric. In particular, the systematic waveform smoothing visible in the ensemble plots (\sfigref{fig:pgrf_raja_centred}) is consistent with attenuated high-frequency and peak features within stance, which can shift both the effective impulse distribution and the instantaneous moment arm about the joints if accompanied by errors in the predicted line of action and CoP trajectory. Because inverse dynamics couples external loading with segment accelerations, even modest within-stance shape differences in GRF magnitude can produce disproportionately large errors in the recovered net joint moments once propagated through the multi-segment chain. This sensitivity is greatest proximally, hip moments are typically smaller in magnitude than ankle moments and are computed after accumulating distal force and inertial contributions, making them more vulnerable to combined errors in GRF magnitude, timing, and application point. Consequently, a strong pGRF--GRF agreement at the force-magnitude level does not guarantee an equivalently strong agreement in fully markerless joint moments, especially at the hip, where errors in both GRF direction and line of action can dominate the net-moment balance.

\subsection*{Inverse-dynamics configurations and reliability considerations}

\begin{figure}[H]
  \centering
  \includegraphics[width=0.8\linewidth]{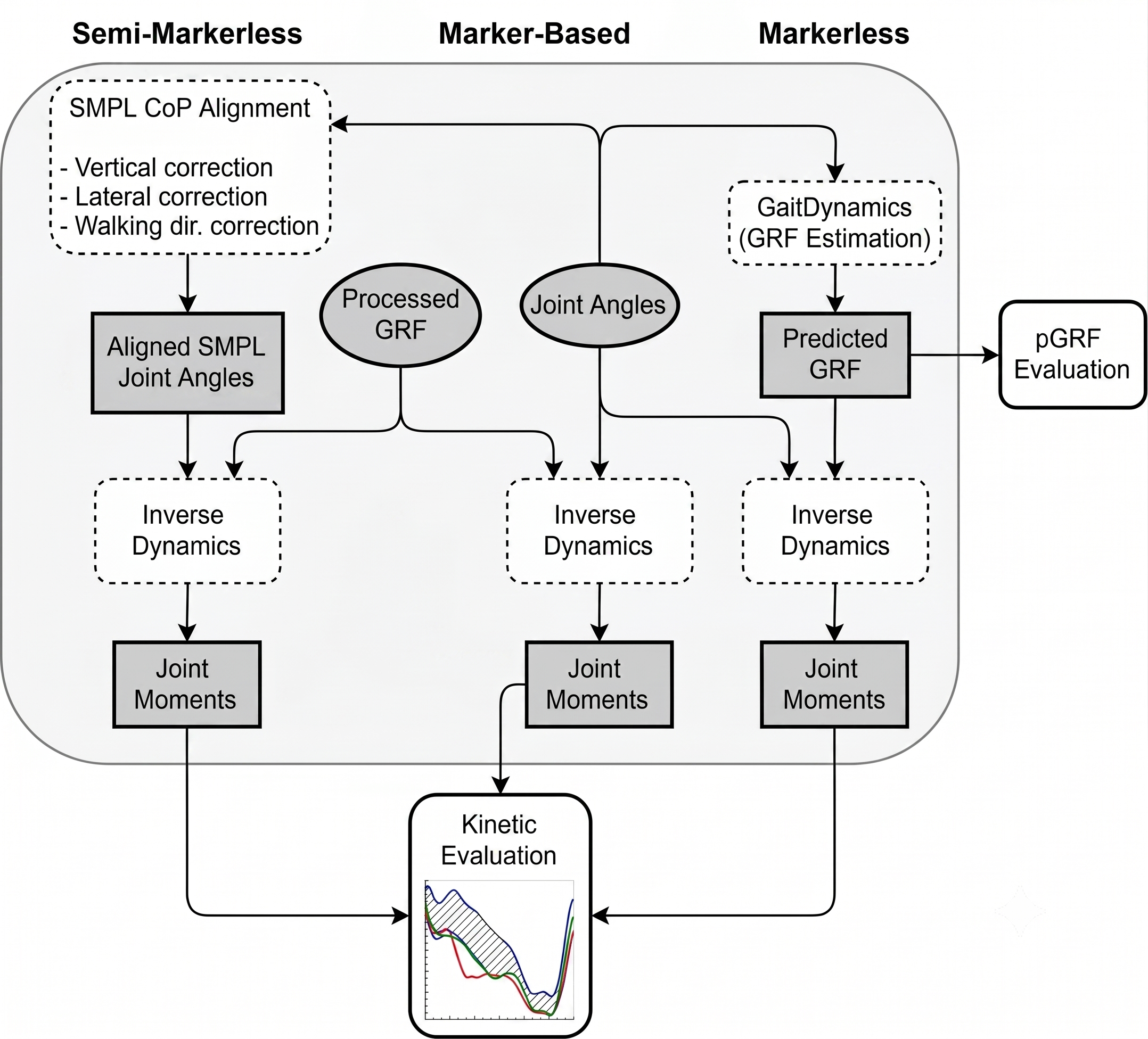}
  \caption{\small{\textbf{Kinetic Pipeline Overview.} 
  Schematic workflow of the kinetic pipeline. Three variations were made: \textit{marker-based} (MB-MoCap with GRF), \textit{semi-markerless} (SMPL with GRF), and \textit{markerless} (SMPL with pGRF). \texttt{GaitDynamics} was used to estimate GRF (pGRF) without force plate using the subject-specific scaled model, joint angles, and anthropometric data. The joint angles and pGRF were used to obtain \textit{markerless} joint moments. To obtain \textit{semi-markerless} joint moments, \texttt{SMPL} joint angles had to be aligned to the CoP of GRF prior to ID. All resulting joint moments time series were trimmed to the GRF recorded duration and evaluated.}}
  \label{fig:kinetic_pipeline}
\end{figure}

\subsubsection*{Kinetics: detailed implementation (moved from the main Methods)}
Kinetic quantities were computed with \texttt{OpenSim 4.5} inverse dynamics (ID) using the subject-specific scaled musculoskeletal model and the corresponding joint kinematics from IK. We analyzed two markerless kinetics configurations that progressively relax laboratory requirements while preserving a common computational workflow and reference definition. \textbf{Semi-markerless kinetics} combined \texttt{SMPL}-driven kinematics (virtual-marker IK) with \textbf{measured} force-plate GRF, enabling us to isolate how markerless kinematic uncertainty propagates into joint moments when the external load is known. \textbf{Fully markerless kinetics} replaced the force plate by using \textbf{predicted GRF (pGRF)} from \texttt{GaitDynamics} as the external load input to ID, yielding an end-to-end markerless pipeline (video $\rightarrow$ kinematics $\rightarrow$ pGRF $\rightarrow$ moments). In all cases, the marker-based MoCap pipeline with measured GRF served as the reference for benchmarking of joint moments.

\begin{figure}[H]
  \centering

  \begin{subfigure}[t]{0.49\linewidth}
    \centering
    \includegraphics[width=\linewidth]{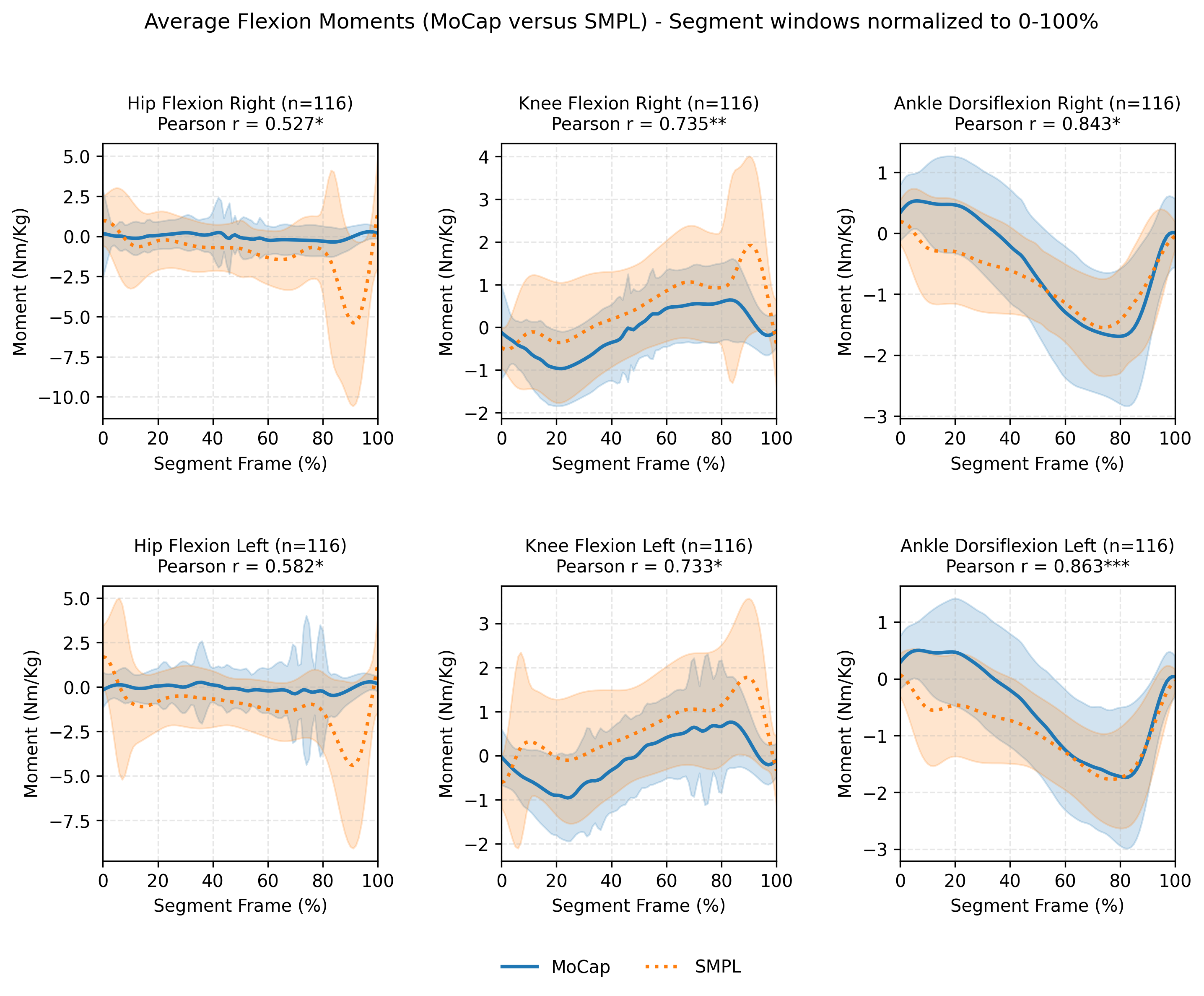}
    \caption{\small{\textbf{Semi-markerless kinetics (measured GRF):} Averaged \texttt{SMPL} inverse-dynamics (ID) joint moments driven by measured GRF, compared with averaged MB-MoCap ID moments (also driven by measured GRF) in \texttt{OpenSim 4.5} using the \texttt{Pose2Sim} musculoskeletal model (\textbf{116} matched trials).}}
    \label{fig:id_p2s}
  \end{subfigure}\hfill
  \begin{subfigure}[t]{0.49\linewidth}
    \centering
    \includegraphics[width=\linewidth]{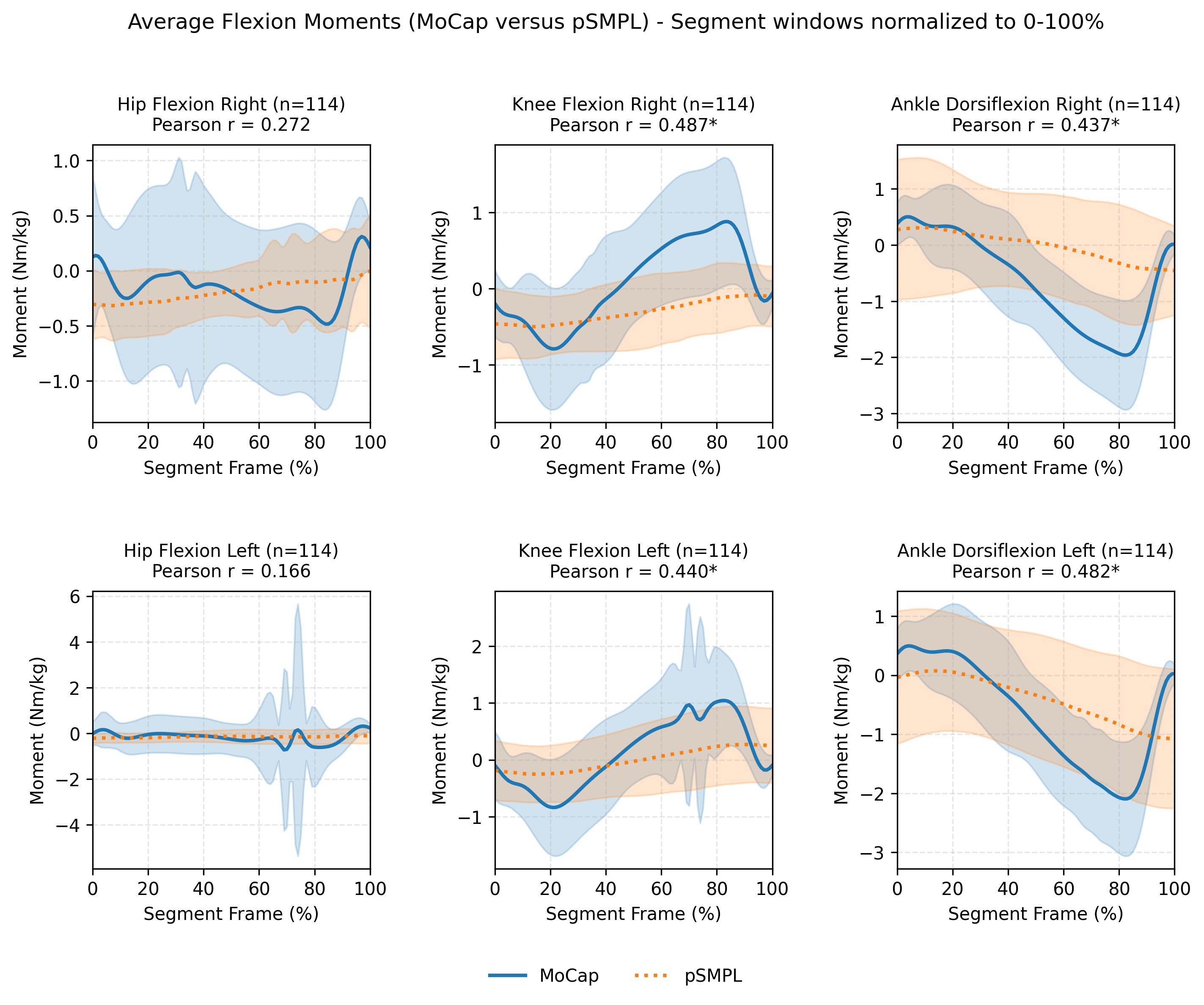}
    \caption{\small{\textbf{Fully markerless kinetics (predicted GRF):} Averaged \texttt{SMPL} ID joint moments driven by predicted GRF (pGRF), compared with MB-MoCap ID moments driven by measured GRF in \texttt{OpenSim 4.5} using the \texttt{Pose2Sim} musculoskeletal model (\textbf{114} matched trials).}}
    \label{fig:pid_p2s}
  \end{subfigure}
  \caption{\small{\textbf{Averaged sagittal-plane lower-limb joint moments from inverse dynamics for the Pose2Sim model under semi-markerless versus fully markerless loading conditions.} Each subfigure shows the ensemble-averaged hip, knee, and ankle flexion moments for the left and right limbs, normalized by bodyweight and time-trimmed to the duration of the recorded force-plate contact. In (a), both \texttt{SMPL}-based and MB-MoCap inverse dynamics were driven by the same measured GRF, isolating the effect of markerless (mesh-based) kinematics on the recovered joint-moment waveforms (\textbf{116} matched trials). In (b), \texttt{SMPL}-based inverse dynamics was driven by predicted GRF (pGRF), while the MB-MoCap reference remained driven by measured GRF, reflecting a fully markerless kinetics pipeline in which external loading must be estimated (\textbf{114} matched trials). For both conditions, the results are reported after dynamic time warping (\texttt{DTW}) alignment of \texttt{z}-scored \texttt{SMPL} segments to the corresponding \texttt{z}-scored MB-MoCap segments to account for inter-trial timing differences before waveform comparisons. The strongest agreement is observed distally (ankle and knee) and degrades proximally (hip), consistent with distal-to-proximal error amplification in inverse dynamics. Significance levels: $^{*}p<0.05$, $^{**}p<0.01$, $^{***}p<0.001$.}}
  \label{fig:avg_moments_p2s}
\end{figure}

We computed joint moments using the \texttt{OpenSim} ID Tool by supplying an ExternalLoads definition that referenced either measured GRF or predicted GRF (pGRF) time series (\texttt{.mot}) and applied the left/right ground-reaction wrenches to the calcaneus segments in the laboratory ground coordinate system. Prior to running ID, joint-angle coordinates were low-pass filtered using a low-pass \textbf{4th-order Butterworth} filter with a \textbf{6~Hz} cut-off (consistent with the harmonized filtering across modalities; see \figref{fig:osim_pipeline_overview}). ID outputs were exported as \texttt{.sto} files and analyzed for sagittal-plane hip, knee, and ankle flexion/extension moments. For paired benchmarking, we restricted GRF/pGRF and moment time series to identical force-plate contact windows (vertical GRF $>20$~N) and retained only trials containing two consecutive, clearly ordered steps (\textit{Left--Right} or \textit{Right--Left}); left/right gait cycles were trimmed to the corresponding force-plate frame indices, normalized by bodyweight (GRF in $\mathrm{N/kg}$; moments in $\mathrm{Nm/kg}$), saved as per-step \texttt{.sto} files (time reset to 0), and pooled into per-source \texttt{.csv} tables for downstream statistics (see \figref{fig:trials_included}).

For waveform benchmarking, gait cycles were z-scored, and where necessary, temporally aligned using DTW prior to computing agreement metrics, following the same evaluation framework as kinematics. pGRF benchmarking was performed on the resultant GRF magnitude computed from the anterior--posterior, vertical, and medio--lateral components. Joint-moment benchmarking was performed on the hip, knee, and ankle sagittal-plane moments within the corresponding left/right step of each paired trial.

\paragraph*{Semi-markerless kinetics: CoP-consistent alignment of markerless kinematics}
A key requirement for physically plausible ID is that the external GRF wrench is applied with a realistic line of action relative to the model. However, monocular HMR reconstructs motion in a camera-centred coordinate system, and without explicit camera calibration the global translation and walking direction in the reconstructed \texttt{SMPL} sequence are ambiguous \cite{shen2024gvhmr,kim2022camera,yang2024ofcam}. In our data, this manifested as stationary or inconsistently directed overground trajectories in which the \texttt{SMPL} marker cloud remained near the origin instead of translating across the instrumented force plates, producing large offsets between the reconstructed foot locations and the force-plate center of pressure (CoP). Because joint moments depend on the GRF moment arm, these offsets yielded unrealistically large moments if ID was run directly.

\begin{figure}[H]
  \centering
  \begin{subfigure}[t]{0.49\linewidth}
    \centering
    \includegraphics[width=\linewidth]{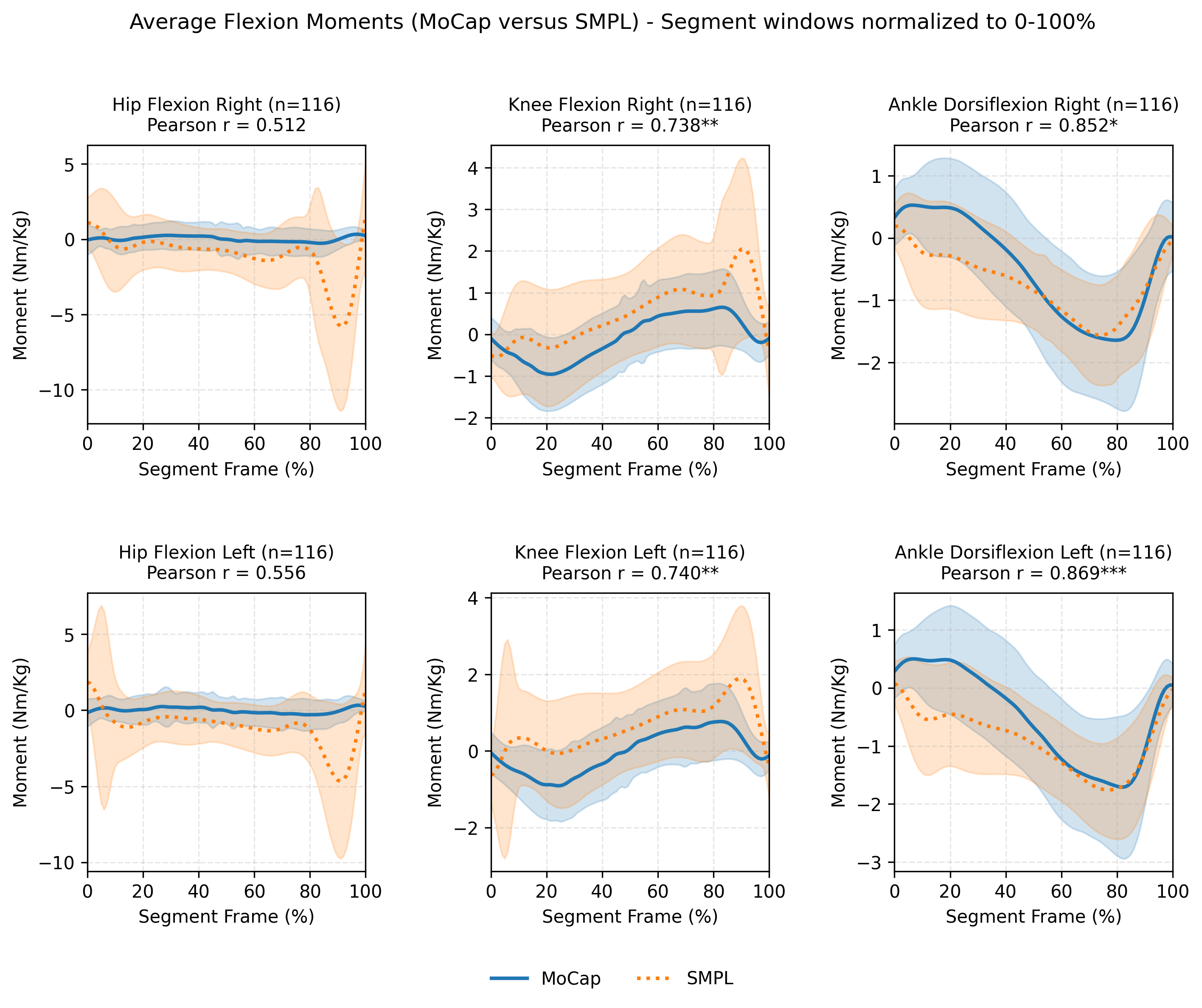}
    \caption{\small{\textbf{Semi-markerless kinetics (measured GRF):} Averaged \texttt{SMPL} inverse-dynamics (ID) joint moments driven by measured GRF, compared with averaged MB-MoCap ID moments (also driven by measured GRF) in \texttt{OpenSim 4.5} using the Rajagopal musculoskeletal model (\textbf{116} matched trials).}}
  \end{subfigure}
  \hfill
  \begin{subfigure}[t]{0.49\linewidth}
    \centering
    \includegraphics[width=\linewidth]{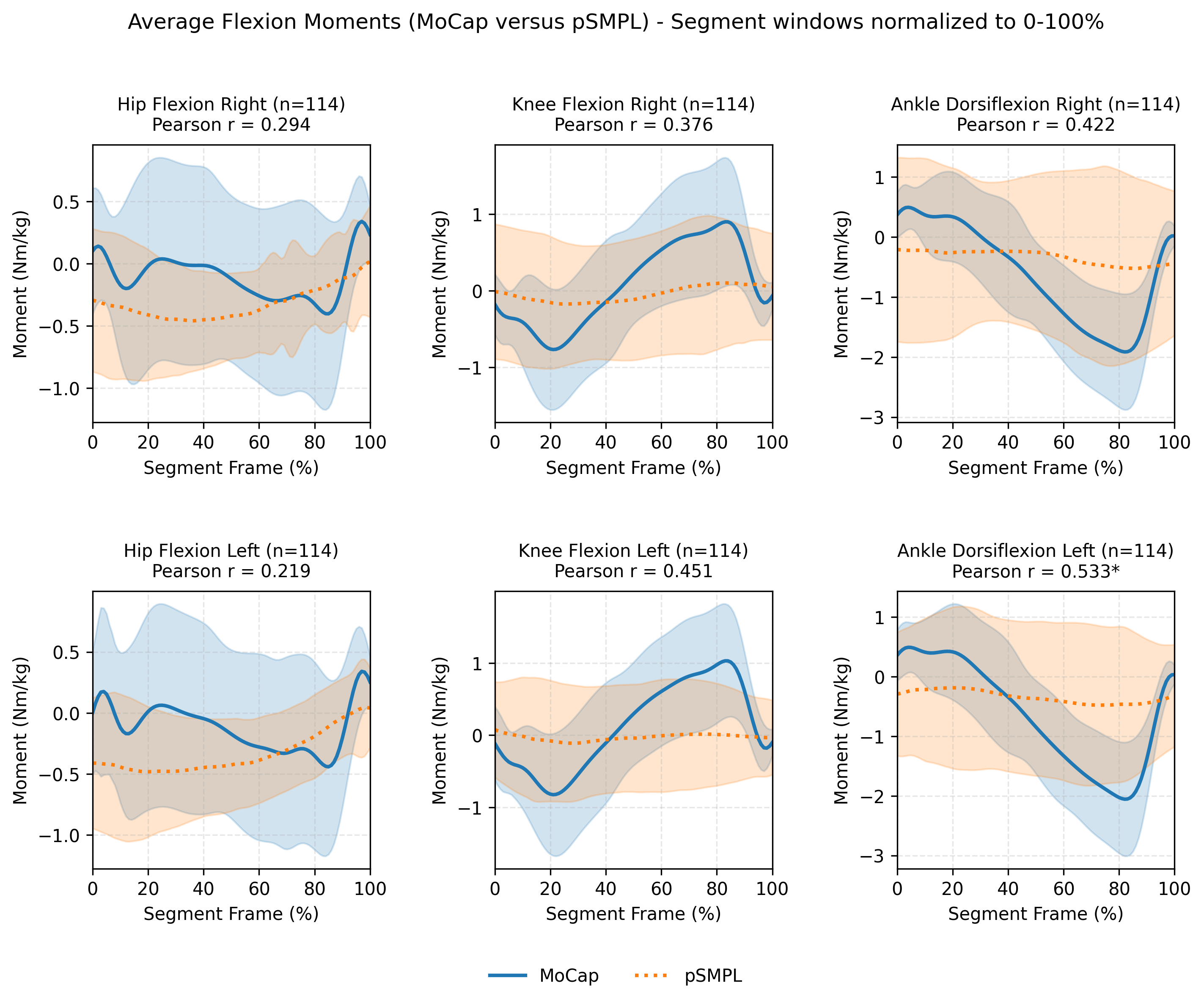}
    \caption{\small{\textbf{Fully markerless kinetics (predicted GRF):} Averaged \texttt{SMPL} ID joint moments driven by predicted GRF (pGRF), compared with MB-MoCap ID moments driven by measured GRF in \texttt{OpenSim 4.5} using the Rajagopal musculoskeletal model (\textbf{114} matched trials).}}
  \end{subfigure}

  \caption{\small{\textbf{Averaged (semi-)markerless sagittal-plane joint moments vs.
  marker-based reference (Rajagopal model).}
  Each panel shows the ensemble-average flexion/extension moments for the six lower-limb sagittal degrees of freedom (right/left hip, knee, and ankle flexion moments) computed with inverse dynamics in OpenSim~4.5.
  In both panels, the markerless moments (\texttt{SMPL}-driven kinematics) are compared against the marker-based reference moments (MB-MoCap-driven kinematics), with the reference always driven by measured force-plate GRF.
  \textbf{(a) Semi-markerless kinetics (GRF-driven ID):} inverse dynamics driven by measured GRF combined with markerless kinematics after GRF centre-of-pressure (CoP) alignment.
  \textbf{(b) Fully markerless kinetics (pGRF-driven ID):} inverse dynamics driven by predicted GRF (pGRF) combined with the same markerless kinematics.
  Trials were segmented into individual steps and trimmed to the recorded GRF time window; moments were normalized by body mass (reported in Nm/kg).
  The analysis included \textbf{116} matched trials for GRF-driven ID and \textbf{114} matched trials for pGRF-driven ID.
  For waveform comparison, step segments were z-scored and temporally aligned using dynamic time warping (DTW); the displayed curves represent the average waveform across aligned segments.
  Significance levels for the corresponding segment-level correlations are indicated as $^{*}p<0.05$, $^{**}p<0.01$, $^{***}p<0.001$.}}%
  \label{figure:id_raja}
  \label{figure:pid_raja}
\end{figure}

To obtain mechanically consistent semi-markerless kinetics, we aligned the \texttt{SMPL}-driven kinematics to the measured force-plate CoP before ID. Foot contact was detected using a vertical GRF threshold of \textbf{20~N}. A \emph{foot point} was defined per frame as the mean of the \texttt{SMPL}-derived heel and toe virtual markers; frames with the foot on the plate were labelled \emph{stance frames}. For trials with inconsistent walking direction across the camera view, we horizontally mirrored the source \texttt{SCAI SCI Gait} videos before mesh reconstruction so the recovered motion matched the GRF/MoCap direction (Figure~\ref{fig:osim_pipeline_overview}).

\textbf{Horizontal correction.} For each frame $t$, we computed the offset between each stance foot point and the corresponding force-plate CoP in anterior--posterior ($X$) and medial--lateral ($Z$). Offsets were combined using vertical-GRF weights $w_{L}(t)$ and $w_{R}(t)$ (normalised so $w_{L}(t)+w_{R}(t)=1$), so the loaded limb dominates in single support with smooth transitions in double support. The resulting translation was applied to $pelvis\_tx$ and $pelvis\_tz$:
\begin{equation}
\label{xz_corr_eqn}
\begin{aligned}
  \Delta X(t) & = w_{L}(t)\,\bigl(\text{COP}_{L,x}(t) - FP_{L,x}(t)\bigr) + w_{R}(t)\,\bigl(\text{COP}_{R,x}(t) - FP_{R,x}(t)\bigr), \\
  \Delta Z(t) & = w_{L}(t)\,\bigl(\text{COP}_{L,z}(t) - FP_{L,z}(t)\bigr) + w_{R}(t)\,\bigl(\text{COP}_{R,z}(t) - FP_{R,z}(t)\bigr),
\end{aligned}
\end{equation}
where $FP_{L,\cdot}(t)$ and $FP_{R,\cdot}(t)$ are the left/right foot-point coordinates. To reduce jitter and avoid impulsive translations, $\Delta X(t)$ and $\Delta Z(t)$ were smoothed with a \textbf{0.05~s} moving-average window before application.

\textbf{Vertical correction.} We applied a per-trial constant vertical offset to $pelvis\_ty$ to set a plausible ground height. Over stance frames, we computed the GRF-weighted distribution of foot-point heights $Y_{stance}$ and used a robust lower-tail statistic to reduce sensitivity to transient foot--ground penetration:
\begin{equation}
\label{y_corr_eqn}\Delta Y = Y_{\text{ground}}- P_{5}\bigl(Y_{stance}\bigr),
\end{equation}
with $Y_{\text{ground}}=0$ and $P_{5}(\cdot)$ the 5th percentile. We applied $\Delta Y$ uniformly to all frames of the \texttt{SMPL}-driven kinematics.

After these corrections, we re-exported an \texttt{OpenSim}-compatible \texttt{.mot} in the laboratory ground frame, applying the corrected pelvis translations ($pelvis\_tx$, $pelvis\_ty$, $pelvis\_tz$) frame-by-frame and preserving all other IK joint coordinates. We then paired this aligned \texttt{.mot} with the measured force-plate GRF/CoP and ran the \texttt{OpenSim} ID Tool using the subject-specific scaled \texttt{.osim} model and an ExternalLoads setup that applies each GRF wrench at the recorded CoP under the appropriate foot segment (ground frame). Net sagittal-plane moments for ankle, knee, and hip were exported as \texttt{.sto} (\sfigref{fig:kinetic_pipeline}, \figref{fig:osim_pipeline_overview}).

\begin{table}[H]
\centering
\begin{tabular}{llllllll}
\toprule
\multicolumn{8}{c}{\texttt{\textbf{Marker-based vs Semi-Markerless SCI Flexion Moments}}} \\
\midrule
\textbf{Input} & \textbf{Model} & \textbf{Joints} & \textbf{Segments} & \textbf{Pearson R} & \textbf{Pearson P} & \textbf{DTW Dist.} & \textbf{RMSE (Nm/kg)} \\
\midrule
\multirow{12}{*}{\textit{GRF}} & \multirow{6}{*}{\texttt{Pose2Sim}}
& Right Ankle  & 116 & 0.843 & 0.0198$^{*}$   & 3.156 & 0.587 \\
& & Left  Ankle  & 116 & 0.863 & 0.0002$^{***}$ & 3.114 & 0.608 \\
& & Right Knee   & 116 & 0.735 & 0.0090$^{**}$  & 4.527 & 1.095 \\
& & Left  Knee   & 116 & 0.733 & 0.0110$^{*}$   & 4.620 & 1.066 \\
& & Right Hip    & 116 & 0.527 & 0.0466$^{*}$   & 6.587 & 2.411 \\
& & Left  Hip    & 116 & 0.582 & 0.0288$^{*}$   & 6.319 & 2.215 \\
\cmidrule(lr){2-8}
& \multirow{6}{*}{\texttt{Rajagopal}}
& Right Ankle  & 116 & 0.852 & 0.0107$^{*}$   & 3.116 & 0.578 \\
& & Left  Ankle  & 116 & 0.869 & 0.0003$^{***}$ & 3.063 & 0.603 \\
& & Right Knee   & 116 & 0.738 & 0.0077$^{**}$  & 4.518 & 1.109 \\
& & Left  Knee   & 116 & 0.740 & 0.0054$^{**}$  & 4.638 & 1.069 \\
& & Right Hip    & 116 & 0.512 & 0.0570         & 6.765 & 2.533 \\
& & Left  Hip    & 116 & 0.556 & 0.0531         & 6.588 & 2.298 \\
\midrule
\multicolumn{8}{c}{\texttt{\textbf{Marker-based vs Markerless SCI Flexion Moments}}} \\
\midrule
\multirow{12}{*}{\textit{pGRF}} & \multirow{6}{*}{\texttt{Pose2Sim}}
& Right Ankle  & 114 & 0.437 & 0.0413$^{*}$  & 7.555 & 1.323 \\
& & Left  Ankle  & 114 & 0.482 & 0.0382$^{*}$  & 7.202 & 1.277 \\
& & Right Knee   & 114 & 0.487 & 0.0224$^{*}$  & 7.132 & 0.872 \\
& & Left  Knee   & 114 & 0.440 & 0.0496$^{*}$  & 7.376 & 0.952 \\
& & Right Hip    & 114 & 0.272 & 0.1114        & 8.059 & 0.694 \\
& & Left  Hip    & 114 & 0.166 & 0.1654        & 9.012 & 0.817 \\
\cmidrule(lr){2-8}
& \multirow{6}{*}{\texttt{Rajagopal}}
& Right Ankle  & 114 & 0.422 & 0.0583        & 6.988 & 1.399 \\
& & Left  Ankle  & 114 & 0.533 & 0.0242$^{*}$  & 6.017 & 1.338 \\
& & Right Knee   & 114 & 0.376 & 0.0545        & 7.303 & 0.978 \\
& & Left  Knee   & 114 & 0.451 & 0.0673        & 6.635 & 0.988 \\
& & Right Hip    & 114 & 0.294 & 0.1292        & 7.993 & 0.716 \\
& & Left  Hip    & 114 & 0.219 & 0.1439        & 8.375 & 0.759 \\
\bottomrule
\end{tabular}
\caption{\small{\textbf{Benchmarking of \texttt{SMPL}-based (Semi-)Markerless Kinetics:} This table compares \texttt{SMPL} ID solutions driven by measured GRF (\texttt{GRF}; semi-markerless) and by predicted GRF (\texttt{pGRF}; markerless) against MB-MoCap ID solutions driven by GRF, all computed in OpenSim 4.5. The analysis included 116 (\texttt{GRF}) and 114 (\texttt{pGRF}) matched trials, normalized by body weight and trimmed to the recorded GRF duration, for six lower limb sagittal joint moments (bilateral hip, knee, and ankle flexion moments). The results are reported after \texttt{DTW} alignment of the \texttt{z-scored} \texttt{SMPL} segments to the corresponding \texttt{z-scored} MB-MoCap segments. Significance levels: $^{*}p<0.05$, $^{**}p<0.01$, $^{***}p<0.001$.}}%
\label{tab:semi_markerless_id}
\end{table}

\paragraph*{Fully markerless kinetics: pGRF-driven inverse dynamics}
For fully markerless kinetics, ID used \textbf{predicted GRF (pGRF)} instead of force-plate data. From the \texttt{SMPL}-driven IK kinematics, we generated pGRF with \texttt{GaitDynamics} using the participant-specific scaled \texttt{.osim} model, the joint-coordinate time series (\texttt{.mot}), and anthropometrics (height, bodyweight). Using the default overground setup, we set walking speed to \textbf{0~m/s} so speed was inferred from kinematics. \texttt{GaitDynamics} operates on fixed windows (\textbf{1.5~s} at \textbf{100~Hz}) and outputs time-varying loads consistent with the model coordinate definitions used during training.

\paragraph*{How similarity was identified (practical interpretation)}
Similarity was defined using multiple complementary criteria: (i) \textbf{phase similarity} via \texttt{DTW}, which compensates for small time shifts between segments; low DTW cost and high warping efficiency indicate close phase matching. (ii) \textbf{shape similarity} via \texttt{Pearson $r$} (after alignment where relevant), which reflects preservation of peaks/troughs and relative timing even with a constant offset. (iii) \textbf{amplitude/bias similarity} via \texttt{RMSE/MAE} and the static offset, separating baseline shifts from true excursion differences. (iv) \textbf{time-localised similarity} via \texttt{SPM1D} and \texttt{Cohen's $d$}, identifying whether disagreements are diffuse or concentrated in specific gait phases (e.g., loading response or swing).

We exported the predicted loads as an \texttt{OpenSim}-compatible pGRF \texttt{.mot} and used it to populate ExternalLoads for the \texttt{OpenSim} ID Tool. Left/right channels were mapped to the corresponding calcaneus segments and expressed in the laboratory ground frame, matching measured-GRF conventions so the ID workflow was unchanged. ID on the same scaled \texttt{.osim} model with \texttt{SMPL}-driven kinematics and pGRF produced sagittal-plane hip, knee, and ankle flexion/extension moments exported as \texttt{.sto}.

This is the strictest end-to-end test because ID combines two coupled error sources: markerless kinematics (segment orientations, pelvis translation, and derivatives) and predicted loading (magnitude/direction, timing, and implied moment arm). We therefore interpret fully markerless kinetics mainly as a test of whether the pipeline reproduces clinically interpretable joint-moment \emph{waveforms} without laboratory instrumentation, and we report it separately from the semi-markerless condition with measured loading (\sfigref{fig:kinetic_pipeline}).

\subsubsection*{Semi-markerless inverse dynamics (\texttt{SMPL} kinematics + measured GRF)}
Semi-markerless ID combines \textbf{markerless kinematics} (joint angles from \texttt{SMPL}-derived virtual-marker IK) with \textbf{measured external loading} (force-plate GRF). This removes physical markers while keeping laboratory loading, isolating the effect of \texttt{SMPL}-driven kinematics on inverse-dynamics moments when the external wrench is known.

Because \texttt{SMPL} is reconstructed in a camera-centric world frame, global translation and foot placement can be offset from the force-plate frame. Before ID, we therefore \textbf{aligned the \texttt{SMPL} kinematics to the measured GRF CoP} so the GRF wrench acts at the correct point under the foot; even small CoP errors change the ankle GRF moment arm and propagate to proximal moments via the Newton--Euler recursion.

With the GRF held fixed, the observed agreement patterns quantify the net impact of kinematic differences (segment orientations, pelvis translation, and resulting coordinate accelerations) on the recovered moments.

\textbf{Pose2Sim model} (semi-markerless; measured GRF; 116 matched trials). The most accurate moments were distal and the least accurate were proximal (\stabref{tab:semi_markerless_id}). At the \textbf{ankle}, correlations were highest and errors were lowest: Right ankle $R=0.843$ ($p=0.0198$), DTW distance 3.156, RMSE $0.587\,\mathrm{Nm/kg}$; Left ankle $R=0.863$ ($p=0.0002$), DTW distance 3.114, RMSE $0.608\,\mathrm{Nm/kg}$. At the \textbf{knee}, agreement remained strong but degraded relative to the ankle: Right knee $R=0.735$ ($p=0.0090$), DTW distance 4.527, RMSE $1.095\,\mathrm{Nm/kg}$; Left knee $R=0.733$ ($p=0.0110$), DTW distance 4.620, RMSE $1.066\,\mathrm{Nm/kg}$. The largest discrepancies occurred at the \textbf{hip}, consistent with distal-to-proximal amplification: Right hip $R=0.527$ ($p=0.0466$), DTW distance 6.587, RMSE $2.411\,\mathrm{Nm/kg}$; Left hip $R=0.582$ ($p=0.0288$), DTW distance 6.319, RMSE $2.215\,\mathrm{Nm/kg}$.

\textbf{Rajagopal model} (semi-markerless; measured GRF; 116 matched trials). The same distal-to-proximal ordering was observed, with slightly higher distal correlations but similarly elevated hip errors (\stabref{tab:semi_markerless_id}). Right ankle $R=0.852$ ($p=0.0107$), DTW distance 3.116, RMSE $0.578\,\mathrm{Nm/kg}$; Left ankle $R=0.869$ ($p=0.0003$), DTW distance 3.063, RMSE $0.603\,\mathrm{Nm/kg}$. Right knee $R=0.738$ ($p=0.0077$), DTW distance 4.518, RMSE $1.109\,\mathrm{Nm/kg}$; Left knee $R=0.740$ ($p=0.0054$), DTW distance 4.638, RMSE $1.069\,\mathrm{Nm/kg}$. Right hip $R=0.512$ ($p=0.0570$), DTW distance 6.765, RMSE $2.533\,\mathrm{Nm/kg}$; Left hip $R=0.556$ ($p=0.0531$), DTW distance 6.588, RMSE $2.298\,\mathrm{Nm/kg}$. Notably, hip correlations were the only semi-markerless cases that did \emph{not} reach conventional statistical significance (both $p>0.05$), reinforcing that proximal kinetics were the least stable output under \texttt{SMPL}-driven kinematics, even when GRF was measured.

\subsubsection*{Fully markerless inverse dynamics (\texttt{SMPL} kinematics + predicted GRF)}
Fully markerless ID removes the force plate by replacing measured GRF with \textbf{predicted GRF (pGRF)}. In this pipeline, \texttt{GaitDynamics} estimates the external ground-reaction wrench using the subject-specific scaled model, \texttt{SMPL}-derived joint angles, and anthropometric quantities; the resulting pGRF is then supplied to the \texttt{OpenSim} ID tool. This configuration therefore combines two coupled sources of error: \textbf{(i) kinematic uncertainty from \texttt{SMPL}} and \textbf{(ii) loading uncertainty from pGRF} (force magnitude and direction, timing, and CoP trajectory), with the latter directly modulating the external moment arms that drive the joint-moment recursion.

Consistent with this compounded uncertainty, agreement dropped across \emph{all} joints, and the distal-to-proximal gradient became more pronounced (\stabref{tab:semi_markerless_id}).

\textbf{Pose2Sim model} (fully markerless; pGRF; 114 matched trials). At the \textbf{ankle}, correlations reduced to Right ankle $R=0.437$ ($p=0.0413$), DTW distance 7.555, RMSE $1.323\,\mathrm{Nm/kg}$ and Left ankle $R=0.482$ ($p=0.0382$), DTW distance 7.202, RMSE $1.277\,\mathrm{Nm/kg}$. At the \textbf{knee}, agreement was comparable but slightly higher on the right: Right knee $R=0.487$ ($p=0.0224$), DTW distance 7.132, RMSE $0.872\,\mathrm{Nm/kg}$; Left knee $R=0.440$ ($p=0.0496$), DTW distance 7.376, RMSE $0.952\,\mathrm{Nm/kg}$. The \textbf{hip} showed the weakest association and the clearest collapse in reliability: Right hip $R=0.272$ ($p=0.1114$), DTW distance 8.059, RMSE $0.694\,\mathrm{Nm/kg}$; Left hip $R=0.166$ ($p=0.1654$), DTW distance 9.012, RMSE $0.817\,\mathrm{Nm/kg}$. Despite lower hip RMSE values (in absolute terms) than semi-markerless hip RMSE, the low correlations and large DTW distances indicate poor recovery of waveform timing/shape, consistent with sensitivity of proximal moments to pGRF errors in direction and line of action.

\textbf{Rajagopal model} (fully markerless; pGRF; 114 matched trials). Agreement was again best distally but generally weaker than the semi-markerless case, with only one distal correlation reaching significance: Right ankle $R=0.422$ ($p=0.0583$), DTW distance 6.988, RMSE $1.399\,\mathrm{Nm/kg}$; Left ankle $R=0.533$ ($p=0.0242$), DTW distance 6.017, RMSE $1.338\,\mathrm{Nm/kg}$. Knees were moderate but not statistically significant: Right knee $R=0.376$ ($p=0.0545$), DTW distance 7.303, RMSE $0.978\,\mathrm{Nm/kg}$; Left knee $R=0.451$ ($p=0.0673$), DTW distance 6.635, RMSE $0.988\,\mathrm{Nm/kg}$. Hips remained low and non-significant: Right hip $R=0.294$ ($p=0.1292$), DTW distance 7.993, RMSE $0.716\,\mathrm{Nm/kg}$; Left hip $R=0.219$ ($p=0.1439$), DTW distance 8.375, RMSE $0.759\,\mathrm{Nm/kg}$.

\subsubsection*{Reliability of the standard \texttt{OpenSim} inverse-dynamics implementation (distal-to-proximal degradation)}
The detailed metric patterns above show two consistent effects: (i) \textbf{distal joints (ankle, then knee) are more robust} than proximal joints (hip) under \texttt{SMPL}-driven kinematics, and (ii) \textbf{replacing measured GRF with pGRF reduces agreement} even at distal joints. These observations align with inverse-dynamics error propagation and with known sensitivity of \texttt{OpenSim} ID to imperfectly consistent kinematics and external loads.

\textbf{Inverse dynamics is a sequential balance of distal loads and inertial terms.} In a Newton--Euler formulation, the net moment at a proximal joint must balance the sum of (i) distal external wrenches (GRF/pGRF applied at the foot), (ii) inertial moments from segment accelerations, and (iii) transmitted joint reaction forces and moments from all distal joints and segments. Any mismatch introduced at the foot (CoP, GRF direction, GRF magnitude, or timing) or in distal segment kinematics is carried upward and can grow with each recursion step. Even when ankle and knee waveforms appear qualitatively similar, small distal inconsistencies can accumulate into a large hip-moment error because the hip must balance the compounded distal residuals.

\textbf{The \texttt{OpenSim} ID tool is highly sensitive to kinematic derivatives.} The generalized forces were computed from the coordinates, velocities, and accelerations. In common workflows, velocities and accelerations are obtained by differentiating the coordinate trajectories (often after filtering). Markerless kinematics frequently contains small frame-to-frame jitter in pelvis translation and segment orientation. Differentiation amplifies this jitter, thereby perturbing inertial forces and moments. Because proximal joints must balance the inertial effects of multiple distal segments, this sensitivity is typically expressed most strongly in hip moments.

\textbf{Dynamic inconsistency manifests as residual (unmodelled) forces/torques that primarily contaminate proximal moments.} If the supplied kinematics and applied external wrenches do not satisfy Newton's laws for the chosen rigid-body model (because of measurement noise, scaling error in segment inertial properties, soft-tissue artifacts, imperfect markerless global translation, or pGRF inaccuracies), the inverse-dynamics equations cannot be satisfied without implicit residual wrenches. In \texttt{OpenSim}, these inconsistencies are often reflected as non-physiological ``residual'' forces and torques applied at the pelvis (or expressed through base reactions), and their effect propagates into the computed proximal net moments.

\textbf{Errors in GRF application point and coordinate alignment disproportionately affect the hip.} The moment contribution of the GRF depends on the line of action relative to the joint centers. A CoP shift alters the GRF lever arm about each joint; because the hip is farthest from the foot, the same absolute spatial error produces a larger moment perturbation about the hip than about the ankle. This is why CoP alignment is required in the semi-markerless pipeline and why inaccuracies in pGRF CoP/force direction are particularly detrimental to hip moments.

\textbf{Model scaling and simplifications compound proximal uncertainty.} Both models rely on scaled generic inertial parameters (segment masses, centers of mass, and inertia tensors). Any mismatch between the true subject inertial properties and the scaled model affects inertial terms because proximal moments must balance the inertial contributions of multiple segments; these effects accumulate proximally. Likewise, differences in joint definitions, coordinate conventions, and constrained degrees of freedom can redistribute net moments and change proximal moment sensitivity. Taken together, these factors mean that the standard \texttt{OpenSim} inverse dynamics workflow can be \textbf{comparatively stable for distal moments} while becoming \textbf{increasingly unreliable for proximal joint moments} as one moves away from the foot. 

\subsection*{Graph-based conditional-dependence analysis}

\begin{table}[H]
  \centering

  \begin{subtable}[t]{0.62\linewidth}
    \centering
    \scriptsize
    \setlength{\tabcolsep}{5pt}
    \renewcommand{\arraystretch}{0.92}
    \resizebox{\linewidth}{!}{%
      \begin{tabular}{@{}l l r@{}}
        \toprule
        \textbf{Variable} & \textbf{Level / statistic} & \textbf{Value} \\
        \midrule
        \multicolumn{3}{@{}l}{\textit{Sampling unit}} \\
        Participants                & $n$                        & 105 \\
        Gait segments               & $n$                        & 328 \\
        Segments per participant    & median (IQR)               & 3 (2--4) \\
                                    & range                      & 1--8 \\
        \midrule
        \multicolumn{3}{@{}l}{\textit{Demographics}} \\
        Age at examination (years)  & median (IQR)               & 51.0 (32.5--60.0) \\
        Sex                         & male                       & 72 (69\%) \\
                                    & female                     & 31 (30\%) \\
                                    & missing                    & 2 (2\%) \\
        Time since injury (years)   & median (IQR)               & 7.0 (4.0--25.5) \\
        \midrule
        \multicolumn{3}{@{}l}{\textit{Graph target: functional independence}} \\
        SCIM mobility total         & median (IQR)               & 24.0 (19.0--29.0) \\
                                    & range                      & 5--30 \\
                                    & missing                    & 12 (11\%) \\
        \midrule
        \multicolumn{3}{@{}l}{\textit{Graph target: neurological severity}} \\
        AIS grade                   & D                          & 97 (92\%) \\
                                    & A                          & 3 (3\%) \\
                                    & B                          & 2 (2\%) \\
                                    & C                          & 1 (1\%) \\
                                    & missing                    & 2 (2\%) \\
        \midrule
        \multicolumn{3}{@{}l}{\textit{Graph target: injury category}} \\
        Lesion category             & paraplegic                 & 60 (57\%) \\
                                    & tetraplegic                & 43 (41\%) \\
                                    & missing                    & 2 (2\%) \\
        \midrule
        \multicolumn{3}{@{}l}{\textit{Graph target: injury etiology}} \\
        Etiology                    & sports                     & 32 (30\%) \\
                                    & fall                       & 20 (19\%) \\
                                    & other non-traumatic        & 15 (14\%) \\
                                    & traumatic (unspecified)    & 13 (12\%) \\
                                    & transport                  & 9 (9\%) \\
                                    & degenerative non-traumatic & 4 (4\%) \\
                                    & tumour (benign)            & 3 (3\%) \\
                                    & infection                  & 2 (2\%) \\
                                    & tumour (malignant)         & 2 (2\%) \\
                                    & congenital                 & 1 (1\%) \\
                                    & assault                    & 1 (1\%) \\
                                    & unknown                    & 1 (1\%) \\
                                    & missing                    & 2 (2\%) \\
        \midrule
        \multicolumn{3}{@{}l}{\textit{Ambulatory status}} \\
        Examination status          & stationary                 & 66 (63\%) \\
                                    & ambulatory                 & 36 (34\%) \\
                                    & unknown / missing          & 3 (3\%) \\
        \bottomrule
      \end{tabular}%
    }
    \caption{\textbf{Cohort used for the graph-based conditional-dependence analysis.} Participant-level summary of the 105 individuals contributing the 328 peak-anchored gait segments that enter the graphs, together with the distribution of every clinical variable used as a graph target. Continuous variables are median (IQR); categorical variables are $n$ (\%) of participants; therefore, categories including missing entries sum to 100\%. Because multiple segments come from the same individual, all edge-stability estimates were obtained by participant-level cluster bootstrap.}
    \label{tab:graph_cohort}
  \end{subtable}
  \hfill
  \begin{subtable}[t]{0.35\linewidth}
    \centering
    \scriptsize
    \setlength{\tabcolsep}{6pt}
    \renewcommand{\arraystretch}{0.95}
    \resizebox{\linewidth}{!}{%
      \begin{tabular}{@{}l c c c@{}}
        \toprule
        \textbf{Gait waveform} & \textbf{PC} & $\widehat{\pi}$ & $\bar{w}$ \\
        \midrule
        Vertical GRF, plate 2 & PC5 & 0.99 & 0.076 \\
        Hip flexion, left & PC3 & 0.93 & 0.059 \\
        Hip flexion, right & PC2 & 0.90 & 0.030 \\
        Pelvis list & PC2 & 0.90 & 0.050 \\
        Hip flexion, left & PC2 & 0.87 & 0.029 \\
        Vertical GRF, plate 1 & PC1 & 0.87 & 0.017 \\
        Vertical GRF, plate 2 & PC1 & 0.86 & 0.018 \\
        Hip flexion, left & PC1 & 0.85 & 0.011 \\
        Knee angle, right & PC1 & 0.85 & 0.025 \\
        Ankle angle, left & PC1 & 0.83 & 0.023 \\
        Hip flexion, right & PC4 & 0.83 & 0.072 \\
        Vertical GRF, plate 1 & PC3 & 0.82 & 0.015 \\
        Hip flexion, right & PC1 & 0.81 & 0.009 \\
        Hip flexion, right & PC3 & 0.78 & 0.025 \\
        Ankle angle, right & PC3 & 0.75 & 0.030 \\
        Vertical GRF, plate 1 & PC5 & 0.75 & 0.029 \\
        Ankle angle, right & PC1 & 0.74 & 0.027 \\
        Ankle angle, right & PC4 & 0.71 & 0.022 \\
        Ankle angle, right & PC2 & 0.70 & 0.022 \\
        \bottomrule
      \end{tabular}%
    }
    \caption{\textbf{Markov blanket of SCIM mobility Total} The 19 gait-waveform principal components retaining a direct conditional association with SCIM mobility Total after adjusting for all other retained descriptors, out of 55 candidates (11 waveforms $\times$ 5 PCs). $\widehat{\pi}$ is the participant-level cluster-bootstrap edge stability ($B=300$); only edges with $\widehat{\pi}\ge0.70$ were retained. $\bar{w}$ is the mean symmetrized edge weight across bootstrap replicates. Rows are ordered by decreasing stability. Edge weights index reproducibility of a conditional association, not effect size or causal strength.}
    \label{tab:scim_blanket}
  \end{subtable}

  \caption{\textbf{Graph-based conditional-dependence analysis: cohort and SCIM Markov blanket.} Subtable~\textbf{(a)} shows the cohort and clinical target distributions used to learn the graphs. Subtable~\textbf{(b)} lists the Markov-blanket members of SCIM mobility Total with participant-level cluster-bootstrap stabilities and mean edge weights.}
  \label{tab:graph_cohort_scim_blanket}
\end{table}

We next evaluated whether the markerless waveform descriptors encode clinically interpretable multivariate structure using a sparse conditional-dependence graph learned over waveform principal components and clinical targets. The graph feature set was restricted to the most reliable markerless signals (lower-limb sagittal kinematics, pelvic kinematics, and vertical GRF descriptors). In particular, net joint moments from inverse dynamics were \emph{not} used as graph inputs. (\stabref{tab:semi_markerless_id}) demonstrates substantial distal-to-proximal degradation and strong sensitivity to the external-load source (measured GRF vs pGRF), implying that moment PCs would be more likely to reflect solver/consistency artifacts than stable patient-specific gait structure.

\subsection*{Primary speed-strata clustering: marker-based pipeline}

Analyses were restricted to participants with complete clinical records and valid marker-based gait-feature extraction ($n=172$; Slow $n=36$, Medium $n=89$, Fast $n=47$). The clinical and gait-domain characteristics of the primary speed strata are reported jointly for the marker-based and markerless pipelines in \stabref{tab:clinical_gait_primaryspeed_markerbased} (merged table with modality-specific columns). The table is organized by domain (Clinical; Gait) and, within the Gait domain, by spatiotemporal descriptors, sagittal-plane lower-limb kinematics, gait-quality indices (Gait Variable Score components and cyclogram-derived metrics), and ground reaction force features; between-strata statistical comparisons are reported as $p$-values (separately for each modality).

Across the marker-based speed strata, functional mobility differed strongly: SCIM mobility total increased from Slow to Medium to Fast (15.8, 19.7, 22.7; $p=0.001$), consistent with progressively higher gait speed, shorter cycle time, longer stride length, longer step length, shorter stance and swing times, and reduced double-limb support time (all $p=0.001$ for the listed spatiotemporal measures). Range-of-motion measures also differentiated strata, with larger hip and knee flexion ROM in the Fast group (ROM hip flexion 109.6$^\circ$; ROM knee flexion 136.0$^\circ$; both significant across strata), and higher hip flexor strength in the Fast group (LEMS hip flexors 4.9; $p=0.001$).

Kinematic and gait-quality measures demonstrated systematic speed-associated changes. Peak knee flexion increased with walking speed for both dominant and nondominant limbs (both $p=0.001$), while peak ankle dorsiflexion decreased across strata (dominant $p=0.017$; nondominant $p=0.001$). The maximum pelvic anterior tilt was highest in the slow group (11.6$^\circ$) and lower in the medium and fast groups (8.8$^\circ$, 8.1$^\circ$; $p=0.013$). Gait quality indices improved with higher speed strata, as reflected by lower GVS values for hip, knee, and ankle dorsiflexion (all $p=0.001$), reduced SSD for knee--hip and ankle--knee cyclograms (both $p=0.001$), and larger cyclogram areas for both joint-pair couplings (both $p=0.001$), indicating broader joint coordination loops with increasing speed.

Ground reaction force features differed across speed strata in a manner consistent with greater dynamic loading at higher speeds: the minimum anterior GRF became more negative, vertical GRF maxima increased, and medial GRF minima became more negative with increasing speed for both limbs (all listed GRF features $p=0.001$).


\subsubsection*{Primary speed-strata clustering: markerless pipeline}

Analyses were restricted to participants with complete clinical records and valid markerless gait-feature extraction ($n=239$; Slow $n=42$, Medium $n=110$, Fast $n=87$). Clinical and gait-domain characteristics of the primary speed strata for the markerless pipeline are shown in the markerless columns of the merged \stabref{tab:clinical_gait_primaryspeed_markerless}. As in the marker-based pipeline, the clinical function differed across speed strata. The SCIM mobility total increased from Slow to Medium to Fast (15.3, 19.7, 23.6; $p=0.001$). Lower limb impairment measures showed concomitant differences, including larger hip, knee, and ankle dorsiflexion ROM in the Fast group relative to the Slow and Medium groups (ROM hip flexion $p=0.017$; ROM knee flexion $p=0.017$; ROM ankle dorsiflexion $p=0.026$) and higher hip flexor strength with increasing speed (LEMS hip flexors $p=0.001$).

Spatiotemporal variables separated the speed strata with strong monotonic patterns: gait speed increased, cycle time decreased, stride and step lengths increased, stance and swing times decreased, and double limb support time decreased (all $p=0.001$ for the listed measures). Stride width also differed across speed strata ($p=0.015$), indicating a detectable change in the mediolateral foot placement with speed.

Kinematic and gait quality indices exhibited speed-associated differences, although the directionality differed from the marker-based pipeline for some gait quality outcomes. Peak knee flexion increased with speed for the dominant limb ($p=0.032$). GVS hip flexion and GVS knee flexion for the dominant limb increased across strata (GVS hip flexion $p=0.038$; GVS knee flexion $p=0.001$), and cyclogram areas increased with speed (knee--hip $p=0.001$; ankle--knee $p=0.005$). SSD measures also increased with speed (knee--hip $p=0.001$; ankle--knee $p=0.030$), indicating larger deviations of joint-pair trajectories from the reference patterns as speed increased in this markerless representation.

\begin{figure}[H]
  \vspace*{-5.8em}
  \centering
  \includegraphics[width=0.83\linewidth]{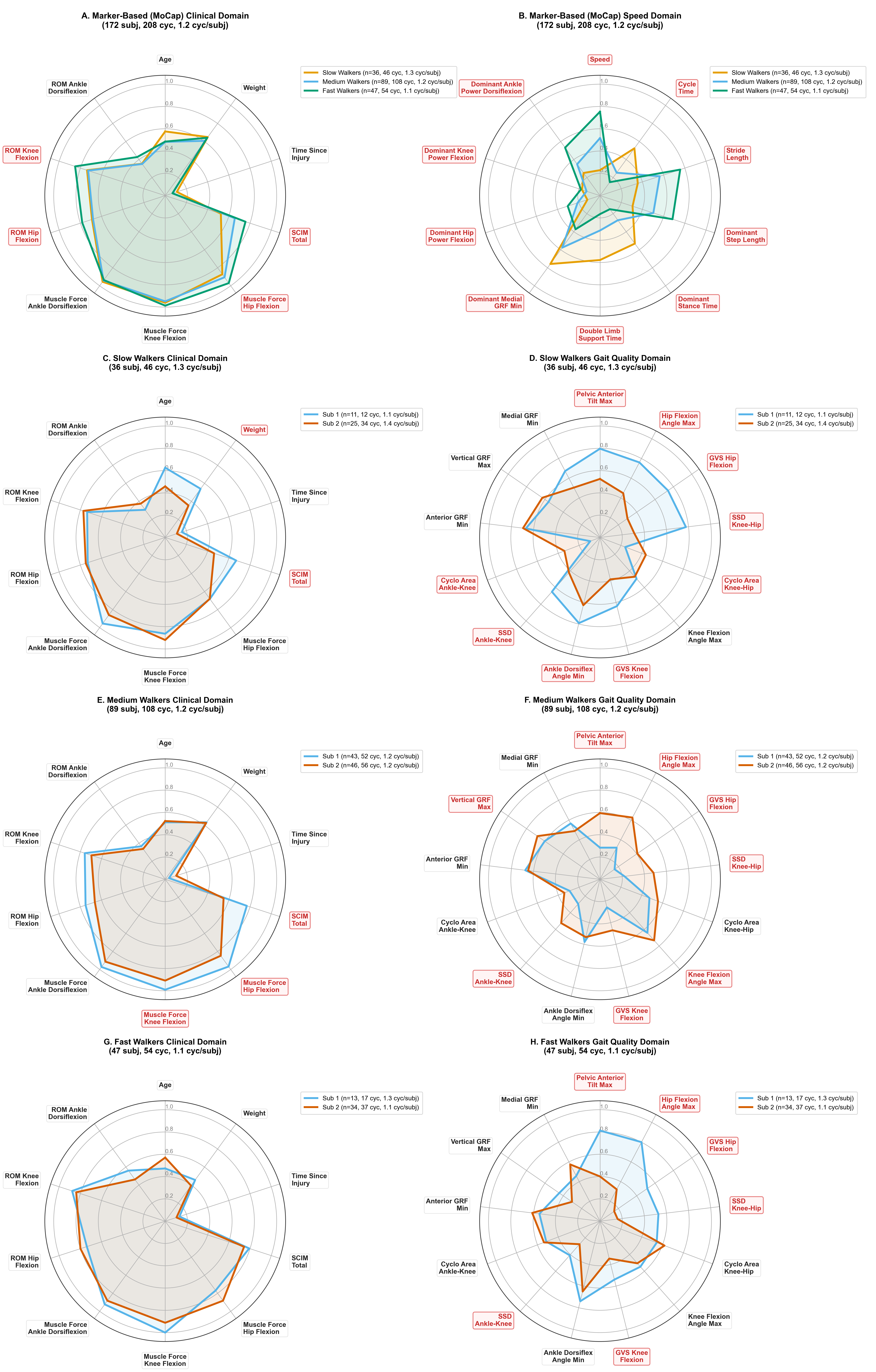}
  \caption{\small{\textbf{Unified spider-plot summary of clustering phenotypes (marker-based pipeline).} Radar plots summarise the multivariate cluster profiles on a shared normalised scale (larger radius indicates a higher normalised value). \textbf{(A)} Primary (speed-strata) clustering: clinical-domain profile across \textbf{Slow}, \textbf{Medium}, and \textbf{Fast} groups. \textbf{(B)} Primary clustering: speed/spatiotemporal gait-domain profile across \textbf{Slow}, \textbf{Medium}, and \textbf{Fast}. \textbf{(C--D)} Within-stratum sub-clustering in the \textbf{Slow} group: \textbf{(C)} clinical-domain profile and \textbf{(D)} gait-quality-domain profile comparing Sub-cluster~1 vs Sub-cluster~2. \textbf{(E--F)} Within-stratum sub-clustering in the \textbf{Medium} group: \textbf{(E)} clinical-domain profile and \textbf{(F)} gait-quality-domain profile comparing Sub-cluster~1 vs Sub-cluster~2. \textbf{(G--H)} Within-stratum sub-clustering in the \textbf{Fast} group: \textbf{(G)} clinical-domain profile and \textbf{(H)} gait-quality-domain profile comparing Sub-cluster~1 vs Sub-cluster~2.}} 
  \label{fig:supp_spider_markerbased_unified}
\end{figure}

Among the GRF-derived measures, the minimum anterior GRF differed across strata for both limbs (dominant $p=0.001$; non-dominant $p=0.002$), consistent with increasing braking forces at higher speeds, while the vertical and medial components showed weaker or non-significant stratification in this markerless-derived feature set for the maxima and minima reported. \sfigref{fig:supp_spider_markerbased_unified} provides a compact multivariate visualisation of the marker-based clustering results using radar (spider) plots on a shared normalised scale, where larger radii indicate higher normalised values for the corresponding features. 

Panels A and B summarize the primary speed strata phenotypes. Panel A aggregates the clinical-domain signature of each speed stratum into a multivariate profile. In line with the quantitative patterns in \stabref{tab:clinical_gait_primaryspeed_markerbased}, the Fast stratum exhibits a profile consistent with higher functional mobility and greater available joint excursion, while the Slow stratum exhibits a profile consistent with reduced mobility and more constrained lower-limb joint excursion. Panel B depicts the spatiotemporal speed-related phenotype: the Fast stratum occupies the largest radius on features aligned with faster locomotion and longer steps, while the Slow stratum occupies the largest radius on features aligned with prolonged gait cycle timing and increased support-phase timing.

Panels C through H visualize within-stratum sub-cluster contrasts after conditioning on walking-speed stratum. For each speed stratum, the clinical-domain panel indicates which functional and impairment measures contribute most strongly to the separation between Sub-cluster~1 and Sub-cluster~2, while the paired gait-quality panel indicates which deviation-based and coordination-based indices differentiate the sub-phenotypes. Across strata, the spider plots separate speed-driven effects captured in Panels A and B from phenotype-driven effects within a given speed stratum captured in Panels C through H, enabling identification of feature blocks that consistently discriminate sub-cluster membership.

\subsubsection*{Within-speed sub-cluster analysis}

To complement the primary speed-strata analysis (\stabref{tab:clinical_gait_primaryspeed_markerbased} and \stabref{tab:clinical_gait_primaryspeed_markerless}), we performed secondary clustering \emph{within} each speed stratum to identify distinct gait-mechanics phenotypes among participants with comparable walking speed.

\begin{table*}[ht]
\centering
\tiny
\setlength{\tabcolsep}{0.8pt}
\renewcommand{\arraystretch}{1.0}

\definecolor{MBslowSig}{RGB}{217,234,247} 
\definecolor{MLslowSig}{RGB}{255,231,204} 

\resizebox{\textwidth}{!}{%
\begin{tabular}{cl*{4}{c}c*{4}{c}c}
\toprule
\textbf{Domain} & \textbf{Parameter} & \multicolumn{4}{c}{\textbf{Marker-based}} & \textbf{$p$} & \multicolumn{4}{c}{\textbf{Markerless}} & \textbf{$p$} \\
\cmidrule(lr){3-6}\cmidrule(lr){8-11}
 &  & \textbf{Total} & \textbf{Sub 1} & \textbf{Sub 2} & \textbf{$n$} &  & \textbf{Total} & \textbf{Sub 1} & \textbf{Sub 2} & \textbf{$n$} &  \\
\midrule
\multirow{15}{*}{\rotatebox{90}{\textbf{Clinical domain}}}
& Total & $n=36$ & $n=11$ & $n=25$ &  &  & $n=42$ & $n=24$ & $n=18$ &  &  \\
& Age [years] & 51.6 (15.3) & 59.1 (13.0) & 48.4 (15.3) &  & 0.169 & 49.7 (16.6) & 53.2 (15.6) & 45.1 (17.3) &  & 0.242 \\
& Sex & 66.7\% male & 72.7\% male & 64.0\% male &  & 0.898 & 66.7\% male & 75.0\% male & 55.6\% male &  & 0.321 \\
& Weight [kg] & \cellcolor{MBslowSig}75.4 (14.2) & \cellcolor{MBslowSig}\textbf{83.3 (12.3)} & \cellcolor{MBslowSig}\textbf{71.9 (13.8)} &  & \cellcolor{MBslowSig}\textbf{0.028} & 73.9 (14.4) & 74.8 (11.1) & 72.8 (18.1) &  & 0.394 \\
& Level of injury & 55.9\% paraplegic & 44.4\% paraplegic & 60.0\% paraplegic &  & 0.679 & 51.3\% paraplegic & 45.5\% paraplegic & 58.8\% paraplegic &  & 0.613 \\
& Diagnosis & 72.2\% traumatic & 63.6\% traumatic & 76.0\% traumatic &  & 0.720 & 76.2\% traumatic & 83.3\% traumatic & 66.7\% traumatic &  & 0.374 \\
& ASIA & 86.1\% d & 81.8\% d & 88.0\% d &  & 0.117 & 88.1\% d & 83.3\% d & 94.4\% d &  & 0.487 \\
& SCIM mobility total & 15.8 (8.5) & 20.1 (7.0) & 13.8 (8.5) &  & 0.050 & 15.3 (8.7) & 16.7 (6.9) & 13.4 (10.6) &  & 0.332 \\
& Time since injury [months] & 60.7 (122.0) & 76.3 (125.7) & 53.8 (122.4) &  & 0.214 & 53.3 (104.8) & 38.4 (66.6) & 73.2 (140.5) &  & 0.868 \\
& ROM Hip flexion [$^\circ$] & 96.7 (36.1) & 95.0 (32.3) & 97.4 (38.2) &  & 0.228 & 93.3 (39.8) & 93.1 (37.3) & 93.6 (44.1) &  & 0.367 \\
& ROM Knee flexion [$^\circ$] & 117.9 (43.7) & 114.1 (39.5) & 119.6 (46.1) &  & 0.086 & 114.0 (48.1) & 112.7 (44.6) & 115.8 (53.8) &  & 0.050 \\
& ROM Ankle dorsiflexion [$^\circ$] & 14.2 (9.9) & 12.3 (7.5) & 15.0 (10.8) &  & 0.452 & 13.5 (9.7) & 12.9 (9.7) & 14.2 (9.9) &  & 0.514 \\
& LEMS Hip flexors (0--5) & 4.4 (0.6) & 4.4 (0.7) & 4.4 (0.6) &  & 0.985 & 4.5 (0.6) & 4.5 (0.6) & 4.3 (0.6) &  & 0.239 \\
& LEMS Knee flexors (0--5) & 4.8 (0.5) & 4.7 (0.5) & 4.8 (0.5) &  & 0.313 & 4.8 (0.5) & 4.9 (0.3) & 4.7 (0.6) &  & 0.205 \\
& LEMS Ankle dorsal flexors (0--5) & 4.8 (0.5) & 4.9 (0.3) & 4.7 (0.5) &  & 0.306 & 4.7 (0.4) & 4.8 (0.4) & 4.7 (0.5) &  & 0.854 \\
\midrule
\multirow{18}{*}{\rotatebox{90}{\textbf{Gait domain}}}
& \multicolumn{10}{c}{\textbf{Spatiotemporal}} \\
& Gait speed [m/s] & 0.7 (0.1) & 0.7 (0.1) & 0.7 (0.1) &  & 0.492 & 0.7 (0.1) & 0.7 (0.1) & 0.7 (0.1) &  & 0.849 \\
& Cycle time [s] & 1.4 (0.1) & 1.4 (0.2) & 1.4 (0.1) &  & 0.372 & 1.4 (0.1) & 1.4 (0.1) & 1.4 (0.1) &  & 0.970 \\
& Stride length [m] & 1.0 (0.1) & 0.9 (0.1) & 1.0 (0.1) &  & 0.080 & 1.0 (0.1) & 1.0 (0.1) & 1.0 (0.1) &  & 0.732 \\
& Dominant step length [m] & \cellcolor{MBslowSig}0.5 (0.1) & \cellcolor{MBslowSig}\textbf{0.4 (0.1)} & \cellcolor{MBslowSig}\textbf{0.5 (0.1)} &  & \cellcolor{MBslowSig}\textbf{0.036} & 0.5 (0.1) & 0.5 (0.1) & 0.5 (0.1) &  & 0.517 \\
\cmidrule(lr){2-12}
& \multicolumn{10}{c}{\textbf{Kinematics}} \\
& \textbf{Hip Flexion Max [$^\circ$] (Dom)} & \cellcolor{MBslowSig}38.1 (8.5) & \cellcolor{MBslowSig}\textbf{45.5 (2.4)} & \cellcolor{MBslowSig}\textbf{34.9 (8.2)} &  & \cellcolor{MBslowSig}\textbf{0.001} & \cellcolor{MLslowSig}38.2 (10.3) & \cellcolor{MLslowSig}\textbf{34.5 (7.4)} & \cellcolor{MLslowSig}\textbf{43.1 (11.7)} &  & \cellcolor{MLslowSig}\textbf{0.015} \\
& Knee Flexion Max [$^\circ$] (Dom) & 62.6 (7.2) & 63.2 (6.7) & 62.3 (7.5) &  & 0.837 & \cellcolor{MLslowSig}52.8 (14.9) & \cellcolor{MLslowSig}\textbf{47.3 (13.4)} & \cellcolor{MLslowSig}\textbf{60.2 (13.9)} &  & \cellcolor{MLslowSig}\textbf{0.005} \\
& Ankle Dorsiflexion Max [$^\circ$] (Dom) & \cellcolor{MBslowSig}13.2 (6.0) & \cellcolor{MBslowSig}\textbf{16.6 (4.8)} & \cellcolor{MBslowSig}\textbf{11.7 (5.9)} &  & \cellcolor{MBslowSig}\textbf{0.020} & 23.6 (6.7) & 22.0 (7.8) & 25.9 (4.2) &  & 0.073 \\
\cmidrule(lr){2-12}
& \multicolumn{10}{c}{\textbf{Gait Quality Indices}} \\
& GVS Hip Flexion (Dom) & \cellcolor{MBslowSig}11.8 (5.3) & \cellcolor{MBslowSig}\textbf{17.5 (2.5)} & \cellcolor{MBslowSig}\textbf{9.3 (4.0)} &  & \cellcolor{MBslowSig}\textbf{0.001} & 22.1 (3.8) & 21.1 (3.1) & 23.5 (4.3) &  & 0.055 \\
& \textbf{GVS Knee Flexion (Dom)} & \cellcolor{MBslowSig}11.7 (3.2) & \cellcolor{MBslowSig}\textbf{13.4 (3.1)} & \cellcolor{MBslowSig}\textbf{10.9 (3.0)} &  & \cellcolor{MBslowSig}\textbf{0.028} & \cellcolor{MLslowSig}27.5 (3.7) & \cellcolor{MLslowSig}\textbf{26.2 (3.0)} & \cellcolor{MLslowSig}\textbf{29.2 (3.8)} &  & \cellcolor{MLslowSig}\textbf{0.009} \\
& \textbf{Cyclogram Area Knee--Hip (Dom)} & \cellcolor{MBslowSig}1250.0 (318.2) & \cellcolor{MBslowSig}\textbf{1057.3 (268.6)} & \cellcolor{MBslowSig}\textbf{1334.8 (305.1)} &  & \cellcolor{MBslowSig}\textbf{0.016} & \cellcolor{MLslowSig}966.1 (482.5) & \cellcolor{MLslowSig}\textbf{757.4 (397.8)} & \cellcolor{MLslowSig}\textbf{1244.5 (451.3)} &  & \cellcolor{MLslowSig}\textbf{0.002} \\
\cmidrule(lr){2-12}
& \multicolumn{10}{c}{\textbf{Ground Reaction Forces}} \\
& Anterior GRF Min [BW] (Dom) & -0.0 (0.0) & -0.0 (0.0) & -0.0 (0.0) &  & 0.514 & \cellcolor{MLslowSig}-0.0 (0.0) & \cellcolor{MLslowSig}\textbf{-0.0 (0.0)} & \cellcolor{MLslowSig}\textbf{-0.1 (0.0)} &  & \cellcolor{MLslowSig}\textbf{0.002} \\
& Vertical GRF Max [BW] (Dom) & 0.9 (0.2) & 0.9 (0.2) & 1.0 (0.2) &  & 0.514 & \cellcolor{MLslowSig}0.8 (0.2) & \cellcolor{MLslowSig}\textbf{0.7 (0.2)} & \cellcolor{MLslowSig}\textbf{1.0 (0.1)} &  & \cellcolor{MLslowSig}\textbf{0.001} \\
& Medial GRF Min [BW] (Dom) & -0.1 (0.0) & -0.1 (0.0) & -0.1 (0.0) &  & 0.054 & \cellcolor{MLslowSig}-0.0 (0.0) & \cellcolor{MLslowSig}\textbf{-0.0 (0.0)} & \cellcolor{MLslowSig}\textbf{-0.0 (0.0)} &  & \cellcolor{MLslowSig}\textbf{0.001} \\
\bottomrule
\end{tabular}%
}

\caption{\small{\textbf{Sub-cluster analysis within the Slow walkers stratum (marker-based and markerless).} Within-stratum comparisons between Sub-cluster~1 and Sub-cluster~2 for the marker-based pipeline (Total $n=36$, Sub 1 $n=11$, Sub 2 $n=25$) and the markerless pipeline (Total $n=42$, Sub 1 $n=24$, Sub 2 $n=18$). Continuous variables are reported as mean (SD) and categorical variables as \%. $p$-values correspond to within-stratum tests comparing Sub-cluster~1 vs Sub-cluster~2. Rows with $p<0.05$ are shown in \textbf{bold}. Row ordering and domain structure match \stabref{tab:clinical_gait_primaryspeed_markerless}.}} 
\label{tab:supp_subclusters_slow_mb}
\label{tab:supp_subclusters_slow_ml}
\end{table*}

Within-stratum results are reported separately for marker-based and markerless pipelines and for each stratum: Slow (marker-based: \stabref{tab:supp_subclusters_slow_mb}; markerless: \stabref{tab:supp_subclusters_slow_ml}), Medium (marker-based: \stabref{tab:supp_subclusters_medium_mb}; markerless: \stabref{tab:supp_subclusters_medium_ml}), and Fast (marker-based: \stabref{tab:supp_subclusters_fast_mb}; markerless: \stabref{tab:supp_subclusters_fast_ml}). Across all sub-cluster tables, continuous variables are reported as mean (SD) and categorical variables as \%; $p$-values correspond to within-stratum comparisons between Sub-cluster~1 and Sub-cluster~2, using the same domain hierarchy and row ordering as the primary speed-strata tables. Within-stratum comparisons reveal phenotype separation that is not explained by walking speed alone.

\textbf{Slow stratum.} In the marker-based pipeline (\stabref{tab:supp_subclusters_slow_mb}), Sub-cluster~1 was heavier than Sub-cluster~2 (83.3 vs 71.9~kg; $p=0.028$) and exhibited shorter dominant step length (0.4 vs 0.5~m; $p=0.036$). The same sub-cluster also showed larger dominant hip flexion maximum (45.5$^\circ$ vs 34.9$^\circ$; $p=0.001$) and larger dominant ankle dorsiflexion maximum (16.6$^\circ$ vs 11.7$^\circ$; $p=0.020$), coupled with larger deviation-based gait-quality indices for hip and knee flexion (GVS hip flexion 17.5 vs 9.3; $p=0.001$; GVS knee flexion 13.4 vs 10.9; $p=0.028$). The cyclogram area for the knee--hip coupling was smaller in Sub-cluster~1 (1057.3 vs 1334.8; $p=0.016$), consistent with a more constrained joint-coupling loop despite larger peak hip flexion. In the markerless pipeline (\stabref{tab:supp_subclusters_slow_ml}), sub-cluster separation was dominated by kinematic and kinetic features: Sub-cluster~2 showed larger dominant hip flexion maximum (43.1$^\circ$ vs 34.5$^\circ$; $p=0.015$) and larger dominant knee flexion maximum (60.2$^\circ$ vs 47.3$^\circ$; $p=0.005$), accompanied by higher knee-flexion deviation (GVS knee flexion 29.2 vs 26.2; $p=0.009$), larger knee--hip cyclogram area (1244.5 vs 757.4; $p=0.002$), and stronger GRF magnitudes (more negative minimum anterior GRF, higher vertical GRF maximum, and more negative minimum medial GRF; all significant within the reported GRF block).

\begin{table*}[ht]
\centering
\tiny
\setlength{\tabcolsep}{0.8pt}
\renewcommand{\arraystretch}{1.0}

\definecolor{MBmedSig}{RGB}{217,234,247} 
\definecolor{MLmedSig}{RGB}{255,231,204} 

\resizebox{\textwidth}{!}{%
\begin{tabular}{clccc c ccc c}
\toprule
\textbf{Domain} & \textbf{Parameter} & \multicolumn{3}{c}{\textbf{Marker-based}} & \textbf{$p$} & \multicolumn{3}{c}{\textbf{Markerless}} & \textbf{$p$} \\
\cmidrule(lr){3-5}\cmidrule(lr){7-9}
 &  & \textbf{Total} & \textbf{Sub 1} & \textbf{Sub 2} &  & \textbf{Total} & \textbf{Sub 1} & \textbf{Sub 2} &  \\
\midrule
\multirow{14}{*}{\rotatebox{90}{\textbf{Clinical domain}}}
& Age [years] & 44.7 (17.1) & 44.3 (16.8) & 45.1 (17.5) & 0.805 & 45.8 (16.6) & 45.1 (16.7) & 46.1 (16.6) & 0.841 \\
& Sex & 70.8\% male & 72.1\% male & 69.6\% male & 0.977 & 76.4\% male & 75.9\% male & 76.5\% male & 1.000 \\
& Weight [kg] & 70.5 (22.4) & 70.9 (18.9) & 70.1 (25.4) & 0.805 & 73.5 (23.2) & 70.9 (26.3) & 74.4 (22.1) & 0.578 \\
& Level of injury & 60.9\% paraplegic & 63.4\% paraplegic & 58.7\% paraplegic & 0.818 & 60.2\% paraplegic & 55.2\% paraplegic & 62.0\% paraplegic & 0.672 \\
& Diagnosis & 69.3\% traumatic & 62.8\% traumatic & 75.6\% traumatic & 0.286 & 67.9\% traumatic & 65.5\% traumatic & 68.8\% traumatic & 0.930 \\
& ASIA & 92.1\% d & 88.4\% d & 95.7\% d & 0.571 & 89.9\% d & 79.3\% d & 93.8\% d & 0.101 \\
& SCIM mobility total & \cellcolor{MBmedSig}19.7 (9.7) & \cellcolor{MBmedSig}\textbf{23.1 (7.7)} & \cellcolor{MBmedSig}\textbf{16.5 (10.3)} & \cellcolor{MBmedSig}\textbf{0.002} & 19.7 (9.3) & 18.7 (10.3) & 20.0 (9.0) & 0.686 \\
& Time since injury [months] & 39.3 (81.4) & 19.8 (40.8) & 57.4 (103.4) & 0.073 & 50.1 (94.9) & 70.2 (114.4) & 42.9 (86.6) & 0.224 \\
& ROM Hip flexion [$^\circ$] & 95.3 (41.0) & 101.4 (34.9) & 89.6 (45.7) & 0.273 & 96.0 (40.9) & 90.0 (43.2) & 98.2 (40.1) & 0.072 \\
& ROM Knee flexion [$^\circ$] & 116.2 (52.0) & 121.4 (47.3) & 111.4 (56.2) & 0.275 & 117.2 (51.1) & 114.3 (53.9) & 118.2 (50.4) & 0.511 \\
& ROM Ankle dorsiflexion [$^\circ$] & 14.0 (9.2) & 14.7 (9.2) & 13.5 (9.2) & 0.977 & 13.5 (9.0) & 14.7 (10.5) & 13.1 (8.4) & 0.660 \\
& LEMS Hip flexors (0--5) & \cellcolor{MBmedSig}4.5 (1.0) & \cellcolor{MBmedSig}\textbf{4.8 (0.4)} & \cellcolor{MBmedSig}\textbf{4.2 (1.3)} & \cellcolor{MBmedSig}\textbf{0.008} & 4.5 (0.9) & 4.4 (1.1) & 4.6 (0.9) & 0.598 \\
& LEMS Knee flexors (0--5) & \cellcolor{MBmedSig}4.7 (0.9) & \cellcolor{MBmedSig}\textbf{5.0 (0.2)} & \cellcolor{MBmedSig}\textbf{4.5 (1.2)} & \cellcolor{MBmedSig}\textbf{0.018} & 4.8 (0.8) & 4.7 (1.0) & 4.8 (0.7) & 0.698 \\
& LEMS Ankle dorsal flexors (0--5) & 4.7 (0.9) & 4.9 (0.5) & 4.6 (1.1) & 0.168 & 4.7 (0.8) & 4.6 (1.0) & 4.7 (0.8) & 0.718 \\
\midrule
\multirow{17}{*}{\rotatebox{90}{\textbf{Gait domain}}}
& \multicolumn{8}{c}{\textbf{Spatiotemporal}} \\
& Gait speed [m/s] & 1.0 (0.1) & 1.0 (0.1) & 1.0 (0.1) & 0.740 & 1.0 (0.1) & 1.0 (0.1) & 1.0 (0.1) & 0.328 \\
& Cycle time [s] & 1.2 (0.1) & 1.2 (0.1) & 1.2 (0.1) & 0.918 & 1.2 (0.1) & 1.2 (0.1) & 1.2 (0.1) & 0.113 \\
& Stride length [m] & 1.2 (0.1) & 1.2 (0.1) & 1.2 (0.1) & 0.860 & \cellcolor{MLmedSig}1.2 (0.1) & \cellcolor{MLmedSig}\textbf{1.1 (0.1)} & \cellcolor{MLmedSig}\textbf{1.2 (0.1)} & \cellcolor{MLmedSig}\textbf{0.003} \\
& Dominant step length [m] & 0.6 (0.1) & 0.6 (0.1) & 0.6 (0.1) & 0.690 & \cellcolor{MLmedSig}0.6 (0.1) & \cellcolor{MLmedSig}\textbf{0.5 (0.1)} & \cellcolor{MLmedSig}\textbf{0.6 (0.1)} & \cellcolor{MLmedSig}\textbf{0.022} \\
\cmidrule(lr){2-10}
& \multicolumn{8}{c}{\textbf{Kinematics}} \\
& Hip Flexion Max [$^\circ$] (Dom) & \cellcolor{MBmedSig}36.4 (7.7) & \cellcolor{MBmedSig}\textbf{30.6 (4.8)} & \cellcolor{MBmedSig}\textbf{41.9 (5.4)} & \cellcolor{MBmedSig}\textbf{0.001} & 38.1 (8.9) & 35.4 (5.3) & 39.1 (9.7) & 0.137 \\
& Knee Flexion Max [$^\circ$] (Dom) & \cellcolor{MBmedSig}66.6 (7.0) & \cellcolor{MBmedSig}\textbf{64.5 (7.3)} & \cellcolor{MBmedSig}\textbf{68.6 (6.1)} & \cellcolor{MBmedSig}\textbf{0.019} & 55.5 (12.4) & 52.2 (10.2) & 56.7 (12.9) & 0.111 \\
& Ankle Dorsiflexion Max [$^\circ$] (Dom) & 12.9 (4.9) & 12.3 (4.1) & 13.4 (5.4) & 0.322 & 24.7 (4.7) & 24.3 (5.0) & 24.8 (4.5) & 0.689 \\
\cmidrule(lr){2-10}
& \multicolumn{8}{c}{\textbf{Gait Quality Indices}} \\
& \textbf{GVS Hip Flexion (Dom)} & \cellcolor{MBmedSig}8.9 (4.7) & \cellcolor{MBmedSig}\textbf{5.8 (2.2)} & \cellcolor{MBmedSig}\textbf{11.9 (4.6)} & \cellcolor{MBmedSig}\textbf{0.001} & \cellcolor{MLmedSig}22.7 (3.3) & \cellcolor{MLmedSig}\textbf{21.1 (2.2)} & \cellcolor{MLmedSig}\textbf{23.3 (3.4)} & \cellcolor{MLmedSig}\textbf{0.001} \\
& \textbf{GVS Knee Flexion (Dom)} & \cellcolor{MBmedSig}9.5 (3.9) & \cellcolor{MBmedSig}\textbf{7.6 (2.8)} & \cellcolor{MBmedSig}\textbf{11.3 (4.0)} & \cellcolor{MBmedSig}\textbf{0.001} & \cellcolor{MLmedSig}28.0 (3.9) & \cellcolor{MLmedSig}\textbf{26.5 (3.4)} & \cellcolor{MLmedSig}\textbf{28.6 (4.0)} & \cellcolor{MLmedSig}\textbf{0.023} \\
& Cyclogram Area Knee--Hip (Dom) & 1409.4 (296.2) & 1350.4 (276.0) & 1464.7 (306.5) & 0.063 & \cellcolor{MLmedSig}1144.6 (498.6) & \cellcolor{MLmedSig}\textbf{929.2 (406.3)} & \cellcolor{MLmedSig}\textbf{1221.8 (508.0)} & \cellcolor{MLmedSig}\textbf{0.003} \\
\cmidrule(lr){2-10}
& \multicolumn{8}{c}{\textbf{Ground Reaction Forces}} \\
& Anterior GRF Min [BW] (Dom) & -0.0 (0.0) & -0.0 (0.0) & -0.0 (0.0) & 0.579 & \cellcolor{MLmedSig}-0.1 (0.0) & \cellcolor{MLmedSig}\textbf{-0.0 (0.0)} & \cellcolor{MLmedSig}\textbf{-0.1 (0.0)} & \cellcolor{MLmedSig}\textbf{0.001} \\
& \textbf{Vertical GRF Max [BW] (Dom)} & \cellcolor{MBmedSig}1.1 (0.1) & \cellcolor{MBmedSig}\textbf{1.0 (0.2)} & \cellcolor{MBmedSig}\textbf{1.1 (0.1)} & \cellcolor{MBmedSig}\textbf{0.031} & \cellcolor{MLmedSig}0.9 (0.2) & \cellcolor{MLmedSig}\textbf{0.7 (0.1)} & \cellcolor{MLmedSig}\textbf{1.0 (0.1)} & \cellcolor{MLmedSig}\textbf{0.001} \\
& Medial GRF Min [BW] (Dom) & -0.1 (0.0) & -0.1 (0.0) & -0.1 (0.0) & 0.182 & \cellcolor{MLmedSig}-0.0 (0.0) & \cellcolor{MLmedSig}\textbf{-0.0 (0.0)} & \cellcolor{MLmedSig}\textbf{-0.0 (0.0)} & \cellcolor{MLmedSig}\textbf{0.001} \\
\bottomrule
\end{tabular}%
}

\caption{\small{\textbf{Sub-cluster analysis within the Medium walkers stratum (marker-based and markerless).} Within-stratum comparisons between Sub-cluster~1 and Sub-cluster~2 for the marker-based pipeline (Total $n=89$, Sub 1 $n=43$, Sub 2 $n=46$) and the markerless pipeline (Total $n=110$, Sub 1 $n=29$, Sub 2 $n=81$). Continuous variables are reported as mean (SD) and categorical variables as \%. $p$-values correspond to within-stratum tests comparing Sub-cluster~1 vs Sub-cluster~2. Rows with $p<0.05$ are shown in \textbf{bold} and highlighted with modality-specific colors. Row ordering and domain structure match \stabref{tab:clinical_gait_primaryspeed_markerless}.}} 
\label{tab:supp_subclusters_medium_mb}
\label{tab:supp_subclusters_medium_ml}
\end{table*}

\textbf{Medium stratum.} In the marker-based pipeline (\stabref{tab:supp_subclusters_medium_mb}), clinical-domain separation involved both function and strength: Sub-cluster~1 had higher SCIM mobility total (23.1 vs 16.5; $p=0.002$) and higher LEMS for hip and knee flexors (both significant), indicating a higher functional and strength profile at the same walking-speed stratum. The gait-domain separation showed that Sub-cluster~1 had lower dominant hip flexion maximum (30.6$^\circ$ vs 41.9$^\circ$; $p=0.001$) and lower deviation-based indices (GVS hip flexion 5.8 vs 11.9; $p=0.001$; GVS knee flexion 7.6 vs 11.3; $p=0.001$), while Sub-cluster~2 exhibited larger vertical GRF maximum (1.1 vs 1.0~BW; $p=0.031$). In the markerless pipeline (\stabref{tab:supp_subclusters_medium_ml}), no within-stratum variables reached the reported significance threshold, indicating weaker or more heterogeneous phenotype separation in this stratum for the subset of features listed.

\textbf{Fast stratum.} In the marker-based pipeline (\stabref{tab:supp_subclusters_fast_mb}), sub-clusters were separated primarily by dominant hip kinematics and associated deviation indices. Sub-cluster ~1 exhibited larger dominant hip flexion maximum (44.4$^\circ$ vs 33.0$^\circ$; $p=0.001$) and higher deviation-based gait-quality indices for hip and knee flexion (both significant), while spatiotemporal and GRF measures showed no significant separation within the reported set. In the markerless pipeline (\stabref{tab:supp_subclusters_fast_ml}), Sub-cluster~2 displayed consistently larger peak joint excursions (dominant hip, knee, and ankle dorsiflexion maxima; all $p=0.001$), higher deviation-based indices for hip and knee flexion (both $p=0.001$), larger knee--hip cyclogram area ($p=0.001$), and a higher vertical GRF maximum ($p=0.009$), indicating a higher-amplitude kinematic and loading phenotype within the Fast speed stratum.

\begin{table*}[ht]
\centering
\tiny
\setlength{\tabcolsep}{0.8pt}
\renewcommand{\arraystretch}{1.0}

\definecolor{MBfastSig}{RGB}{217,234,247} 
\definecolor{MLfastSig}{RGB}{255,231,204} 

\resizebox{\textwidth}{!}{%
\begin{tabular}{clccc c ccc c}
\toprule
\textbf{Domain} & \textbf{Parameter} & \multicolumn{3}{c}{\textbf{Marker-based}} & \textbf{$p$} & \multicolumn{3}{c}{\textbf{Markerless}} & \textbf{$p$} \\
\cmidrule(lr){3-5}\cmidrule(lr){7-9}
 &  & \textbf{Total} & \textbf{Sub 1} & \textbf{Sub 2} &  & \textbf{Total} & \textbf{Sub 1} & \textbf{Sub 2} &  \\
\midrule
\multirow{14}{*}{\rotatebox{90}{\textbf{Clinical domain}}}
& Age [years] & 44.9 (14.6) & 41.0 (13.2) & 46.4 (15.0) & 0.203 & 44.5 (15.0) & 43.2 (15.4) & 45.7 (14.7) & 0.489 \\
& Sex & 76.6\% male & 69.2\% male & 79.4\% male & 0.725 & 73.6\% male & 81.4\% male & 65.9\% male & 0.163 \\
& Weight [kg] & 74.5 (14.9) & 77.7 (22.2) & 73.3 (11.1) & 0.953 & 71.6 (19.8) & 72.1 (11.9) & 71.2 (25.5) & 0.662 \\
& Level of injury & 57.8\% paraplegic & 75.0\% paraplegic & 51.5\% paraplegic & 0.285 & 60.2\% paraplegic & 63.4\% paraplegic & 57.1\% paraplegic & 0.719 \\
& Diagnosis & 74.5\% traumatic & 84.6\% traumatic & 70.6\% traumatic & 0.540 & 77.0\% traumatic & 79.1\% traumatic & 75.0\% traumatic & 0.844 \\
& ASIA & 87.2\% d & 92.3\% d & 85.3\% d & 0.721 & 89.7\% d & 88.4\% d & 90.9\% d & 0.790 \\
& SCIM mobility total & 22.7 (9.2) & 23.8 (9.8) & 22.3 (9.1) & 0.482 & 23.6 (9.0) & 25.1 (7.9) & 22.2 (9.9) & 0.076 \\
& Time since injury [months] & 37.8 (82.4) & 44.3 (88.2) & 35.4 (81.4) & 0.682 & 32.8 (70.6) & 24.7 (58.3) & 40.7 (80.7) & 0.139 \\
& ROM Hip flexion [$^\circ$] & 109.6 (25.2) & 103.5 (33.8) & 111.9 (21.2) & 0.177 & 107.1 (28.4) & 109.0 (26.2) & 105.3 (30.6) & 0.659 \\
& ROM Knee flexion [$^\circ$] & 136.0 (22.4) & 140.4 (9.5) & 134.3 (25.6) & 0.344 & 131.2 (31.6) & 132.1 (30.3) & 130.3 (33.2) & 0.951 \\
& ROM Ankle dorsiflexion [$^\circ$] & 17.1 (7.6) & 19.6 (9.2) & 16.2 (6.9) & 0.427 & 17.1 (8.7) & 17.1 (8.3) & 17.0 (9.3) & 0.900 \\
& LEMS Hip flexors (0--5) & 4.9 (0.4) & 4.8 (0.4) & 4.9 (0.3) & 0.345 & 4.8 (0.7) & 4.9 (0.5) & 4.7 (0.8) & 0.245 \\
& LEMS Knee flexors (0--5) & 4.9 (0.2) & 5.0 (0.0) & 4.9 (0.3) & 0.286 & 4.9 (0.6) & 5.0 (0.2) & 4.8 (0.8) & 0.098 \\
& LEMS Ankle dorsal flexors (0--5) & 4.7 (0.7) & 4.8 (0.4) & 4.6 (0.8) & 0.987 & 4.7 (0.8) & 4.7 (0.7) & 4.7 (0.9) & 0.945 \\
\midrule
\multirow{17}{*}{\rotatebox{90}{\textbf{Gait domain}}}
& \multicolumn{8}{c}{\textbf{Spatiotemporal}} \\
& Gait speed [m/s] & 1.2 (0.1) & 1.3 (0.1) & 1.2 (0.1) & 0.212 & 1.2 (0.1) & 1.2 (0.1) & 1.2 (0.1) & 0.126 \\
& Cycle time [s] & 1.1 (0.1) & 1.1 (0.1) & 1.1 (0.1) & 0.840 & 1.1 (0.1) & 1.1 (0.1) & 1.1 (0.1) & 0.852 \\
& Stride length [m] & 1.3 (0.1) & 1.4 (0.1) & 1.3 (0.1) & 0.385 & 1.3 (0.1) & 1.3 (0.1) & 1.3 (0.1) & 0.072 \\
& Dominant step length [m] & 0.7 (0.1) & 0.7 (0.1) & 0.7 (0.0) & 0.203 & 0.7 (0.1) & 0.7 (0.0) & 0.6 (0.1) & 0.061 \\
\cmidrule(lr){2-10}
& \multicolumn{8}{c}{\textbf{Kinematics}} \\
& \textbf{Hip Flexion Max [$^\circ$] (Dom)} & \cellcolor{MBfastSig}36.2 (6.6) & \cellcolor{MBfastSig}\textbf{44.4 (3.5)} & \cellcolor{MBfastSig}\textbf{33.0 (4.4)} & \cellcolor{MBfastSig}\textbf{0.001} & \cellcolor{MLfastSig}38.8 (10.8) & \cellcolor{MLfastSig}\textbf{33.0 (5.1)} & \cellcolor{MLfastSig}\textbf{44.5 (11.8)} & \cellcolor{MLfastSig}\textbf{0.001} \\
& Knee Flexion Max [$^\circ$] (Dom) & 68.2 (4.6) & 68.7 (5.4) & 68.0 (4.3) & 0.609 & \cellcolor{MLfastSig}59.6 (14.2) & \cellcolor{MLfastSig}\textbf{49.4 (10.2)} & \cellcolor{MLfastSig}\textbf{69.5 (10.0)} & \cellcolor{MLfastSig}\textbf{0.001} \\
& Ankle Dorsiflexion Max [$^\circ$] (Dom) & 10.6 (4.0) & 12.5 (5.2) & 9.9 (3.3) & 0.094 & \cellcolor{MLfastSig}23.6 (5.7) & \cellcolor{MLfastSig}\textbf{20.3 (4.7)} & \cellcolor{MLfastSig}\textbf{26.9 (4.5)} & \cellcolor{MLfastSig}\textbf{0.001} \\
\cmidrule(lr){2-10}
& \multicolumn{8}{c}{\textbf{Gait Quality Indices}} \\
& \textbf{GVS Hip Flexion (Dom)} & \cellcolor{MBfastSig}7.0 (4.7) & \cellcolor{MBfastSig}\textbf{12.9 (4.6)} & \cellcolor{MBfastSig}\textbf{4.8 (2.0)} & \cellcolor{MBfastSig}\textbf{0.001} & \cellcolor{MLfastSig}23.7 (4.4) & \cellcolor{MLfastSig}\textbf{22.0 (2.3)} & \cellcolor{MLfastSig}\textbf{25.4 (5.3)} & \cellcolor{MLfastSig}\textbf{0.001} \\
& \textbf{GVS Knee Flexion (Dom)} & \cellcolor{MBfastSig}6.5 (2.8) & \cellcolor{MBfastSig}\textbf{8.1 (3.1)} & \cellcolor{MBfastSig}\textbf{5.9 (2.4)} & \cellcolor{MBfastSig}\textbf{0.018} & \cellcolor{MLfastSig}30.2 (5.3) & \cellcolor{MLfastSig}\textbf{27.7 (2.8)} & \cellcolor{MLfastSig}\textbf{32.7 (6.1)} & \cellcolor{MLfastSig}\textbf{0.001} \\
& Cyclogram Area Knee--Hip (Dom) & 1674.2 (324.2) & 1604.6 (428.7) & 1700.9 (277.5) & 0.513 & \cellcolor{MLfastSig}1414.5 (549.9) & \cellcolor{MLfastSig}\textbf{1087.3 (337.9)} & \cellcolor{MLfastSig}\textbf{1734.4 (530.5)} & \cellcolor{MLfastSig}\textbf{0.001} \\
\cmidrule(lr){2-10}
& \multicolumn{8}{c}{\textbf{Ground Reaction Forces}} \\
& Anterior GRF Min [BW] (Dom) & -0.1 (0.0) & -0.1 (0.0) & -0.1 (0.0) & 0.593 & -0.1 (0.0) & -0.1 (0.0) & -0.1 (0.0) & 0.555 \\
& Vertical GRF Max [BW] (Dom) & 1.2 (0.1) & 1.2 (0.1) & 1.1 (0.1) & 0.065 & \cellcolor{MLfastSig}0.9 (0.2) & \cellcolor{MLfastSig}\textbf{0.9 (0.2)} & \cellcolor{MLfastSig}\textbf{1.0 (0.2)} & \cellcolor{MLfastSig}\textbf{0.009} \\
& Medial GRF Min [BW] (Dom) & -0.2 (0.0) & -0.2 (0.0) & -0.2 (0.0) & 0.203 & -0.0 (0.0) & -0.0 (0.0) & -0.0 (0.0) & 0.389 \\
\bottomrule
\end{tabular}%
}

\caption{\small{\textbf{Sub-cluster analysis within the Fast walkers stratum (marker-based and markerless).} Within-stratum comparisons between Sub-cluster~1 and Sub-cluster~2 for the marker-based pipeline (Total $n=47$, Sub 1 $n=13$, Sub 2 $n=34$) and the markerless pipeline (Total $n=87$, Sub 1 $n=43$, Sub 2 $n=44$). Continuous variables are reported as mean (SD) and categorical variables as \%. $p$-values correspond to within-stratum tests comparing Sub-cluster~1 vs Sub-cluster~2. Rows with $p<0.05$ are shown in \textbf{bold} and highlighted with modality-specific colors. Row ordering and domain structure match \stabref{tab:clinical_gait_primaryspeed_markerless}.}} 
\label{tab:supp_subclusters_fast_mb}
\label{tab:supp_subclusters_fast_ml}
\end{table*}

\subsection*{Markerless Cross-speed-combined plots of speed-strata and sub-cluster phenotypes}

For cross-speed cluster analysis, participant-level features were first
standardised to $z$-scores (Equation~\ref{eq:zscore}), then aggregated within each
derived cluster label, and finally displayed on a shared normalised axis so that
phenotype profiles from different speed strata can be visually compared in a
single view. This layer is intended as an interpretability aid (a compact
phenotype ``fingerprint'') rather than an additional inferential step; all
reported statistical comparisons are based on the underlying participant-level
data and the tests described in the main Methods.

\begin{table*}[ht]
\centering
\tiny
\setlength{\tabcolsep}{1.5pt}
\renewcommand{\arraystretch}{1.03}

\definecolor{MBsig}{RGB}{217,234,247} 
\definecolor{MLsig}{RGB}{255,231,204} 

\resizebox{\textwidth}{!}{%
\begin{tabular}{ll*{10}{c}}
\toprule
\textbf{Domain} & \textbf{Parameter} & \multicolumn{5}{c}{\textbf{Marker-based}} & \multicolumn{5}{c}{\textbf{Markerless}} \\
\cmidrule(lr){3-7}\cmidrule(lr){8-12}
 &  & \textbf{Total} & \textbf{Slow} & \textbf{Medium} & \textbf{Fast} & \textbf{$p$} & \textbf{Total} & \textbf{Slow} & \textbf{Medium} & \textbf{Fast} & \textbf{$p$} \\
\midrule
\multirow{15}{*}{\rotatebox{90}{Clinical domain}} & Total & $n=172$ & $n=36$ & $n=89$ & $n=47$ &  & $n=239$ & $n=42$ & $n=110$ & $n=87$ &  \\
 & Age [years] & 46.2 (16.2) & 51.6 (15.3) & 44.7 (17.1) & 44.9 (14.6) & 0.092 & 46.0 (16.1) & 49.7 (16.6) & 45.8 (16.6) & 44.5 (15.0) & 0.179 \\
 & Sex & 71.5\% male & 66.7\% male & 70.8\% male & 76.6\% male & 0.596 & 73.6\% male & 66.7\% male & 76.4\% male & 73.6\% male & 0.479 \\
 & Weight [kg] & 72.6 (19.1) & 75.4 (14.2) & 70.5 (22.4) & 74.5 (14.9) & 0.623 & 72.9 (20.6) & 73.9 (14.4) & 73.5 (23.2) & 71.6 (19.8) & 0.848 \\
 & Level of injury & 59.0\% paraplegic & 55.9\% paraplegic & 60.9\% paraplegic & 57.8\% paraplegic & 0.862 & 58.7\% paraplegic & 51.3\% paraplegic & 60.2\% paraplegic & 60.2\% paraplegic & 0.587 \\
 & Diagnosis & 71.3\% traumatic & 72.2\% traumatic & 69.3\% traumatic & 74.5\% traumatic & 0.813 & 72.7\% traumatic & 76.2\% traumatic & 67.9\% traumatic & 77.0\% traumatic & 0.310 \\
 & ASIA & 89.5\% d & 86.1\% d & 92.1\% d & 87.2\% d & 0.390 & 89.5\% d & 88.1\% d & 89.9\% d & 89.7\% d & 0.429 \\
 & \textbf{SCIM mobility total} & \cellcolor{MBsig}19.7 (9.6) & \cellcolor{MBsig}15.8 (8.5) & \cellcolor{MBsig}19.7 (9.7) & \cellcolor{MBsig}22.7 (9.2) & \cellcolor{MBsig}\textbf{0.001} & \cellcolor{MLsig}20.3 (9.5) & \cellcolor{MLsig}15.3 (8.7) & \cellcolor{MLsig}19.7 (9.3) & \cellcolor{MLsig}23.6 (9.0) & \cellcolor{MLsig}\textbf{0.001} \\
 & Time since injury [months] & 43.4 (91.5) & 60.7 (122.0) & 39.3 (81.4) & 37.8 (82.4) & 0.136 & 44.4 (88.9) & 53.3 (104.8) & 50.1 (94.9) & 32.8 (70.6) & 0.186 \\
 & \textbf{ROM Hip flexion [$^\circ$]} & \cellcolor{MBsig}99.5 (36.6) & \cellcolor{MBsig}96.7 (36.1) & \cellcolor{MBsig}95.3 (41.0) & \cellcolor{MBsig}109.6 (25.2) & \cellcolor{MBsig}\textbf{0.031} & \cellcolor{MLsig}99.6 (36.9) & \cellcolor{MLsig}93.3 (39.8) & \cellcolor{MLsig}96.0 (40.9) & \cellcolor{MLsig}107.1 (28.4) & \cellcolor{MLsig}\textbf{0.017} \\
 & \textbf{ROM Knee flexion [$^\circ$]} & \cellcolor{MBsig}122.0 (44.6) & \cellcolor{MBsig}117.9 (43.7) & \cellcolor{MBsig}116.2 (52.0) & \cellcolor{MBsig}136.0 (22.4) & \cellcolor{MBsig}\textbf{0.029} & \cellcolor{MLsig}121.7 (44.8) & \cellcolor{MLsig}114.0 (48.1) & \cellcolor{MLsig}117.2 (51.1) & \cellcolor{MLsig}131.2 (31.6) & \cellcolor{MLsig}\textbf{0.017} \\
 & ROM Ankle dorsiflexion [$^\circ$] & 14.9 (9.0) & 14.2 (9.9) & 14.0 (9.2) & 17.1 (7.6) & 0.170 & \cellcolor{MLsig}14.8 (9.1) & \cellcolor{MLsig}13.5 (9.7) & \cellcolor{MLsig}13.5 (9.0) & \cellcolor{MLsig}17.1 (8.7) & \cellcolor{MLsig}\textbf{0.026} \\
 & \textbf{LEMS Hip flexors (0--5)} & \cellcolor{MBsig}4.6 (0.8) & \cellcolor{MBsig}4.4 (0.6) & \cellcolor{MBsig}4.5 (1.0) & \cellcolor{MBsig}4.9 (0.4) & \cellcolor{MBsig}\textbf{0.001} & \cellcolor{MLsig}4.6 (0.8) & \cellcolor{MLsig}4.5 (0.6) & \cellcolor{MLsig}4.5 (0.9) & \cellcolor{MLsig}4.8 (0.7) & \cellcolor{MLsig}\textbf{0.001} \\
 & LEMS Knee flexors (0--5) & 4.8 (0.7) & 4.8 (0.5) & 4.7 (0.9) & 4.9 (0.2) & 0.310 & 4.8 (0.7) & 4.8 (0.5) & 4.8 (0.8) & 4.9 (0.6) & 0.128 \\
 & LEMS Ankle dorsal flexors (0--5) & 4.7 (0.8) & 4.8 (0.5) & 4.7 (0.9) & 4.7 (0.7) & 0.650 & 4.7 (0.8) & 4.7 (0.4) & 4.7 (0.8) & 4.7 (0.8) & 0.457 \\
\cmidrule(lr){1-12}
\multirow{35}{*}{\rotatebox{90}{Gait domain}} & \multicolumn{11}{c}{\textbf{Spatiotemporal}} \\
 & \textbf{Gait speed [m/s]} & \cellcolor{MBsig}1.0 (0.2) & \cellcolor{MBsig}0.7 (0.1) & \cellcolor{MBsig}1.0 (0.1) & \cellcolor{MBsig}1.2 (0.1) & \cellcolor{MBsig}\textbf{0.001} & \cellcolor{MLsig}1.0 (0.2) & \cellcolor{MLsig}0.7 (0.1) & \cellcolor{MLsig}1.0 (0.1) & \cellcolor{MLsig}1.2 (0.1) & \cellcolor{MLsig}\textbf{0.001} \\
 & \textbf{Cycle time [s]} & \cellcolor{MBsig}1.2 (0.1) & \cellcolor{MBsig}1.4 (0.1) & \cellcolor{MBsig}1.2 (0.1) & \cellcolor{MBsig}1.1 (0.1) & \cellcolor{MBsig}\textbf{0.001} & \cellcolor{MLsig}1.2 (0.1) & \cellcolor{MLsig}1.4 (0.1) & \cellcolor{MLsig}1.2 (0.1) & \cellcolor{MLsig}1.1 (0.1) & \cellcolor{MLsig}\textbf{0.001} \\
 & \textbf{Stride length [m]} & \cellcolor{MBsig}1.2 (0.2) & \cellcolor{MBsig}1.0 (0.1) & \cellcolor{MBsig}1.2 (0.1) & \cellcolor{MBsig}1.3 (0.1) & \cellcolor{MBsig}\textbf{0.001} & \cellcolor{MLsig}1.2 (0.2) & \cellcolor{MLsig}1.0 (0.1) & \cellcolor{MLsig}1.2 (0.1) & \cellcolor{MLsig}1.3 (0.1) & \cellcolor{MLsig}\textbf{0.001} \\
 & \textbf{Dominant step length [m]} & \cellcolor{MBsig}0.6 (0.1) & \cellcolor{MBsig}0.5 (0.1) & \cellcolor{MBsig}0.6 (0.1) & \cellcolor{MBsig}0.7 (0.1) & \cellcolor{MBsig}\textbf{0.001} & \cellcolor{MLsig}0.6 (0.1) & \cellcolor{MLsig}0.5 (0.1) & \cellcolor{MLsig}0.6 (0.1) & \cellcolor{MLsig}0.7 (0.1) & \cellcolor{MLsig}\textbf{0.001} \\
 & \textbf{Non-dominant step length [m]} & \cellcolor{MBsig}0.6 (0.1) & \cellcolor{MBsig}0.5 (0.1) & \cellcolor{MBsig}0.6 (0.0) & \cellcolor{MBsig}0.7 (0.0) & \cellcolor{MBsig}\textbf{0.001} & \cellcolor{MLsig}0.6 (0.1) & \cellcolor{MLsig}0.5 (0.1) & \cellcolor{MLsig}0.6 (0.1) & \cellcolor{MLsig}0.7 (0.1) & \cellcolor{MLsig}\textbf{0.001} \\
 & \textbf{Dominant stance time [s]} & \cellcolor{MBsig}0.8 (0.1) & \cellcolor{MBsig}0.9 (0.1) & \cellcolor{MBsig}0.7 (0.1) & \cellcolor{MBsig}0.7 (0.1) & \cellcolor{MBsig}\textbf{0.001} & \cellcolor{MLsig}0.8 (0.1) & \cellcolor{MLsig}0.9 (0.1) & \cellcolor{MLsig}0.8 (0.1) & \cellcolor{MLsig}0.7 (0.0) & \cellcolor{MLsig}\textbf{0.001} \\
 & \textbf{Non-dominant stance time [s]} & \cellcolor{MBsig}0.7 (0.1) & \cellcolor{MBsig}0.9 (0.1) & \cellcolor{MBsig}0.7 (0.1) & \cellcolor{MBsig}0.7 (0.1) & \cellcolor{MBsig}\textbf{0.001} & \cellcolor{MLsig}0.7 (0.1) & \cellcolor{MLsig}0.9 (0.1) & \cellcolor{MLsig}0.7 (0.1) & \cellcolor{MLsig}0.7 (0.0) & \cellcolor{MLsig}\textbf{0.001} \\
 & \textbf{Dominant swing time [s]} & \cellcolor{MBsig}0.4 (0.0) & \cellcolor{MBsig}0.5 (0.1) & \cellcolor{MBsig}0.4 (0.0) & \cellcolor{MBsig}0.4 (0.0) & \cellcolor{MBsig}\textbf{0.001} & \cellcolor{MLsig}0.4 (0.0) & \cellcolor{MLsig}0.5 (0.1) & \cellcolor{MLsig}0.4 (0.0) & \cellcolor{MLsig}0.4 (0.0) & \cellcolor{MLsig}\textbf{0.001} \\
 & \textbf{Non-dominant swing time [s]} & \cellcolor{MBsig}0.4 (0.1) & \cellcolor{MBsig}0.5 (0.1) & \cellcolor{MBsig}0.4 (0.0) & \cellcolor{MBsig}0.4 (0.0) & \cellcolor{MBsig}\textbf{0.001} & \cellcolor{MLsig}0.4 (0.0) & \cellcolor{MLsig}0.5 (0.1) & \cellcolor{MLsig}0.4 (0.0) & \cellcolor{MLsig}0.4 (0.0) & \cellcolor{MLsig}\textbf{0.001} \\
 & \textbf{Double limb support time [s]} & \cellcolor{MBsig}0.3 (0.1) & \cellcolor{MBsig}0.4 (0.1) & \cellcolor{MBsig}0.3 (0.0) & \cellcolor{MBsig}0.2 (0.0) & \cellcolor{MBsig}\textbf{0.001} & \cellcolor{MLsig}0.3 (0.1) & \cellcolor{MLsig}0.4 (0.1) & \cellcolor{MLsig}0.3 (0.0) & \cellcolor{MLsig}0.2 (0.0) & \cellcolor{MLsig}\textbf{0.001} \\
 & Stride width [m] & 0.1 (0.0) & 0.1 (0.0) & 0.1 (0.0) & 0.1 (0.0) & 0.231 & \cellcolor{MLsig}0.1 (0.0) & \cellcolor{MLsig}0.1 (0.0) & \cellcolor{MLsig}0.1 (0.0) & \cellcolor{MLsig}0.1 (0.0) & \cellcolor{MLsig}\textbf{0.015} \\
\cmidrule(lr){2-12}
 & \multicolumn{11}{c}{\textbf{Kinematics}} \\
 & Hip Flexion Max [$^\circ$] (Dom) & 36.7 (7.6) & 38.1 (8.5) & 36.4 (7.7) & 36.2 (6.6) & 0.407 & 38.3 (9.8) & 38.2 (10.3) & 38.1 (8.8) & 38.7 (10.8) & 0.950 \\
 & Hip Flexion Max [$^\circ$] (Non-dom) & 35.4 (7.6) & 36.0 (8.6) & 35.1 (7.6) & 35.4 (6.9) & 0.794 & 37.9 (10.0) & 36.5 (9.3) & 38.7 (9.4) & 37.7 (11.1) & 0.187 \\
 & \textbf{Knee Flexion Max [$^\circ$] (Dom)} & \cellcolor{MBsig}66.2 (6.7) & \cellcolor{MBsig}62.6 (7.2) & \cellcolor{MBsig}66.6 (7.0) & \cellcolor{MBsig}68.2 (4.6) & \cellcolor{MBsig}\textbf{0.001} & \cellcolor{MLsig}56.6 (13.7) & \cellcolor{MLsig}52.8 (14.9) & \cellcolor{MLsig}55.6 (12.4) & \cellcolor{MLsig}59.6 (14.1) & \cellcolor{MLsig}\textbf{0.032} \\
 & Knee Flexion Max [$^\circ$] (Non-dom) & \cellcolor{MBsig}63.7 (9.2) & \cellcolor{MBsig}57.8 (10.9) & \cellcolor{MBsig}64.2 (9.0) & \cellcolor{MBsig}67.0 (5.8) & \cellcolor{MBsig}\textbf{0.001} & 54.0 (13.7) & 49.6 (11.9) & 54.8 (12.2) & 55.2 (15.8) & 0.085 \\
 & Ankle Dorsiflexion Max [$^\circ$] (Dom) & \cellcolor{MBsig}12.3 (5.0) & \cellcolor{MBsig}13.2 (6.0) & \cellcolor{MBsig}12.9 (4.9) & \cellcolor{MBsig}10.6 (4.0) & \cellcolor{MBsig}\textbf{0.017} & 24.1 (5.4) & 23.6 (6.7) & 24.6 (4.7) & 23.6 (5.6) & 0.304 \\
 & Ankle Dorsiflexion Max [$^\circ$] (Non-dom) & \cellcolor{MBsig}12.6 (4.9) & \cellcolor{MBsig}13.0 (5.5) & \cellcolor{MBsig}13.6 (4.7) & \cellcolor{MBsig}10.3 (4.0) & \cellcolor{MBsig}\textbf{0.001} & 23.7 (5.5) & 22.4 (5.5) & 24.1 (4.7) & 23.8 (6.4) & 0.371 \\
 & Pelvic Anterior Tilt Max [$^\circ$] & \cellcolor{MBsig}9.2 (6.1) & \cellcolor{MBsig}11.6 (6.8) & \cellcolor{MBsig}8.8 (5.8) & \cellcolor{MBsig}8.1 (5.5) & \cellcolor{MBsig}\textbf{0.013} & 2.4 (37.5) & -0.5 (79.7) & 4.2 (17.9) & 1.5 (21.9) & 0.972 \\
\cmidrule(lr){2-12}
 & \multicolumn{11}{c}{\textbf{Gait Quality Indices}} \\
 & \textbf{GVS Hip Flexion (Dom)} & \cellcolor{MBsig}9.0 (5.1) & \cellcolor{MBsig}11.8 (5.3) & \cellcolor{MBsig}8.9 (4.7) & \cellcolor{MBsig}7.0 (4.7) & \cellcolor{MBsig}\textbf{0.001} & \cellcolor{MLsig}23.0 (3.9) & \cellcolor{MLsig}22.1 (3.8) & \cellcolor{MLsig}22.7 (3.3) & \cellcolor{MLsig}23.7 (4.4) & \cellcolor{MLsig}\textbf{0.038} \\
 & \textbf{GVS Knee Flexion (Dom)} & \cellcolor{MBsig}9.1 (3.9) & \cellcolor{MBsig}11.7 (3.2) & \cellcolor{MBsig}9.5 (3.9) & \cellcolor{MBsig}6.5 (2.8) & \cellcolor{MBsig}\textbf{0.001} & \cellcolor{MLsig}28.8 (4.6) & \cellcolor{MLsig}27.5 (3.7) & \cellcolor{MLsig}28.1 (3.9) & \cellcolor{MLsig}30.3 (5.3) & \cellcolor{MLsig}\textbf{0.001} \\
 & GVS Ankle Dorsiflexion (Dom) & \cellcolor{MBsig}6.7 (3.1) & \cellcolor{MBsig}8.3 (2.7) & \cellcolor{MBsig}6.8 (2.9) & \cellcolor{MBsig}5.4 (3.2) & \cellcolor{MBsig}\textbf{0.001} & 18.4 (3.2) & 18.2 (3.5) & 18.9 (3.0) & 17.9 (3.3) & 0.056 \\
 & \textbf{Cyclogram Area Knee--Hip (Dom)} & \cellcolor{MBsig}1448.4 (342.4) & \cellcolor{MBsig}1250.0 (318.2) & \cellcolor{MBsig}1409.4 (296.2) & \cellcolor{MBsig}1674.2 (324.2) & \cellcolor{MBsig}\textbf{0.001} & \cellcolor{MLsig}1213.4 (537.6) & \cellcolor{MLsig}966.1 (482.5) & \cellcolor{MLsig}1147.8 (497.4) & \cellcolor{MLsig}1414.1 (546.7) & \cellcolor{MLsig}\textbf{0.001} \\
 & \textbf{Cyclogram Area Ankle--Knee (Dom)} & \cellcolor{MBsig}439.0 (311.4) & \cellcolor{MBsig}277.7 (263.8) & \cellcolor{MBsig}405.1 (294.4) & \cellcolor{MBsig}626.8 (288.1) & \cellcolor{MBsig}\textbf{0.001} & \cellcolor{MLsig}342.2 (239.9) & \cellcolor{MLsig}258.0 (129.1) & \cellcolor{MLsig}315.0 (212.8) & \cellcolor{MLsig}416.7 (289.2) & \cellcolor{MLsig}\textbf{0.005} \\
 & \textbf{SSD Knee--Hip (Dom)} & \cellcolor{MBsig}131.9 (58.3) & \cellcolor{MBsig}169.9 (52.2) & \cellcolor{MBsig}133.5 (56.3) & \cellcolor{MBsig}99.7 (48.0) & \cellcolor{MBsig}\textbf{0.001} & \cellcolor{MLsig}372.5 (54.1) & \cellcolor{MLsig}357.6 (45.6) & \cellcolor{MLsig}365.6 (44.0) & \cellcolor{MLsig}388.4 (64.9) & \cellcolor{MLsig}\textbf{0.001} \\
 & \textbf{SSD Ankle--Knee (Dom)} & \cellcolor{MBsig}116.3 (44.5) & \cellcolor{MBsig}145.5 (34.6) & \cellcolor{MBsig}119.9 (42.8) & \cellcolor{MBsig}87.2 (37.4) & \cellcolor{MBsig}\textbf{0.001} & \cellcolor{MLsig}346.6 (46.6) & \cellcolor{MLsig}335.3 (37.3) & \cellcolor{MLsig}343.1 (40.8) & \cellcolor{MLsig}356.3 (55.4) & \cellcolor{MLsig}\textbf{0.030} \\
\cmidrule(lr){2-12}
 & \multicolumn{11}{c}{\textbf{Ground Reaction Forces}} \\
 & \textbf{Anterior GRF Min [BW] (Dom)} & \cellcolor{MBsig}-0.0 (0.0) & \cellcolor{MBsig}-0.0 (0.0) & \cellcolor{MBsig}-0.0 (0.0) & \cellcolor{MBsig}-0.1 (0.0) & \cellcolor{MBsig}\textbf{0.001} & \cellcolor{MLsig}-0.1 (0.0) & \cellcolor{MLsig}-0.0 (0.0) & \cellcolor{MLsig}-0.1 (0.0) & \cellcolor{MLsig}-0.1 (0.0) & \cellcolor{MLsig}\textbf{0.001} \\
 & Vertical GRF Max [BW] (Dom) & \cellcolor{MBsig}1.1 (0.2) & \cellcolor{MBsig}0.9 (0.2) & \cellcolor{MBsig}1.1 (0.1) & \cellcolor{MBsig}1.2 (0.1) & \cellcolor{MBsig}\textbf{0.001} & 0.9 (0.2) & 0.8 (0.2) & 0.9 (0.2) & 0.9 (0.2) & 0.101 \\
 & Medial GRF Min [BW] (Dom) & \cellcolor{MBsig}-0.1 (0.0) & \cellcolor{MBsig}-0.1 (0.0) & \cellcolor{MBsig}-0.1 (0.0) & \cellcolor{MBsig}-0.2 (0.0) & \cellcolor{MBsig}\textbf{0.001} & -0.0 (0.0) & -0.0 (0.0) & -0.0 (0.0) & -0.0 (0.0) & 0.065 \\
 & \textbf{Anterior GRF Min [BW] (Non-dom)} & \cellcolor{MBsig}-0.0 (0.0) & \cellcolor{MBsig}-0.0 (0.0) & \cellcolor{MBsig}-0.0 (0.0) & \cellcolor{MBsig}-0.0 (0.0) & \cellcolor{MBsig}\textbf{0.001} & \cellcolor{MLsig}-0.1 (0.0) & \cellcolor{MLsig}-0.0 (0.0) & \cellcolor{MLsig}-0.0 (0.0) & \cellcolor{MLsig}-0.1 (0.0) & \cellcolor{MLsig}\textbf{0.002} \\
 & Vertical GRF Max [BW] (Non-dom) & \cellcolor{MBsig}1.0 (0.1) & \cellcolor{MBsig}1.0 (0.2) & \cellcolor{MBsig}1.0 (0.1) & \cellcolor{MBsig}1.1 (0.1) & \cellcolor{MBsig}\textbf{0.001} & 0.9 (0.2) & 0.9 (0.2) & 0.9 (0.2) & 0.9 (0.2) & 0.195 \\
 & Medial GRF Min [BW] (Non-dom) & \cellcolor{MBsig}-0.1 (0.0) & \cellcolor{MBsig}-0.1 (0.0) & \cellcolor{MBsig}-0.1 (0.0) & \cellcolor{MBsig}-0.2 (0.0) & \cellcolor{MBsig}\textbf{0.001} & -0.0 (0.0) & -0.0 (0.0) & -0.0 (0.0) & -0.0 (0.0) & 0.377 \\
\bottomrule
\end{tabular}%
}

\caption{\small{\textbf{Clinical and gait-domain characterization of primary speed clusters (marker-based and markerless).} Values are reported as mean (standard deviation; SD) for continuous measures and as \% for categorical variables. Gait-domain variables are grouped into spatiotemporal measures, lower-limb kinematics, gait-quality indices, and ground reaction forces. Results are shown for the full cohort (Total) and for each primary speed cluster (Slow, Medium, Fast), separately for each modality. Between-group comparisons across speed clusters used Kruskal--Wallis tests for continuous variables and chi-squared tests for categorical variables; $p$-values are reported for each modality. Statistically significant $p$-values ($p<0.05$) are shown in \textbf{bold} and highlighted with modality-specific cell colors. Abbreviations: SCIM, Spinal Cord Independence Measure; ROM, range of motion; LEMS, Lower Extremity Motor Score; GVS, Gait Variable Score; SSD, sum of squared differences; GRF, ground reaction force; BW, body weight.}} 
\label{tab:clinical_gait_primaryspeed_markerbased}
\label{tab:clinical_gait_primaryspeed_markerless}
\end{table*}

All participants (Slow/Medium/Fast) were then pooled and cross-sub-cluster
relationships were recomputed on the combined cohort. This provided a cross-speed
``phenotype map'' that tests whether the within-stratum sub-cluster structure is
preserved when speed is no longer used to separate participants, and it links
(i) \emph{which groups are most similar} to (ii) \emph{which feature domains drive
that similarity} (\sfigref{fig:supp_cross_speed_combined_markerless}; marker-based
analogue: \sfigref{fig:supp_cross_speed_combined_markerbased}).

Pairwise subcluster--subcluster similarity across the full dataset (blue = more
similar, orange = more dissimilar) showcased that several sub-clusters remained
close neighbors despite belonging to adjacent speed strata, implying that the
dominant commonalities were not purely speed-driven. By contrast, the largest
separations occurred between sub-clusters that differed in coordination/compensation
patterns, consistent with distinct motor-strategy archetypes.

Capacity-linked variables (cycle and double-support timing, step/stride length)
vary coherently along the Slow$\rightarrow$Medium$\rightarrow$Fast gradient, whereas
coordination and gait-quality descriptors (GVS/SSD and cyclogram areas) preserve
within-stratum separations and therefore also organize similarity \emph{across}
strata. In the markerless profiles, the strongest cross-speed separations are
carried by sagittal-plane hip/knee excursion--coordination patterns and gross braking/propulsion
loading descriptors (robust over many gait cycles), whereas convention- and tracking-sensitive
measures (notably ankle angles) contribute less uniquely, consistent with the
waveform-validation results reported above.

\subsubsection*{Markerless}

\begin{figure}[H]
  \centering
  \includegraphics[width=\linewidth]{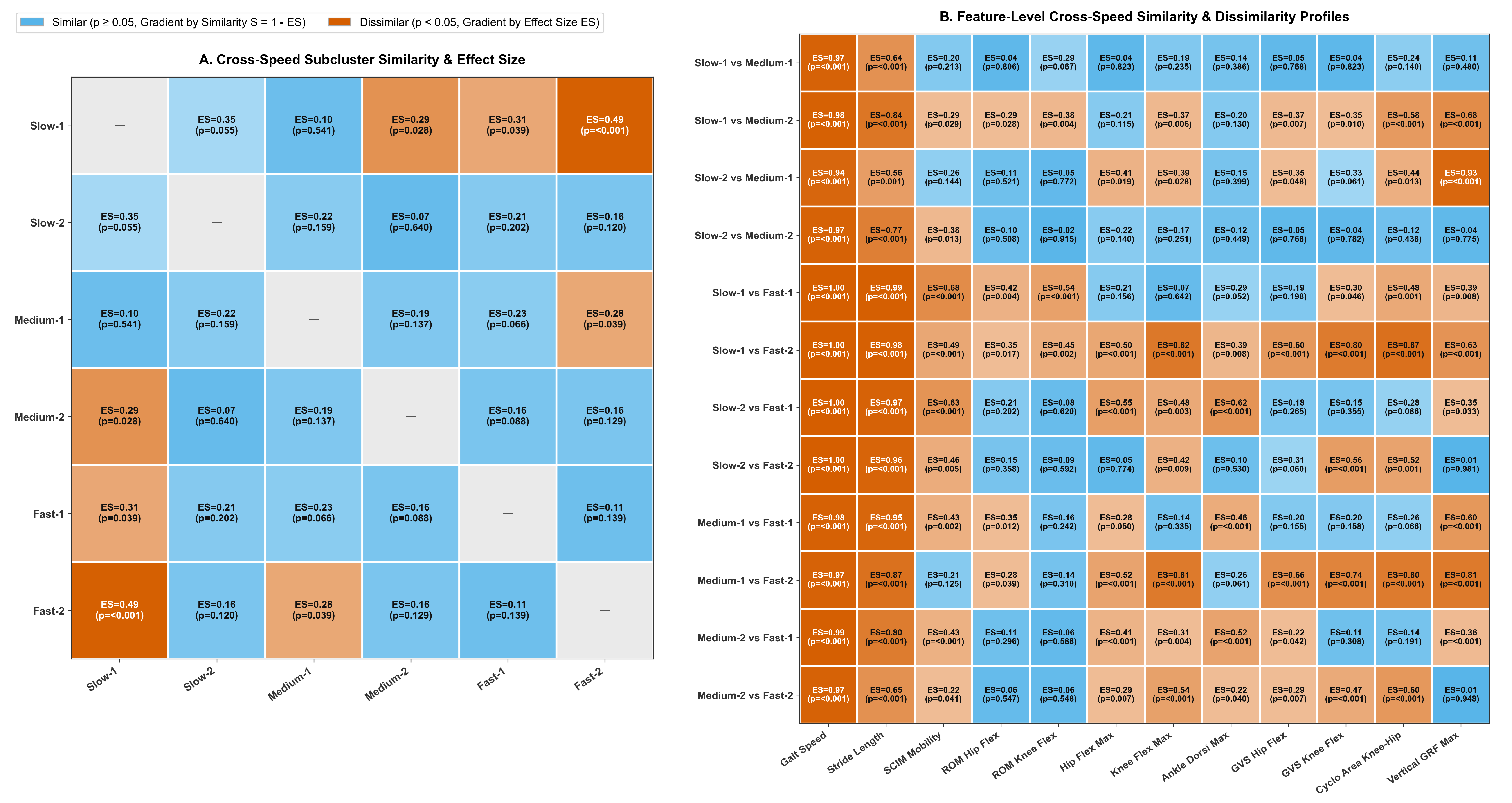}
  \caption{\small{\textbf{Cross-speed-combined phenotype visualisations (markerless pipeline).} Participants from the Slow/Medium/Fast strata are pooled and plotted on a shared normalized scale to visualise how sub-cluster relationships persist across speed.}}
  \label{fig:supp_cross_speed_combined_markerless}
\end{figure}

Across the combined profiles, the clearest markerless separations are carried by
sagittal-plane hip/knee excursion--coordination patterns and gross braking/propulsion
loading descriptors, which are robustly expressed over many gait cycles.
Conversely, features that depend strongly on pelvis/segment coordinate conventions
or distal tracking (notably ankle angles) tend to compress toward the mean and
contribute less uniquely, consistent with the waveform-validation results
reported above. Overall, \sfigref{fig:supp_cross_speed_combined_markerless}
shows that the markerless pipeline preserves a multi-domain phenotype topology
in which speed-related scaling and strategy-related coordination differences
can be interpreted simultaneously, and where similarities between groups are explicitly
visible as cross-stratum near-neighbor relationships.

\subsubsection*{Marker-based}
\begin{figure}[H]
  \centering
  \includegraphics[width=\linewidth]{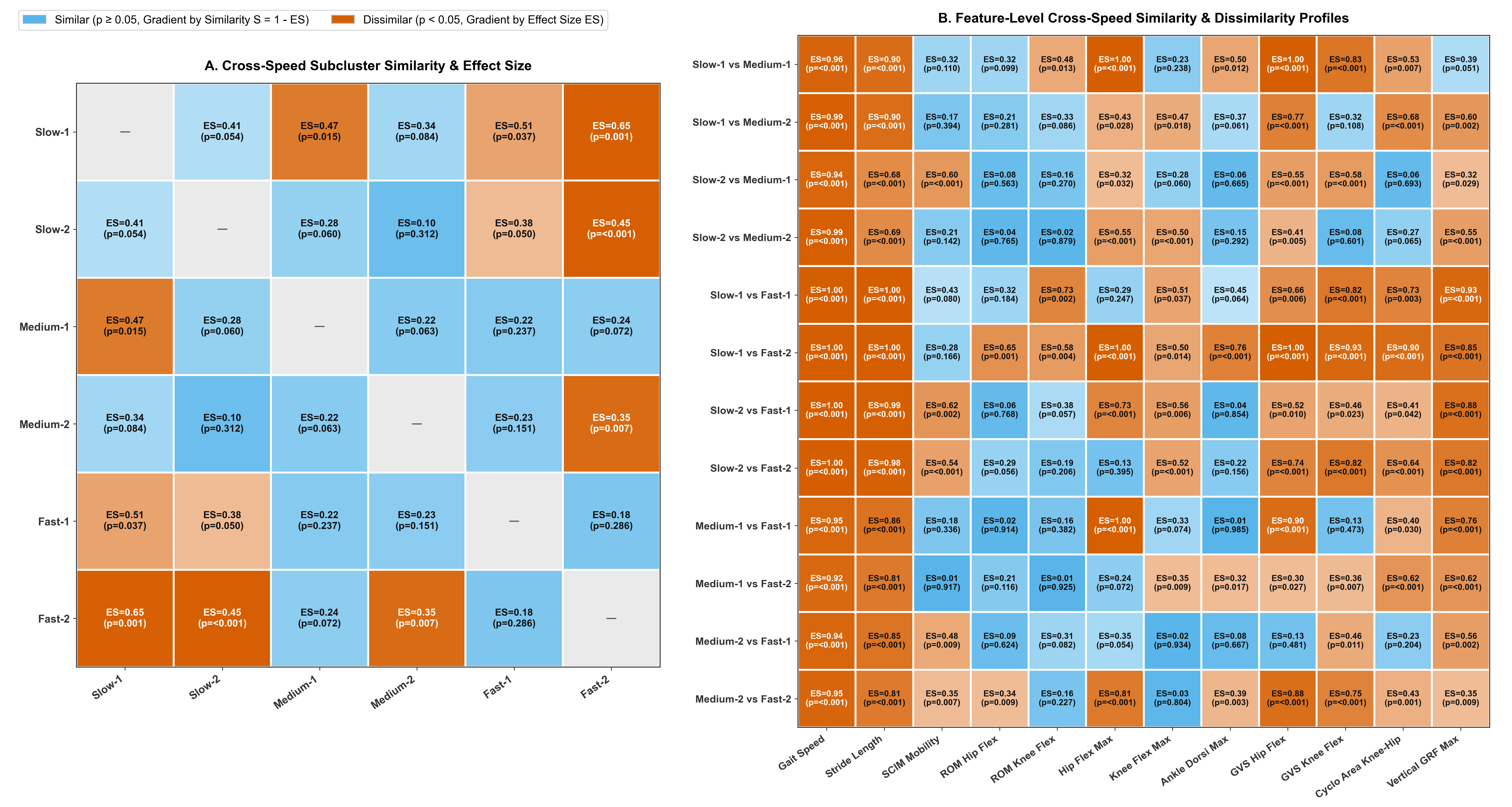}
  \caption{\small{\textbf{Cross-speed-combined phenotype visualisations (marker-based pipeline).} Cluster profiles are shown on a shared normalized scale after combining participants across speed strata.}}
  \label{fig:supp_cross_speed_combined_markerbased}
\end{figure}

To support qualitative interpretation of the clustering structure in the marker-based pipeline, we generated a \emph{cross-speed-combined} visual summary in which participants from the \textbf{Slow}, \textbf{Medium}, and \textbf{Fast} primary strata (and their corresponding Sub-cluster~1/2 assignments) were pooled and compared on a common normalised feature axis (\sfigref{fig:supp_cross_speed_combined_markerbased}). This representation complements the within-stratum tables by addressing a different question: \emph{which sub-clusters remain most similar when walking speed is no longer used to separate individuals, and which feature blocks drive those similarities/differences?}

In the similarity heatmap, each row/column corresponds to a \emph{sub-cluster group} (e.g., Slow--Sub1, Slow--Sub2, etc.). Colour encodes pairwise similarity of the sub-cluster mean feature profiles on the shared scale (blue indicating greater similarity; orange indicating greater dissimilarity). This makes it possible to identify (i) \textbf{near-neighbour relationships} that persist across adjacent speed strata (suggesting a shared mechanics/strategy signature despite different capacity), and (ii) \textbf{isolated or highly separated groups} that represent distinct coordination/compensation phenotypes.

The accompanying feature-block contribution panel provides a compact decomposition of the drivers of between-group separation. Capacity-linked spatiotemporal descriptors (e.g., cycle/double-support timing and step/stride length) tend to organise the Slow$\rightarrow$Medium$\rightarrow$Fast gradient, whereas deviation and coordination descriptors (GVS/SSD/cyclogram-derived features) tend to preserve within-stratum differences and therefore also contribute to similarity structure \emph{across} strata. This cross-speed-combined view is intended for visual synthesis and hypothesis generation; formal inferential comparisons remain those reported in the participant-level tables.

\end{document}